\documentclass[10pt,twocolumn,letterpaper]{article}

\usepackage[pagenumbers]{cvpr} 

\definecolor{cvprblue}{rgb}{0.21,0.49,0.74}
\usepackage[pagebackref,breaklinks,colorlinks,allcolors=cvprblue]{hyperref}

\def\paperID{7168} 
\def\confName{CVPR}
\def\confYear{2025}

\usepackage{array}
\usepackage{caption}
\usepackage{subcaption}
\usepackage{stackengine}
\usepackage{wrapfig}

\usepackage[export]{adjustbox}
\usepackage{booktabs}
\usepackage{multirow}
\usepackage{makecell}
\usepackage{algorithm}
\usepackage{algpseudocode}
\usepackage{url}
\usepackage{tikz}
\usetikzlibrary{calc,spy}
\usepackage{cancel}
\usepackage{diagbox}
\usepackage{pifont}
\usepackage[accsupp]{axessibility}
\usepackage{amsthm}

\newtheorem{definition}{Definition}

\newtheorem{theorem}{Theorem}
\newtheorem*{theorem*}{Theorem}

\newtheorem{proposition}{Proposition}

\newtheorem{lemma}[theorem]{Lemma}

\DeclareSymbolFont{largesymbolsA}{U}{txexa}{m}{n}
\DeclareMathSymbol{\varprod}{\mathop}{largesymbolsA}{16}

\DeclareFontFamily{U}{mathx}{\hyphenchar\font45}
\DeclareFontShape{U}{mathx}{m}{n}{
      <5> <6> <7> <8> <9> <10>
      <10.95> <12> <14.4> <17.28> <20.74> <24.88>
      mathx10
      }{}
\DeclareSymbolFont{mathx}{U}{mathx}{m}{n}
\DeclareMathSymbol{\bigtimes}{1}{mathx}{"91}

\usepackage[utf8]{inputenc} 
\usepackage[T1]{fontenc}    
\usepackage{hyperref}       
\usepackage{url}            
\usepackage{booktabs}       
\usepackage{amsfonts}       
\usepackage{nicefrac}       
\usepackage{microtype}      
\usepackage{xcolor}         

\newcommand*\samethanks[1][\value{footnote}]{\footnotemark[#1]}

\title{Towards Lossless Implicit Neural Representation via Bit Plane Decomposition}

\author{
Woo Kyoung Han$^{1}$ \quad Byeonghun Lee$^1$ \quad Hyunmin Cho$^1$ \quad Sunghoon Im$^2$\thanks{Corresponding author.} \quad Kyong Hwan Jin$^1$\samethanks\\
\vspace{-10pt} \\
$^1$Korea University \quad\qquad $^2$DGIST\\
\vspace{-10pt} \\
{\tt\small \{wookyoung0727, byeonghun\_lee, hyun\_cho, kyong\_jin\}@korea.ac.kr, sunghoonim@dgist.ac.kr}
}

\begin{document}

\tikzstyle{largewindow_w} = [white, line width=0.30mm]
\tikzstyle{smallwindow_w} = [white, line width=0.10mm]
\tikzstyle{largewindow_b} = [blue, line width=0.30mm]
\tikzstyle{smallwindow_b} = [blue, line width=0.10mm]

\tikzstyle{smallwindow_r} = [red, line width=0.10mm]
\tikzstyle{largewindow_r} = [red, line width=0.30mm]

\tikzstyle{smallwindow_k} = [black, line width=0.10mm]
\tikzstyle{largewindow_k} = [black, line width=0.30mm]

\tikzstyle{closeup_b} = [
  opacity=1.0,          
  height=1cm,         
  width=1cm,          
  connect spies, blue  
]
\tikzstyle{closeup_w} = [
  opacity=1.0,          
  height=1cm,         
  width=1cm,          
  connect spies, white  
]
\tikzstyle{closeup_w_2} = [
  opacity=1.0,          
  height=1.7cm,         
  width=1.7cm,          
  connect spies, white  
]
\tikzstyle{closeup_w_3} = [
  opacity=1.0,          
  height=1.35cm,         
  width=1.35cm,          
  connect spies, white  
]
\tikzstyle{closeup_w_4} = [
  opacity=1.0,          
  height=1.cm,         
  width=1.cm,          
  connect spies, red  
]

\tikzstyle{closeup_k_4} = [
  opacity=1.0,          
  height=1.cm,         
  width=1.cm,          
  connect spies, black  
]

\maketitle

\begin{abstract}


We quantify the upper bound on the size of the implicit neural representation (INR) model from a digital perspective.
The upper bound of the model size increases exponentially as the required bit-precision increases.
To this end, we present a bit-plane decomposition method that makes INR predict bit-planes, producing the same effect as reducing the upper bound of the model size. 
We validate our hypothesis that reducing the upper bound leads to faster convergence with constant model size.
Our method achieves lossless representation in 2D image and audio fitting, even for high bit-depth signals, such as 16-bit, which was previously unachievable.
We pioneered the presence of bit bias, which INR prioritizes as the most significant bit (MSB).
We expand the application of the INR task to bit depth expansion, lossless image compression, and extreme network quantization.
Our source code is available at \url{https://github.com/WooKyoungHan/LosslessINR}.

\end{abstract}
\section{Introduction}\label{sec:intro}
 Implicit neural representations (INRs), parameterizing the continuous signals with an artificial neural network (ANN), have been in the spotlight in various areas for recent years. From a signed distance function by \citet{deepsdf_park2019cvpr} to the best-known research by \citet{mildenhall2020nerf} for radiance fields, INR shows promising performance in many fields \cite{chen2021learning,abcd,jdec2024han, Local_Implicit_Grid_CVPR20,lee2021local,Occupancy_Networks,sitzmann2019srns,su2022inras}. The fundamental principle of INR, which aims to train real-world signals with parameters operating in a range and domain of a continuous set, inspired various applications such as super-resolution \cite{chen2021learning,lee2021local,lee2022learning} and a novel view synthesis \cite{mildenhall2020nerf}.

\begin{figure}
    \vspace{-5pt}
    \centering
    \includegraphics[trim=0 0 0 0 ,clip,width = 3.2in]{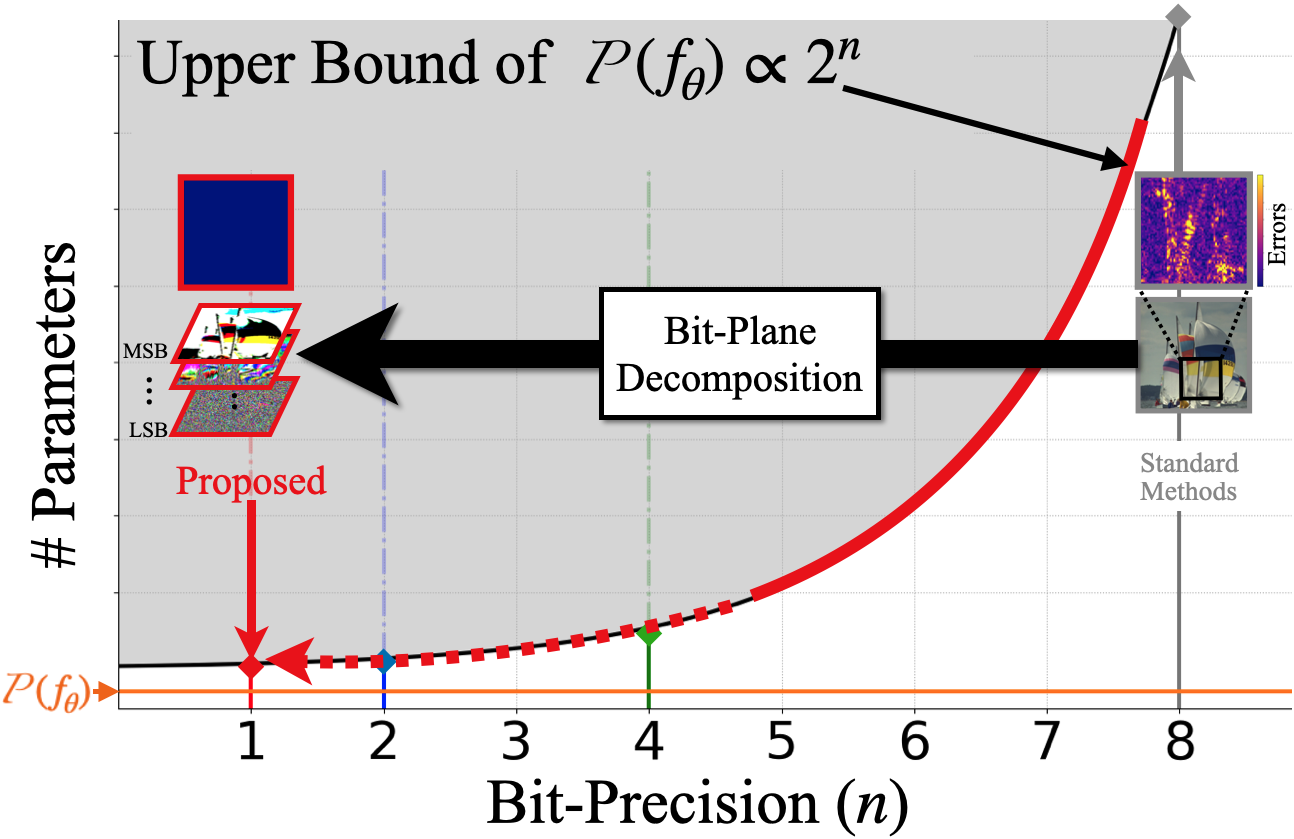}

    \vspace{-10pt}

\caption{\textbf{Overview} of the proposed method and error maps at 1,500 iterations. The upper bound on the number of parameters ($\mathcal{P}(f_\theta)\propto 2^n$) of INR ($f_\theta$) grow proportionally to a bit-precision ($n$). We propose a bit-plane decomposition method, reducing the upper bound, enabling faster convergence, and ultimately achieving a lossless representation. The closer $\mathcal{P}(f_\theta)$ is to the upper bound, the faster it converges, enabling lossless representation}
    \label{fig:concept}
   \vspace{-20pt}
\end{figure}

 \begin{figure*}[ht]
\footnotesize
\centering
\hspace{-18pt}
\stackunder[2pt]{30.79  / 0.246}{\raisebox{0.3in}{\rotatebox{90}{Images}}\includegraphics[trim=0 0 0 0,clip,width = 1.01in]{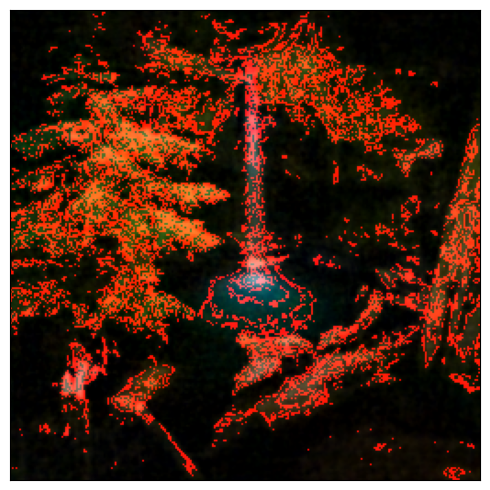}}
\stackunder[2pt]{32.28  / 0.228}{\includegraphics[trim=0 0 0 0,clip,width = 1.01in]{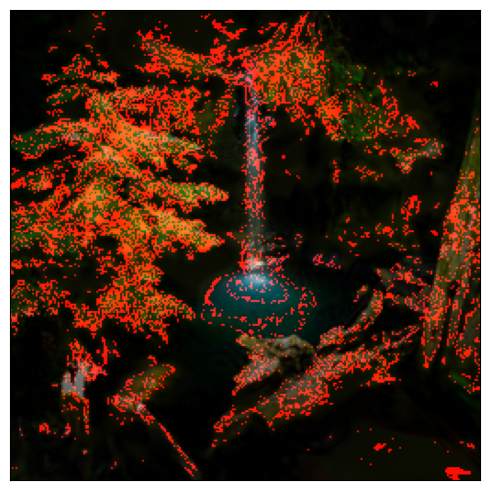}}
\stackunder[2pt]{40.50  / 0.192}{\includegraphics[trim=0 0 0 0,clip,width = 1.01in]{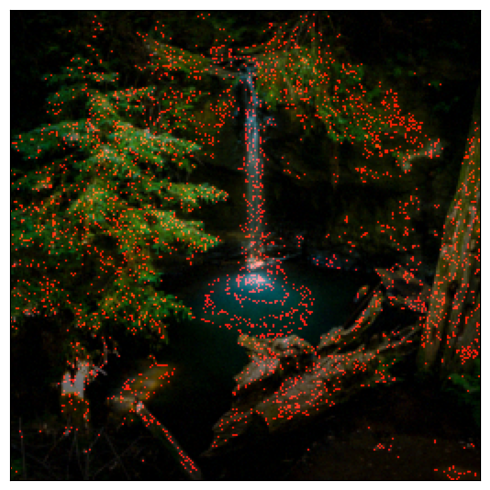}}
\stackunder[2pt]{40.64  / 0.164}{\includegraphics[trim=0 0 0 0,clip,width = 1.01in]{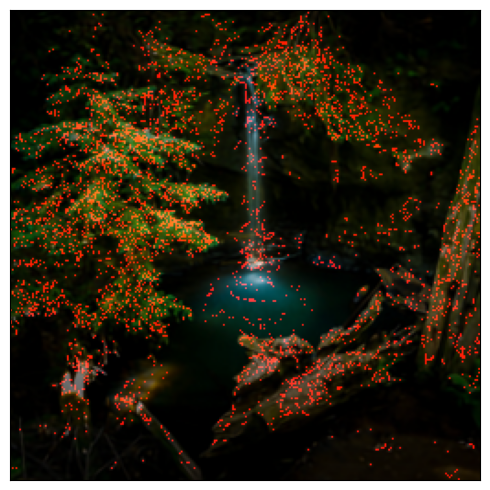}}
\stackunder[2pt]{47.04  / 0.128}{\includegraphics[trim=0 0 0 0,clip,width = 1.01in]{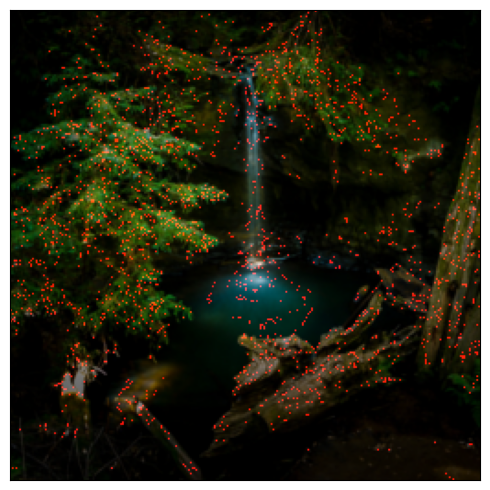}}
\stackunder[2pt]{\color{red}{\textbf{$\mathbf{\infty}$ / 0.00}}}{\includegraphics[trim=0 0 0 0,clip,width = 1.01in]{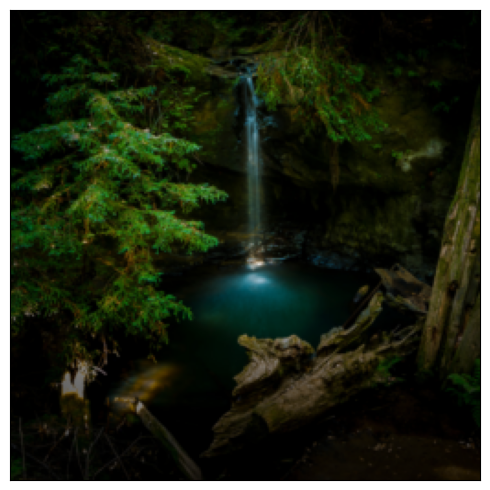}}

\hspace{-2pt}
\raisebox{0.4in}{\rotatebox{90}{Errors}}
\stackunder[2pt]{\includegraphics[trim=0 0 0 0,clip,width = 1.01in]{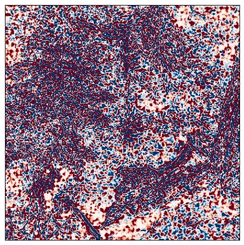}}{ReLU$+$P.E \cite{mildenhall2020nerf}}
\stackunder[2pt]{\includegraphics[trim=0 0 0 0,clip,width = 1.01in]{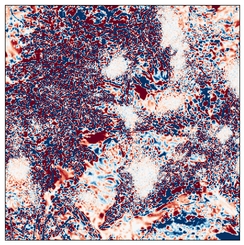}}{WIRE \cite{wirewavelet2023}}
\stackunder[2pt]{\includegraphics[trim=0 0 0 0,clip,width = 1.01in]{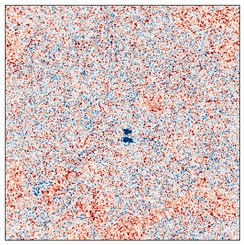}}{Gauss \cite{activationinr_gauss_etc}}
\stackunder[2pt]{\includegraphics[trim=0 0 0 0,clip,width = 1.01in]{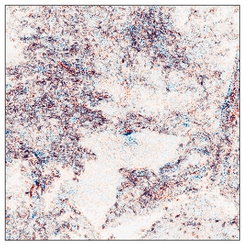}}{SIREN \cite{sitzmann2019siren}}
\stackunder[2pt]{\includegraphics[trim=0 0 0 0,clip,width = 1.01in]{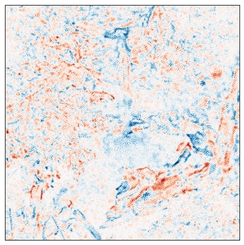}}{FINER \cite{liu2023finer}}
\stackunder[2pt]{\includegraphics[trim=0 0 0 0,clip,height = 1.01in]{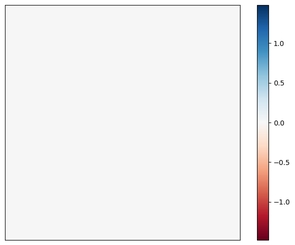}}{\hspace{-12pt}Ours}
\vspace{-10pt}
\caption{\textbf{Visual Demonstration} of representing 2D Image (PSNR(dB) $\uparrow$/ Bit-Error-Rate (BER) $\downarrow$ at top of images). ReLU with position encoding (P.E) \cite{mildenhall2020nerf}, WIRE \cite{wirewavelet2023}, gaussian activation \cite{activationinr_gauss_etc}, SIREN 
 \cite{sitzmann2019siren}, FINER\cite{liu2023finer} and ours. We highlight the occurrence of significant errors with red dots.}
\vspace*{-15pt}
\label{fig:demo_qual}
\end{figure*}
 

However, research exploring precision close to the continuous range, i.e., analog, has not been actively pursued. Computers operate with digital signals, not analog ones, with values constrained by quantization, such as 8-bit or 16-bit for images and 24-bit for audio. 
Therefore, signal representation necessitates the concept of quantization, where bit precision-the number of bits required to represent the signal-serves a pivotal function. Furthermore, lossless representation is defined as satisfying the given bit precision across all input values.
Although existing methods produce high-quality images, achieving complete lossless representation remains challenging, especially for high dynamic range images like 16 bits.


 In this paper, based on ``The implicit ANN approximations with described error tolerance and explicit parameter bounds" by \citet{mathdeeplearningtheory}, we quantify the upper bound of the size of an INR with given bit-precision. 
 \cref{fig:concept} show that the theoretical upper bound of model size increases as an exponential function proportional to required bit-precision. We suggest a method to reduce the quantified upper bound by bit-plane decomposition. We decompose the signal into bit-planes and represent them, leading to lossless representation.
  Our method is based on the hypothesis that INR reaches the target error—the maximum allowable error to ensure a lossless representation—as the upper bound approaches the model size.
  We validate our hypothesis through experiments in \cref{fig:valid_fg1b}. As in \cref{fig:demo_qual}, our method makes INR represent a lossless signal in a bit-for-bit manner, which was previously unachievable.   
 We discovered that INRs learn the most significant bits (MSBs) faster than the least significant bits (LSBs) regardless of activations. We named the observed phenomenon `\textit{Bit Bias}.' Additionally, we empirically show that the frequency of the bit axis also has a bias in learning. We demonstrate three applications utilizing our method: lossless compression through lossless representation, bit-depth expansion through a bit axis, and ternary INR through robustness on weight quantization.

In summary, our main contributions are as follows:
\begin{itemize}

    \item We quantify the upper bound on the number of parameters of the implicit neural network based on the given bit precision.
    
    \item We propose a bit-plane decomposition for lossless implicit neural representation and validate our hypothesis that reducing bit precision lowers the upper bound on the number of parameters, leading to faster convergence to lossless representation compared to other networks.
    
    \item We discovered the existence of bit bias, where most significant bits converge faster than least significant bits, as in spectral bias.

    \item Our approach extends the application of INR to lossless compression, bit-depth expansion, and model quantization.

\end{itemize}

\section{Related Work}\label{sec:related}

\textbf{Implicit Neural Representation} INRs present signals with an ANN that takes spatial coordinates as input. Research to improve the performance of INRs has been conducted to enhance the low representing power of multi-layer perceptrons (MLP). \citet{sitzmann2019siren}, addressed this challenge by employing a sine activation function and inspired researchers to apply various activations to INRs \cite{liu2023finer,activationinr_gauss_etc,wirewavelet2023}. 
Research about enhancing the capacity of INR \cite{chen2023neurbf,lindell2022bacon,muller2022instantngp,xie2023diner,yang2022polynomial_pnf} have been conducted. \citet{muller2022instantngp}, and \citet{xie2023diner} utilize hash-tables in INRs to enable faster training and accommodate larger signals compared to other methods. However, previous studies were not interested in making concrete lossless representations.
The application of INR is also gaining attention. Specifically, several approaches \cite{dupont2021coin,dupont2022coin++,strumpler2022compressINR,guo2023compression_combiner} extend the application of INR to lossy compression.
Recently, the methods \cite{guo2023compression_combiner,he2023recombiner} demonstrate remarkable performance by learning a Bayesian INR and encoding a sample.

\begin{figure*}[t]
\footnotesize
\centering
\includegraphics[height = 1.8in]{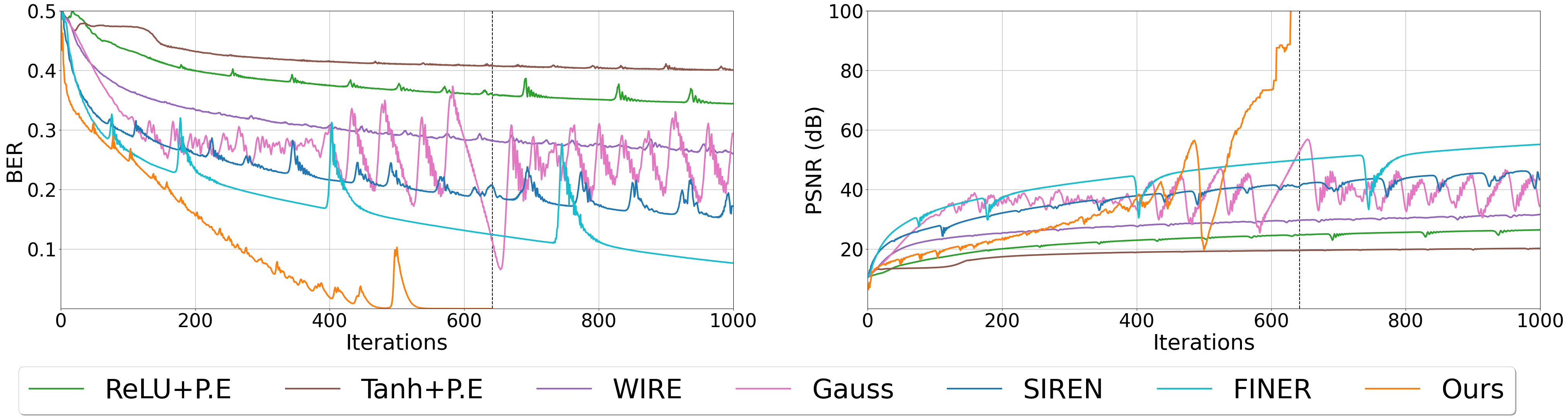}
\vspace{-10pt}
\caption{\textbf{Training curve } on a single image of DIV2K\cite{DIV2kdataset} dataset. Bit-Error-Rate(BER) (left) and PSNR (right). Vertical lines indicate the iteration when the model achieves lossless representation.}
\vspace{-10pt}
\label{fig:learning_curve}
\end{figure*}

Even by increasing the parameters significantly, INRs have difficulty achieving representations aimed at high accuracy as in \cref{fig:learning_curve}. We focus on reducing the required model size based on bit precision and propose a bit-plane decomposition method to achieve lossless representation with a sufficiently sized model. Our approach offers new applications that were not proposed in existing INRs. We devise a lossless compression approach by combining lossless representation with existing methods. Using bit depth as an axis enables a bit-depth expansion through extrapolation. We propose a ternary INR that utilizes the robustness of our approach to weight quantization.

%


\noindent\textbf{Spectral Bias} \citet{spectralbias} have shown the presence of a \textit{spectral bias} which makes it challenging for INR to learn high-frequency components. To address this challenge, approaches \cite{lee2021local,mildenhall2020nerf,sitzmann2019siren,tancik2020fourfeat} that map coordinates into sinusoidal functions have been proposed. The prior works \cite{mildenhall2020nerf,tancik2020fourfeat} suggested fixed frequencies to solve spectral bias known as position encoding, while \citet{lee2021local} proposed a learnable position encoding which enables INR to learn continuous Fourier spectra.
We reinterpret the spectral bias in the bit-plane aspect, which is our proposed method's core concept. Bit-plane is mainly used for image dequantization \cite{abcd,punnappurath2021little} or vision model quantization tasks \cite{zhu2020xornet}. We found the existence of a similar phenomenon like spectral bias, which we call \textit{bit bias}. Our method achieves lossless representations efficiently by mitigating this phenomenon.

\section{Method}\label{sec:problemformulation}
In this section, we quantify the upper bound of the INR based on the given bit-precision, grounded in theory \cite{mathdeeplearningtheory}. Our proposed bit-plane decomposition method reduces the quantified upper bound and accelerates the attainment of the target error bound, leading to lossless representation.
\subsection{Preliminary}\label{sec:prelim}
\noindent\textbf{Quantization} A quantization is an inevitable function for all signals to convert analog to digital, which is defined below:
\vspace{-5pt}
\begin{align}
 \mathcal{Q}_n(\hat x) := \arg\min_{{{x}}} &|| {x}-\hat{{x}} ||_1 \quad ( x \in {Q}_n),
 \label{eq:quantization}
\end{align}
where $n\in \mathbb{N}$ denotes bit precision, and $ Q_n \subset \mathbb{Q} $ is a finite set. We assume the elements of $Q_n$ have normalized and uniformly distributed (e.g., $Q_8 =\{0,\frac{1}{2^8-1},\frac{2}{2^8-1} ... 1\}$ for 8-bit images).
We set the range dimension of a function as 1 without loss of generality.
Let $h: \mathbb{R}^d \rightarrow \mathbb{R}$ be a continuous and analog function in $d$-dimensional space and let $h_n$ be a digital function with $n$-bits precision: 
\begin{align}
    h_n : \mathbb{R}^d \overset{h}{\longrightarrow} \mathbb{R} \overset{\mathcal{Q}_n(\cdot)}{\longrightarrow} Q_n.
    \label{eq:mapping}
\end{align}
With our assumption, a ceiling of error $\epsilon(n)$ between $h$ and $h_n$ is a function of precision defined as below:
\begin{equation}
    \epsilon(n) := \frac{1}{2(2^{n}-1)}.
    \label{eq:required_threshold}
\end{equation}
\noindent\textbf{Explicit Bounds} 
\citet{mathdeeplearningtheory} have demonstrated the explicit upper bounds of the number of parameters of ANNs.
This provides a specific number of parameters regarding the particular error tolerance proposed in the universal approximation theory (UAT) \cite{cybenko1989approximationuat}.

Let $L \in \mathbb{R}$ is a Lipschitz constant that satisfy $\forall \mathbf{x}_i,\mathbf{x}_j \in \mathbb{R}^d$ that $||h(\mathbf{x}_i)-h(\mathbf{x}_j)||_1 \leq L||\mathbf{x}_i-\mathbf{x}_j||_1$. Then, there exists MLP ($:=h_\theta$) that holds:
\begin{enumerate}[leftmargin=0.3in]
    \item The number of parameters: $\mathcal{P}(h_\theta)\leq \mathfrak{C}\epsilon^{-2d} (:=\mathcal{U}_d)$ ,
    \item The ceiling of the error: $\sup||h_\theta(\mathbf{x}) - h(\mathbf{x})||_1\leq \epsilon$,
\end{enumerate}
 where $\mathcal{P}(\cdot)$ is the number of parameters and $\mathfrak{C}$ is a constant determined by the condition of the domain. We provide details of the theorem and $\mathfrak{C}$ in the supplement material.  
\subsection{Problem Formulation}\label{sec:prob}
\noindent\textbf{Lossless Representation} A lossless representation requires having $n$-bit precision, where $n$ remains identical to that of the ground truth digital signal at every point. The INR, coordinate-based MLP ($h_\theta$), aims to parameterize a function $h$ with trainable parameters, $\theta$. Since our target is to represent $h_n$, the output of $h_\theta$ should map to $Q_n$ for digital representation as \cref{eq:mapping}. The parameterized function $h_\theta:\mathbb{R}^d \rightarrow \mathbb{R}$ achieves $n$-bit precision with respect to an analog function $h:\mathbb{R}^d \rightarrow \mathbb{R}$ at $(\mathbf{x}, {h}_{n}{(\mathbf{x})})$, if and only if $\mathbf{x}\in \mathbb{R}^d$, the predicted output $\hat{h}_{\theta}{(\mathbf{x})}$ satisfies:
    \begin{equation}
    \small
    \mathcal{Q}_n(\hat{h}_\theta(\mathbf{x})) = h_n(\mathbf{x}) \leftrightarrow             \hat{h}_{\theta}{(\mathbf{x})} \in [h_n(\mathbf{x})-\epsilon(n), h_n(\mathbf{x})+\epsilon(n)].
    \label{eq:pointwise_lossless}
    \end{equation}
 where $\hat{h}_\theta$ indicate predicted values. If a parameterized function $h_{\theta}$ satisfies \cref{eq:pointwise_lossless} at $\forall \mathbf{x}\in\mathcal{X}$, it is defined to have $n$-bit precision with respect to an analog function $h:\mathbb{R}^d \rightarrow \mathbb{R}$. i.e.:
\begin{equation}
 {\underset{\mathbf{x}\in\mathcal{X}}{\sup}||h_n(\mathbf{x})-\hat{h}_{\theta}(\mathbf{x})||_1} \leq \epsilon(n).
 \label{eq:def_lossless}
\end{equation}
 The lossless representation is identical to make $h_\theta$ satisfy \cref{eq:def_lossless}.
 According to UAT \cite{cybenko1989approximationuat}, it is known that if there is a sufficient number of parameters \cref{eq:def_lossless} be satisfied. However, efficient methods to achieve \cref{eq:def_lossless}, especially for large $n$, have not been well studied.

\begin{table*}[t]
\centering
\setlength{\tabcolsep}{1.2pt}
\footnotesize
\begin{tabular}{
c|l|c|c
||>{\centering\arraybackslash}p{0.85cm}|>{\centering\arraybackslash}p{0.85cm}
||>{\centering\arraybackslash}p{0.85cm}|>{\centering\arraybackslash}p{0.85cm}
||>{\centering\arraybackslash}p{0.85cm}|>{\centering\arraybackslash}p{0.85cm}
||>{\centering\arraybackslash}p{0.85cm}|>{\centering\arraybackslash}p{0.85cm}
||>{\centering\arraybackslash}p{0.85cm}|>{\centering\arraybackslash}p{0.85cm}
||>{\centering\arraybackslash}p{0.85cm}|>{\centering\arraybackslash}p{0.85cm}
||>{\centering\arraybackslash}p{0.85cm}|>{\centering\arraybackslash}p{0.85cm}
}
\toprule
\hline
\multicolumn{4}{c||}{Method} & \multicolumn{2}{c||}{Tanh\footnotesize{+P.E} \cite{mildenhall2020nerf}} & \multicolumn{2}{c||}{ReLU\footnotesize{+P.E} \cite{mildenhall2020nerf}} & \multicolumn{2}{c||}{WIRE \cite{wirewavelet2023}} & \multicolumn{2}{c||}{Gauss \cite{activationinr_gauss_etc}} & \multicolumn{2}{c||}{SIREN \cite{sitzmann2019siren}}& \multicolumn{2}{c||}{FINER \cite{liu2023finer}} &\multicolumn{2}{c}{\textbf{Ours}}\\
\hline
\hline
\multirow{5}{*}{\rotatebox[origin=c]{90}{16-Bit}} & \multicolumn{3}{c||}{Iterations ($\downarrow$)} &\multicolumn{12}{c||}{ 5000}&\multicolumn{2}{c}{\textbf{3450}$(\pm 877)$}\\
\cline{2-18}
& \multirow{2}{*}{\tiny{TESTIMAGES} \cite{asuni2013testimages}} & PSNR&SSIM ($\uparrow$)& 25.63 &0.5447 & 35.91 & 0.8229 &45.36 &0.9341 &69.40&0,9928 &\textcolor{black}{78.52}&0.9969 & \textcolor{blue}{80.17} & \textcolor{blue}{0.9995} & \textcolor{red}{\textbf{$\mathbf{\infty}$}}& \textcolor{red}{{1.0000}} \\
\cline{3-18}
&   & RMSE&BER ($\downarrow$)& 3428.1& 0.4224& 1049.5 & 0.3794&359.17 & 0.3232& 22.217 &0.2167&7.7672 & \textcolor{black}{0.1544}&\textcolor{blue}{6.4222}& \textcolor{blue}{0.1498}&\textcolor{red}{{0.0000}}&\textcolor{red}{{0.0000}}\\
\cline{2-18}
&\multirow{2}{*}{MIT-5k \cite{fivek}} & PSNR&SSIM ($\uparrow$) & 26.95&0.4644 &37.43&0.8064 &45.61&0.9241 &66.90& 0.9795 &\textcolor{black}{78.39}&0.9970&\textcolor{blue}{86.48}&\textcolor{blue}{0.9987} & \textcolor{red}{$\mathbf{\infty}$}& \textcolor{red}{{1.0000}}\\
\cline{3-18}
&   &RMSE&BER ($\downarrow$)&2943.5 & 0.4195& 880.50 &0.3734&343.46 &0.3202&29.611 &0.2306 &7.8885& \textcolor{black}{0.1560}& \textcolor{blue}{3.1079}&\textcolor{blue}{0.1138}&\textcolor{red}{{0.0000}}&\textcolor{red}{{0.0000}}\\
\hline
\hline
\multirow{5}{*}{\rotatebox[origin=c]{90}{8-Bit}} & \multicolumn{3}{c||}{Iterations ($\downarrow$)} &\multicolumn{12}{c||}{ 1000}& \multicolumn{2}{c}{\textbf{790$(\pm 109)$}} \\
\cline{2-18}
 & \multirow{2}{*}{DIV2K \cite{DIV2kdataset}} & PSNR&SSIM ($\uparrow$) &21.08& 0.4956  &27.68& 0.8124 &35.69& 0.9572 &\textcolor{black}{54.70}&0.9940 &47.19&0.9955& \textcolor{blue}{55.03}&\textcolor{blue}{0.9989}& \textcolor{red}{$\infty$}&\textcolor{red}{1.0000}\\
\cline{3-18}
&   &RMSE&BER ($\downarrow$)&22.519 &0.3921& 10.533  &0.3318& 4.1864 &0.2365&0.4694& \textcolor{black}{0.0953}&1.1144&0.1343&\textcolor{blue}{0.4519}&\textcolor{blue}{0.0528}&\textcolor{red}{{0.0000}}& \textcolor{red}{{0.0000}}\\
\cline{2-18}
&   \multirow{2}{*}{Kodak \cite{Kodakdataset}} & PSNR&SSIM ($\uparrow$) & 23.94& 0.5894 &30.94& 0.8473 &37.86& 0.9532 & \textcolor{black}{48.28}&0.9864 &47.28&0.9919&\textcolor{blue}{55.93}&\textcolor{blue}{0.9985} &\textcolor{red}{$\infty$}&\textcolor{red}{1.0000}\\
\cline{3-18}
&   &RMSE&BER ($\downarrow$)& 16.201 &0.3709& 7.2378 & 0.3036& 3.2634 & 0.2234& 0.9830 &0.1456&1.1029 &\textcolor{black}{0.1382}& \textcolor{blue}{0.4074} &\textcolor{blue}{0.0447}& \textcolor{red}{0.0000}&\textcolor{red}{{0.0000}} \\
   \hline
   \bottomrule
\end{tabular}
\caption{\textbf{Quantitative comparison} on 16-bit (top) and 8-bit (bottom) image fitting with existing INR methods. The iteration number of our methods indicates `$mean(\pm std)$' for the total dataset. The text color \textcolor{red}{red}/\textcolor{blue}{blue} indicates the best and second-best, respectively.}
   \vspace{-5pt}
   \label{tab:quan_main}
\end{table*}

\noindent\textbf{Upper Bound}
Based on the \cref{sec:prelim}, we derive the upper bound ($\mathcal{U}_d$). The `upper bound' represents the threshold where more parameters don't improve the representation.
$\mathcal{U}_d$ takes bit precision ($n$), the number of bits required to represent a signal, and signal's dimension ($d$) as a dependent variable:
\begin{equation}
 \mathcal{U}_d(n) := \mathfrak{C} \epsilon(n)^{-2d} = \mathfrak{C} (2^{n+1}-2)^{2d}.
    \label{eq:boundfunction}
\end{equation}
In conclusion, the upper bound $\mathcal{U}_d(n)$ of the $h_\theta$ increases exponentially as a function of the given bit $n$.
We hypothesize that $h_\theta$ achieves \cref{eq:def_lossless} faster more efficiently as $\mathcal{P}(h_\theta)$ approaches $\mathcal{U}_d$. 
We validate the hypothesis in \cref{sec:quali_results} by adjusting $n$ of target signals. We decrease the upper bound $\mathcal{U}_d(n)$  by reducing the required precision ($n$).

\subsection{Methodology}
\noindent\textbf{Bit-Plane Decomposition} 
 Let $n$-bit images be $\mathbf{I}_n:\mathbb{R}^2\rightarrow Q^3_n$. By employing \cref{eq:boundfunction,eq:def_lossless}, $\mathcal{U}_d(n)$ that required to ensure representing lossless $\mathbf{I}_8~\mbox{and}~
\mathbf{I}_{16}$ are significant numbers ($>10^{10}\cdot\mathfrak{C},10^{20}\cdot\mathfrak{C}$, respectively).
To this end, we suggest a \textit{bit-plane decomposition} method for implicit neural representation. Instead of $n$-bit images, we decompose images into bit-planes and represent them. Bit-planes are binary images\footnote{Note that $\{0,1\} = Q_1$} $\mathbf{B}^{(i)} \in \{0,1\}^{H\times W\times 3}$ that satisfy: 
\begin{align}
  \mathbf{I}_n = \frac{1}{2^n-1}\sum_{i=0}^{n-1} 2^i \mathbf{B}^{(i)},\label{eq:bitdecomp} 
\end{align} 
where, $i$ denote $i$-th least significant bit-plane. Our method reduces a bit precision $n$ to $1$, thereby reducing $\mathcal{U}_d(1) = 16\mathfrak{C}$.
 As a result, our approach brings the number of parameters closer to \cref{eq:boundfunction} and makes it easier to achieve \cref{eq:def_lossless}.
 
 A straightforward method for representing an $n$-bit signal is employing parallel sequence of INRs i.e. $[f^{(i)}_{\theta}]$, each representing bit-plane $[\mathbf{B}^{(i)}]$. Inspired by the recent de-quantization approach proposed by \citet{abcd}, we propose a method that employs an additional coordinate ($i$) to represent an $n$-bit image, as shown below:
\begin{align}
 \mathbf{B}^{(i)}(\mathbf{x}) \simeq f_\theta(\mathbf{x},i),
\label{eq:proposedmethod}
\end{align}
where, $\mathbf{x}\in \mathbb{R}^2$ indicates spatial coordinate. \cref{eq:proposedmethod} is motivated by the fact that each element of $[\mathbf{B}^{(i)}]$ is not independent but highly correlated. In other words, our proposed method considers an image as a 3-dimensional function with the bit coordinate. 

\begin{figure}[h]
    \vspace{-5pt}
    \includegraphics[width=3.15in]{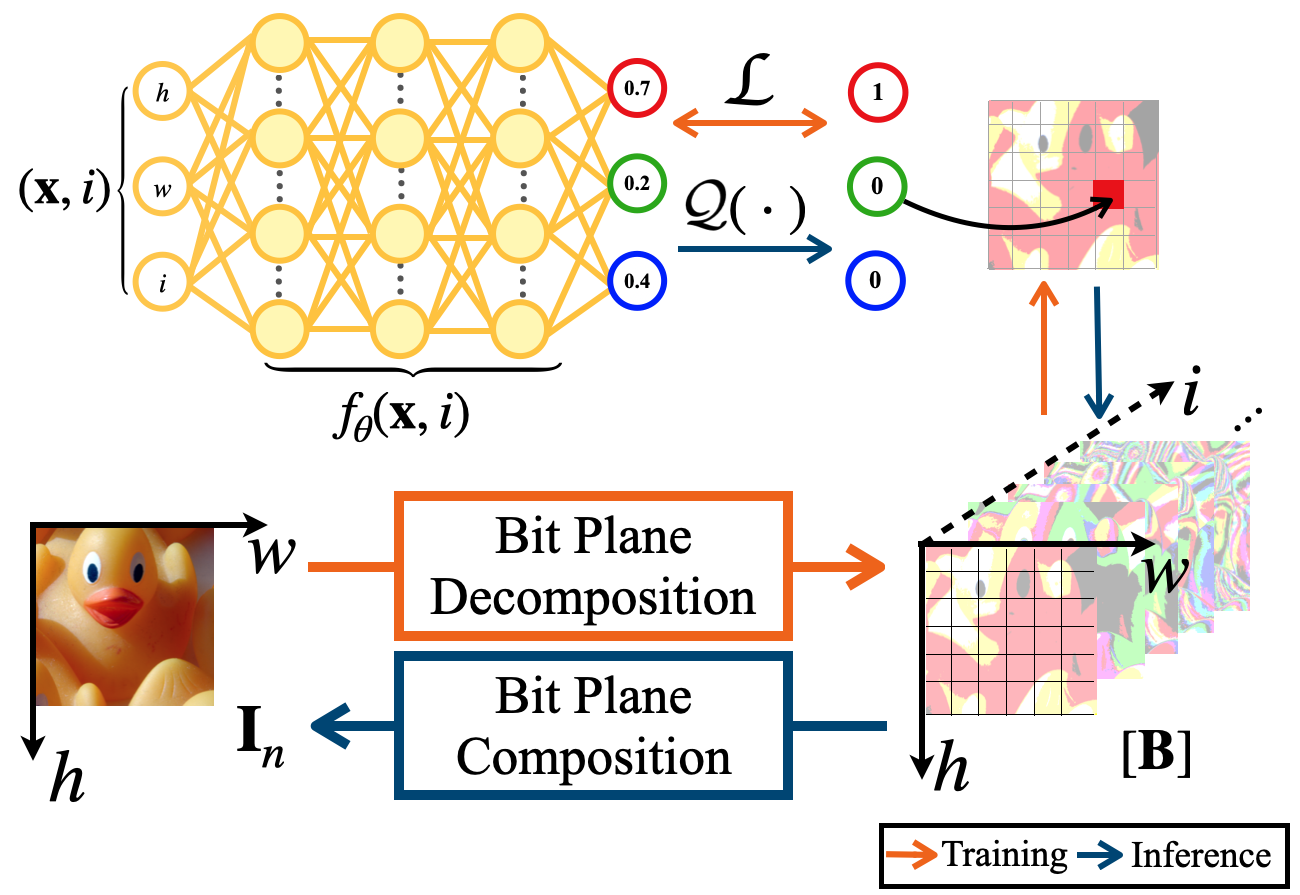}
    \vspace{-5pt}
    \caption{Overall process of our proposed method. We improve the performance of INR by lowering the upper bound of the number of parameters $(\mathcal{U})$ and achieve \textbf{lossless neural representation.}}\label{fig:main}
    \vspace{-10pt}
\end{figure}

\noindent\textbf{Loss Function} We optimize our parameters with the equation below: 
\begin{equation}
\vspace{-5pt}
    \hat{\theta} = \arg\min_{\theta}\mathcal{L}(\mathbf{B}^{(i)}(\mathbf{x}),\hat{f}_{\theta}(\mathbf{x},i)),
    \vspace{-5pt}
\end{equation}
where, $\mathcal{L}$ indicates a loss function. In our approach, candidate loss function $\mathcal{L}$ to optimize parameters $\theta$ include general regression losses such as $||\cdot||_{p}$ with $p=1,2$. Furthermore, the optimization problem in our approach with $k=1$ can be considered as a binary classification problem. We observed that the binary cross-entropy (BCE) loss effectively optimizes $\theta$ and conducted the ablation study in Sec. \ref{sec:abl}.

\cref{fig:main} shows the overall process of our method. After training, we reassemble quantized images to $n$-bit precision representation using \cref{eq:quantization,eq:mapping,eq:bitdecomp} i.e.:
\begin{align}
    \mathbf{I}_{n}(\mathbf{x};\theta) : \underbrace{\overbrace{(\mathbf{x},i) \overset{f_\theta}{\mapsto} [\mathbf{\hat{B}}]}^{\text{Training}} \overset{\mathcal{Q}(\cdot)}{\longrightarrow} [\mathbf{B}]\overset{\cref{eq:bitdecomp}}{\longrightarrow} \mathbf{I}_n}_{\text{Inference}}.
    \label{eq:eval_process}
\end{align}
Note that $\mathbf{\hat{B}} \in \mathbb{Q}^{H\times W\times3}$ which satisfy \cref{eq:pointwise_lossless} for all coordinates. In summary, our method represent $\mathbf{I}_n$ as below:
\begin{align}
  \mathbf{I}_n(\mathbf{x};{\theta}) = \frac{1}{2^n-1}\sum_{i=0}^{n-1} 2^i \mathcal{Q}(\hat{f}_\theta(\mathbf{x},i)).\label{eq:summarize_method} 
\end{align}
The extended description of the method is in the supplement material for $n$-ary representations of \cref{fig:valid_fg1b}.

\begin{figure*}[t]
\footnotesize
\centering

\hspace{-2mm}
\raisebox{0.3in}{\rotatebox{90}{DIV2K \cite{DIV2kdataset}}}%
\hspace{-1.0mm}
\begin{tikzpicture}[x=6cm, y=6cm, spy using outlines={every spy on node/.append style={smallwindow_w}}]
\node[anchor=south] (FigA) at (0,0) {\includegraphics[trim=0 0 0 0 ,clip,width=0.88in]{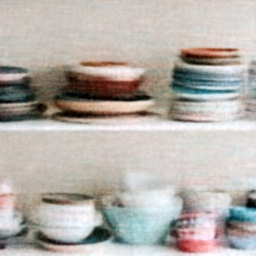}};
\spy [closeup_w_3,magnification=2] on ($(FigA)+( +0.103, -0.12)$) 
    in node[largewindow_w,anchor=east]       at ($(FigA.north) + (+0.035,-0.128)$);
\end{tikzpicture}
\hspace{-1.mm}
\begin{tikzpicture}[x=6cm, y=6cm, spy using outlines={every spy on node/.append style={smallwindow_w}}]
\node[anchor=south] (FigA) at (0,0) {\includegraphics[trim=0 0 0 0 ,clip,width=0.88in]{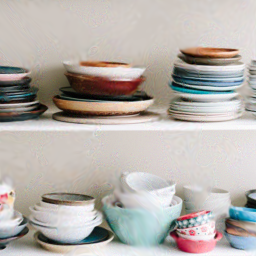}};
\spy [closeup_w_3,magnification=2] on ($(FigA)+( +0.103, -0.12)$) 
    in node[largewindow_w,anchor=east]       at ($(FigA.north) + (+0.035,-0.128)$);
\end{tikzpicture}
\hspace{-1.mm}
\begin{tikzpicture}[x=6cm, y=6cm, spy using outlines={every spy on node/.append style={smallwindow_w}}]
\node[anchor=south] (FigA) at (0,0) {\includegraphics[trim=0 0 0 0 ,clip,width=0.88in]{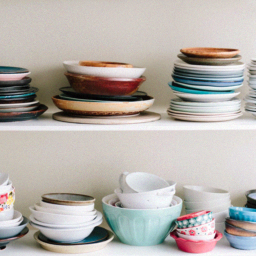}};
\spy [closeup_w_3,magnification=2] on ($(FigA)+( +0.103, -0.12)$) 
    in node[largewindow_w,anchor=east]       at ($(FigA.north) + (+0.035,-0.128)$);
\end{tikzpicture}
\hspace{-1.0mm}
\begin{tikzpicture}[x=6cm, y=6cm, spy using outlines={every spy on node/.append style={smallwindow_w}}]
\node[anchor=south] (FigA) at (0,0) {\includegraphics[trim=0 0 0 0 ,clip,width=0.88in]{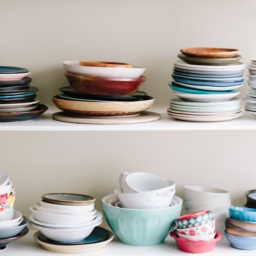}};
\spy [closeup_w_3,magnification=2] on ($(FigA)+( +0.103, -0.12)$) 
    in node[largewindow_w,anchor=east]       at ($(FigA.north) + (+0.035,-0.128)$);
\end{tikzpicture}
\hspace{-1.0mm}
\begin{tikzpicture}[x=6cm, y=6cm, spy using outlines={every spy on node/.append style={smallwindow_w}}]
\node[anchor=south] (FigA) at (0,0) {\includegraphics[trim=0 0 0 0 ,clip,width=0.88in]{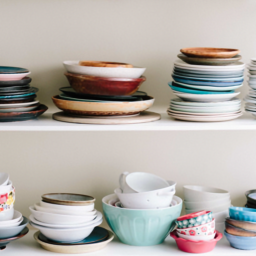}};
\spy [closeup_w_3,magnification=2] on ($(FigA)+( +0.103, -0.12)$) 
    in node[largewindow_w,anchor=east]       at ($(FigA.north) + (+0.035,-0.128)$);
\end{tikzpicture}
\hspace{-1.0mm}
\begin{tikzpicture}[x=6cm, y=6cm, spy using outlines={every spy on node/.append style={smallwindow_w}}]
\node[anchor=south] (FigA) at (0,0) {\includegraphics[trim=0 0 0 0 ,clip,width=0.88in]{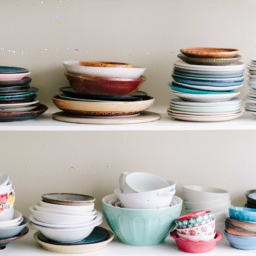}};
\spy [closeup_w_3,magnification=2] on ($(FigA)+( +0.103, -0.12)$) 
    in node[largewindow_w,anchor=east]       at ($(FigA.north) + (+0.035,-0.128)$);
\end{tikzpicture}
\hspace{-1.0mm}
\begin{tikzpicture}[x=6cm, y=6cm, spy using outlines={every spy on node/.append style={smallwindow_w}}]
\node[anchor=south] (FigA) at (0,0) {\includegraphics[trim=0 0 0 0 ,clip,width=0.88in]{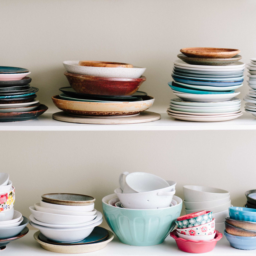}};
\spy [closeup_w_3,magnification=2] on ($(FigA)+( +0.103, -0.12)$) 
    in node[largewindow_w,anchor=east]       at ($(FigA.north) + (+0.035,-0.128)$);
\end{tikzpicture}

\vspace{-5pt}

\hspace{-1mm}
\raisebox{0.3in}{\rotatebox{90}{Kodak \cite{Kodakdataset}}}
\hspace{-2mm}
\stackunder[2pt]{
\begin{tikzpicture}[x=6cm, y=6cm, spy using outlines={every spy on node/.append style={smallwindow_w}}]
\node[anchor=south] (FigA) at (0,0) {\includegraphics[trim=0 0 0 0 ,clip,width=0.88in]{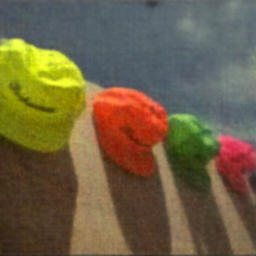}};
\spy [closeup_w_3,magnification=2] on ($(FigA)+( -0.12, +0.025)$) 
    in node[largewindow_w,anchor=east]       at ($(FigA.north) + (0.188,-0.278)$);
\end{tikzpicture}
}{ReLU +P.E \cite{mildenhall2020nerf}}
\hspace{-2.5mm}
\stackunder[2pt]{
\begin{tikzpicture}[x=6cm, y=6cm, spy using outlines={every spy on node/.append style={smallwindow_w}}]
\node[anchor=south] (FigA) at (0,0) {\includegraphics[trim=0 0 0 0 ,clip,width=0.88in]{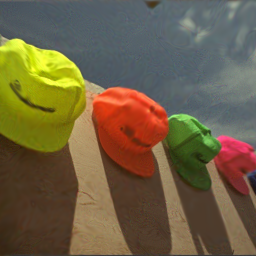}};
\spy [closeup_w_3,magnification=2] on ($(FigA)+( -0.12, +0.025)$) 
    in node[largewindow_w,anchor=east]       at ($(FigA.north) + (0.188,-0.278)$);
\end{tikzpicture}
}{WIRE \cite{wirewavelet2023}}
\hspace{-2.5mm}
\stackunder[2pt]{
\begin{tikzpicture}[x=6cm, y=6cm, spy using outlines={every spy on node/.append style={smallwindow_w}}]
\node[anchor=south] (FigA) at (0,0) {\includegraphics[trim=0 0 0 0 ,clip,width=0.88in]{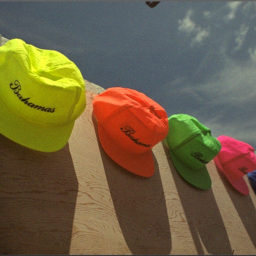}};
\spy [closeup_w_3,magnification=2] on ($(FigA)+( -0.12, +0.025)$) 
    in node[largewindow_w,anchor=east]       at ($(FigA.north) + (0.188,-0.278)$);
\end{tikzpicture}
}{Gauss \cite{activationinr_gauss_etc}}
\hspace{-2.5mm}
\stackunder[2pt]{
\begin{tikzpicture}[x=6cm, y=6cm, spy using outlines={every spy on node/.append style={smallwindow_w}}]
\node[anchor=south] (FigA) at (0,0) {\includegraphics[trim=0 0 0 0 ,clip,width=0.88in]{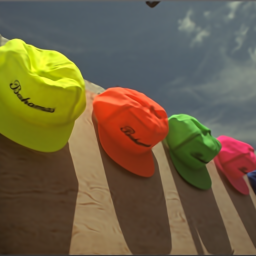}};
\spy [closeup_w_3,magnification=2] on ($(FigA)+( -0.12, +0.025)$) 
    in node[largewindow_w,anchor=east]       at ($(FigA.north) + (0.188,-0.278)$);
\end{tikzpicture}
}{{SIREN} \cite{sitzmann2019siren}}
\hspace{-2.5mm}
\stackunder[2pt]{
\begin{tikzpicture}[x=6cm, y=6cm, spy using outlines={every spy on node/.append style={smallwindow_w}}]
\node[anchor=south] (FigA) at (0,0) {\includegraphics[trim=0 0 0 0 ,clip,width=0.88in]{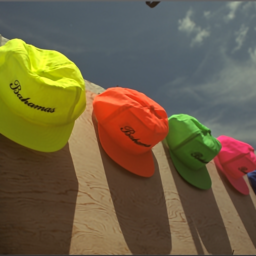}};
\spy [closeup_w_3,magnification=2] on ($(FigA)+( -0.12, +0.025)$) 
    in node[largewindow_w,anchor=east]       at ($(FigA.north) + (0.188,-0.278)$);
\end{tikzpicture}
}{{FINER} \cite{liu2023finer}}
\hspace{-2.5mm}
\stackunder[2pt]{
\begin{tikzpicture}[x=6cm, y=6cm, spy using outlines={every spy on node/.append style={smallwindow_w}}]
\node[anchor=south] (FigA) at (0,0) {\includegraphics[trim=0 0 0 0 ,clip,width=0.88in]{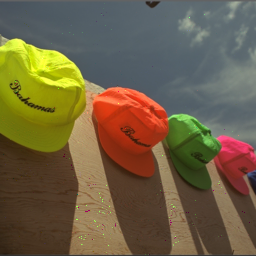}};
\spy [closeup_w_3,magnification=2] on ($(FigA)+( -0.12, +0.025)$) 
    in node[largewindow_w,anchor=east]       at ($(FigA.north) + (0.188,-0.278)$);
\end{tikzpicture}
}{{Ours}}
\hspace{-2.5mm}
\stackunder[2pt]{
\begin{tikzpicture}[x=6cm, y=6cm, spy using outlines={every spy on node/.append style={smallwindow_w}}]
\node[anchor=south] (FigA) at (0,0) {\includegraphics[trim=0 0 0 0 ,clip,width=0.88in]{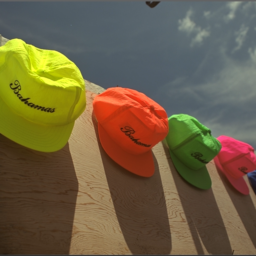}};
\spy [closeup_w_3,magnification=2] on ($(FigA)+( -0.12, +0.025)$) 
    in node[largewindow_w,anchor=east]       at ($(FigA.north) + (0.188,-0.278)$);
\end{tikzpicture}
}{{GT}}

\vspace*{-6pt}
\caption{Qualitative comparison of under-fitted images (\# of Iterations : 400) with existing methods.}
\vspace*{-12pt}
\label{fig:qual_convergence}
\end{figure*}


\begin{figure*}[!ht]
\footnotesize
\centering
\raisebox{0.3in}{\rotatebox{90}{Iteration$\rightarrow$}}
\stackunder[2pt]{\includegraphics[trim=0 0 0 0,height = 1.06in,width = 1.06in]{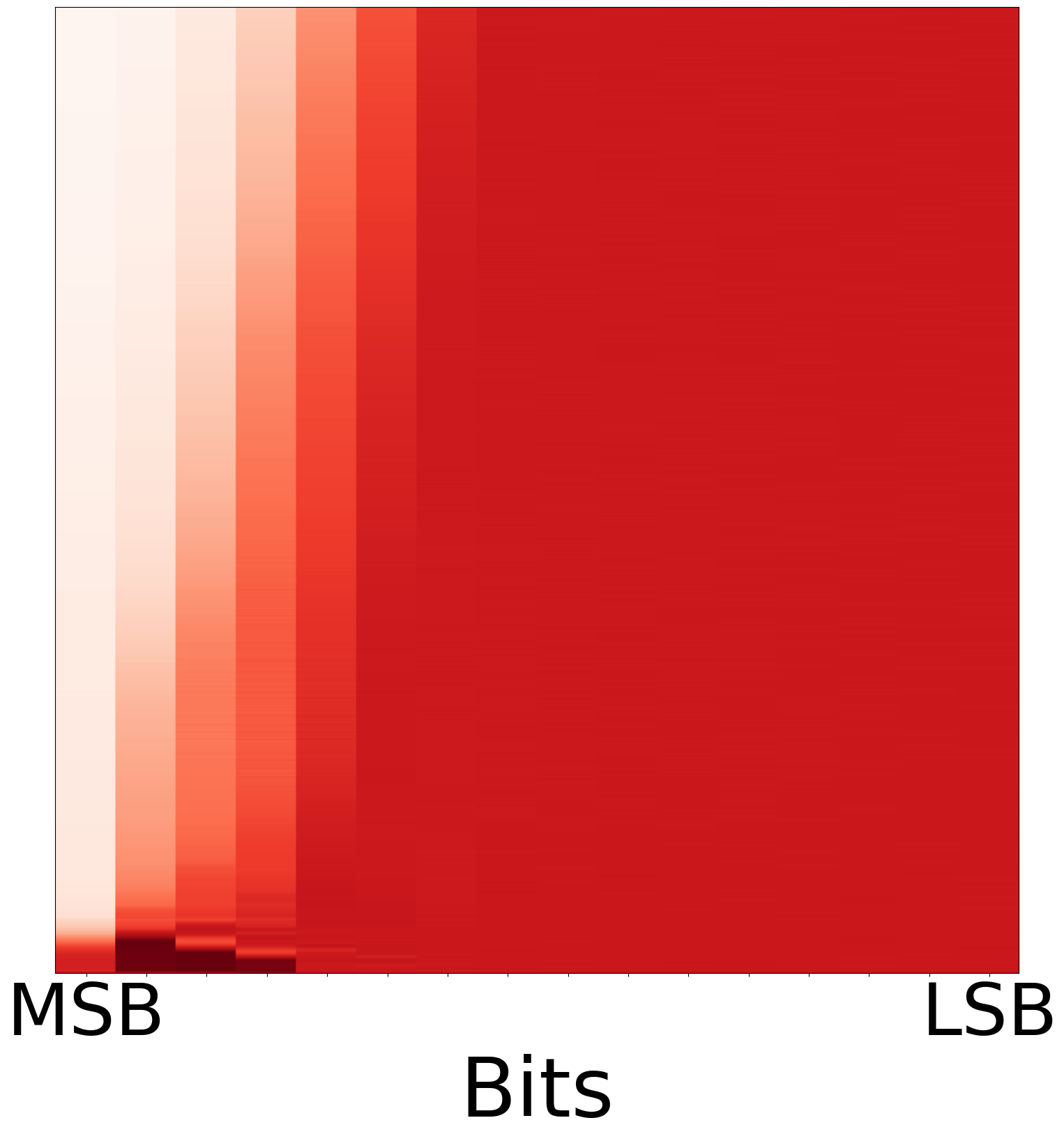}}{ReLU$+$P.E \cite{mildenhall2020nerf}}
\stackunder[2pt]{\includegraphics[trim=0 0 0 0,clip,height = 1.06in,width = 1.06in]{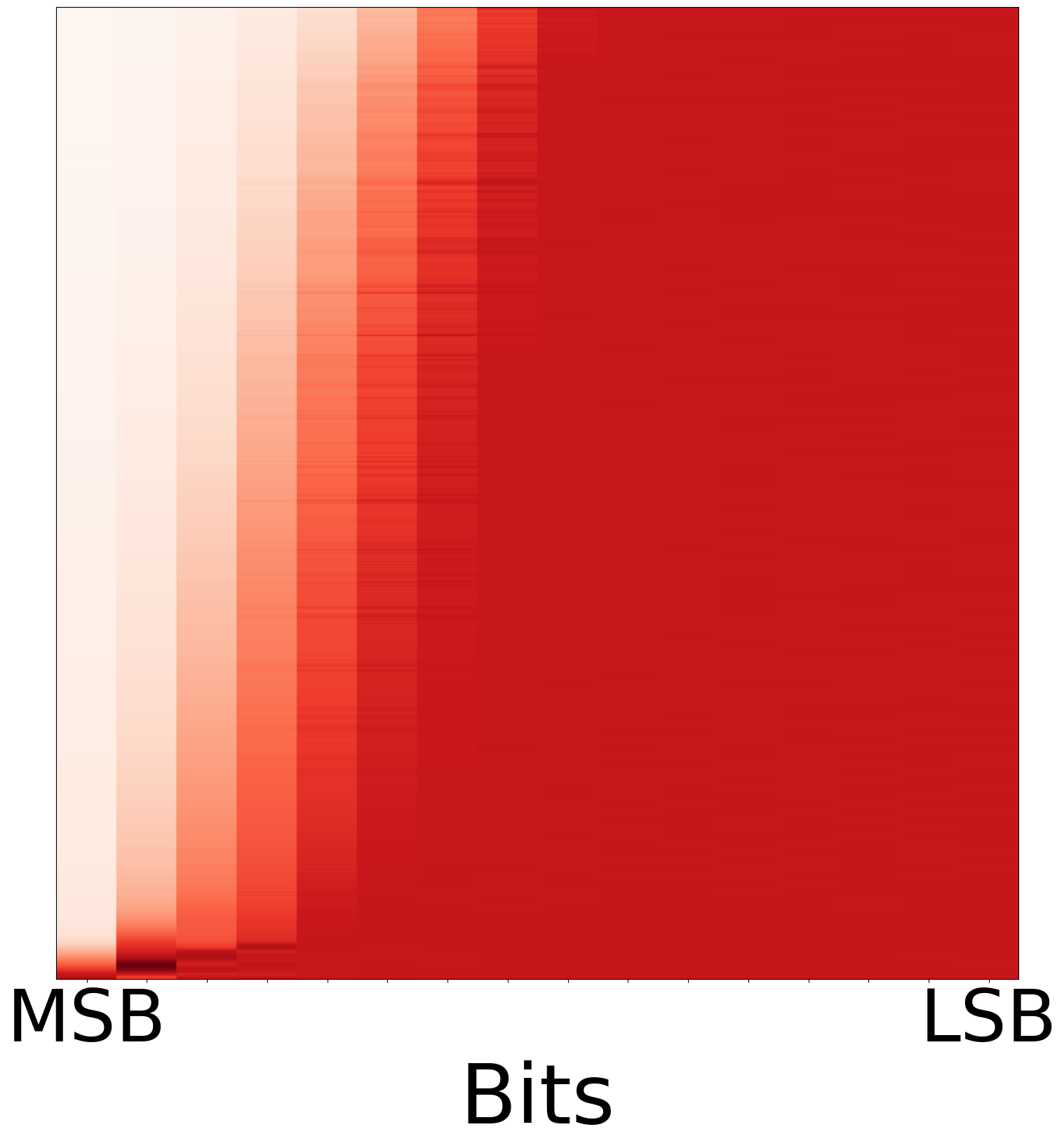}}{WIRE \cite{wirewavelet2023}}
\stackunder[2pt]{\includegraphics[trim=0 0 0 0,clip,height = 1.06in,width = 1.06in]{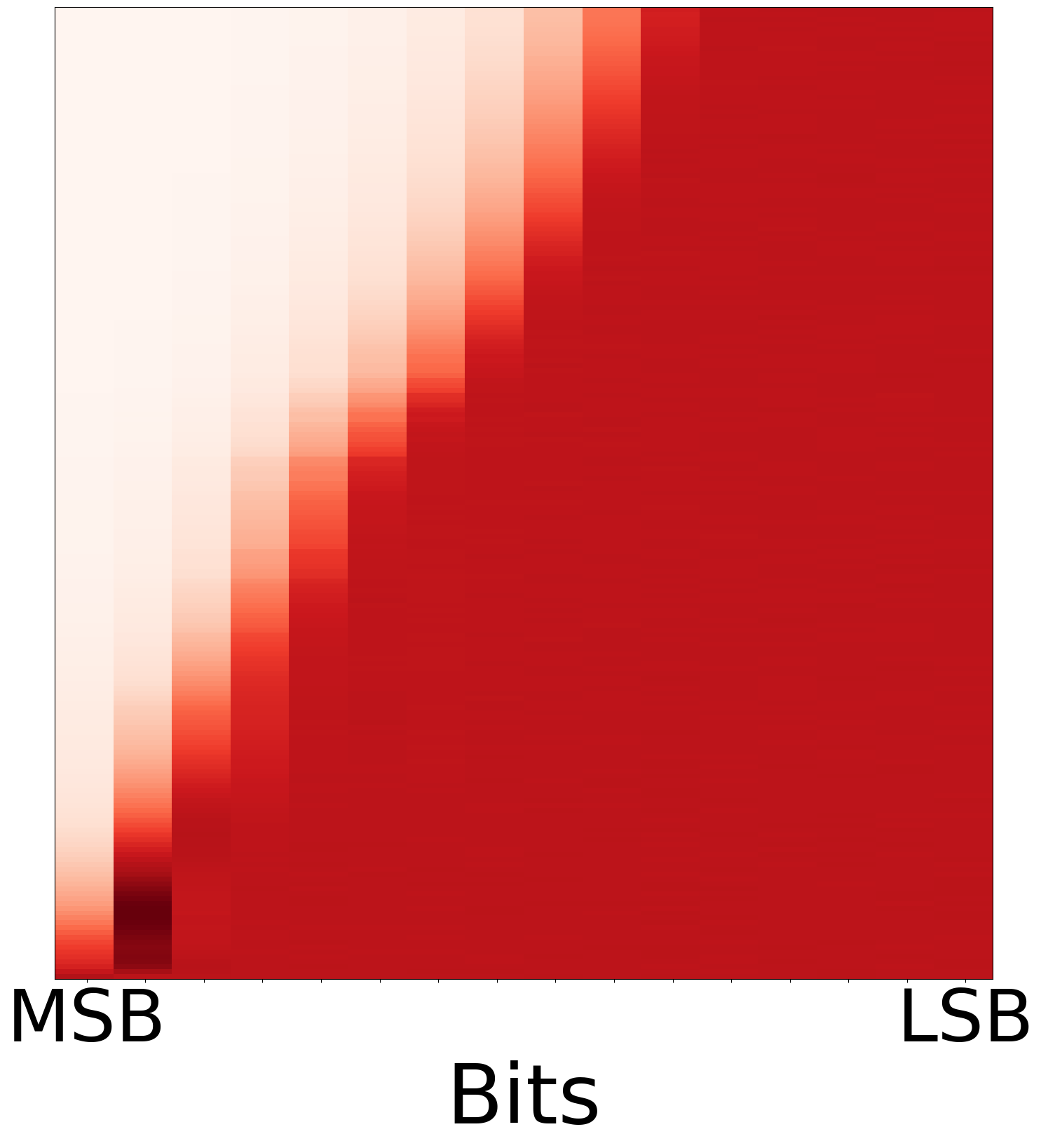}}{Gauss \cite{activationinr_gauss_etc}}
\stackunder[2pt]{\includegraphics[trim=0 0 0 0,clip,height = 1.06in,width = 1.06in]{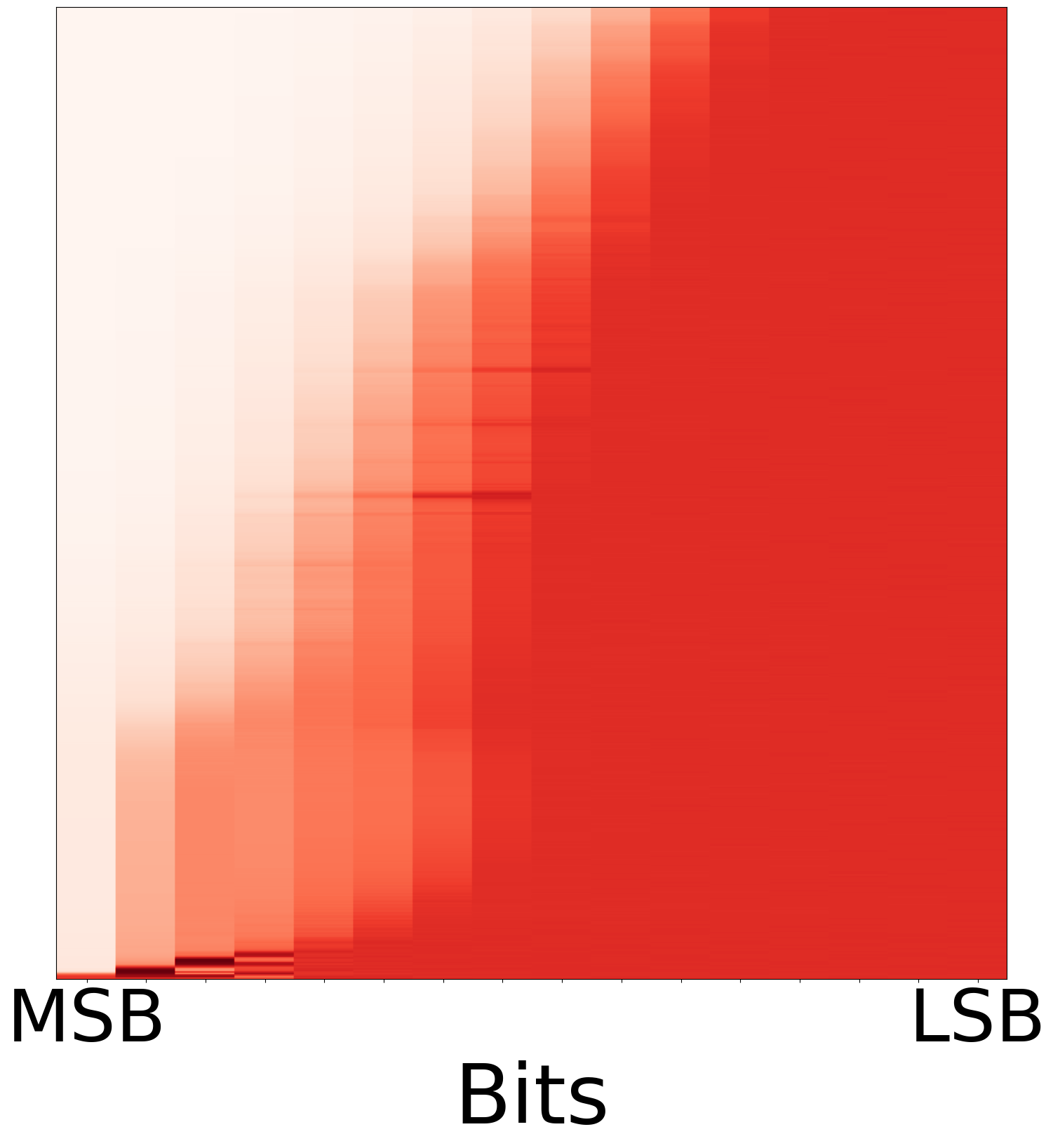}}{SIREN \cite{sitzmann2019siren}}
\stackunder[2pt]{\includegraphics[trim=0 0 0 0,clip,height = 1.06in,width = 1.06in]{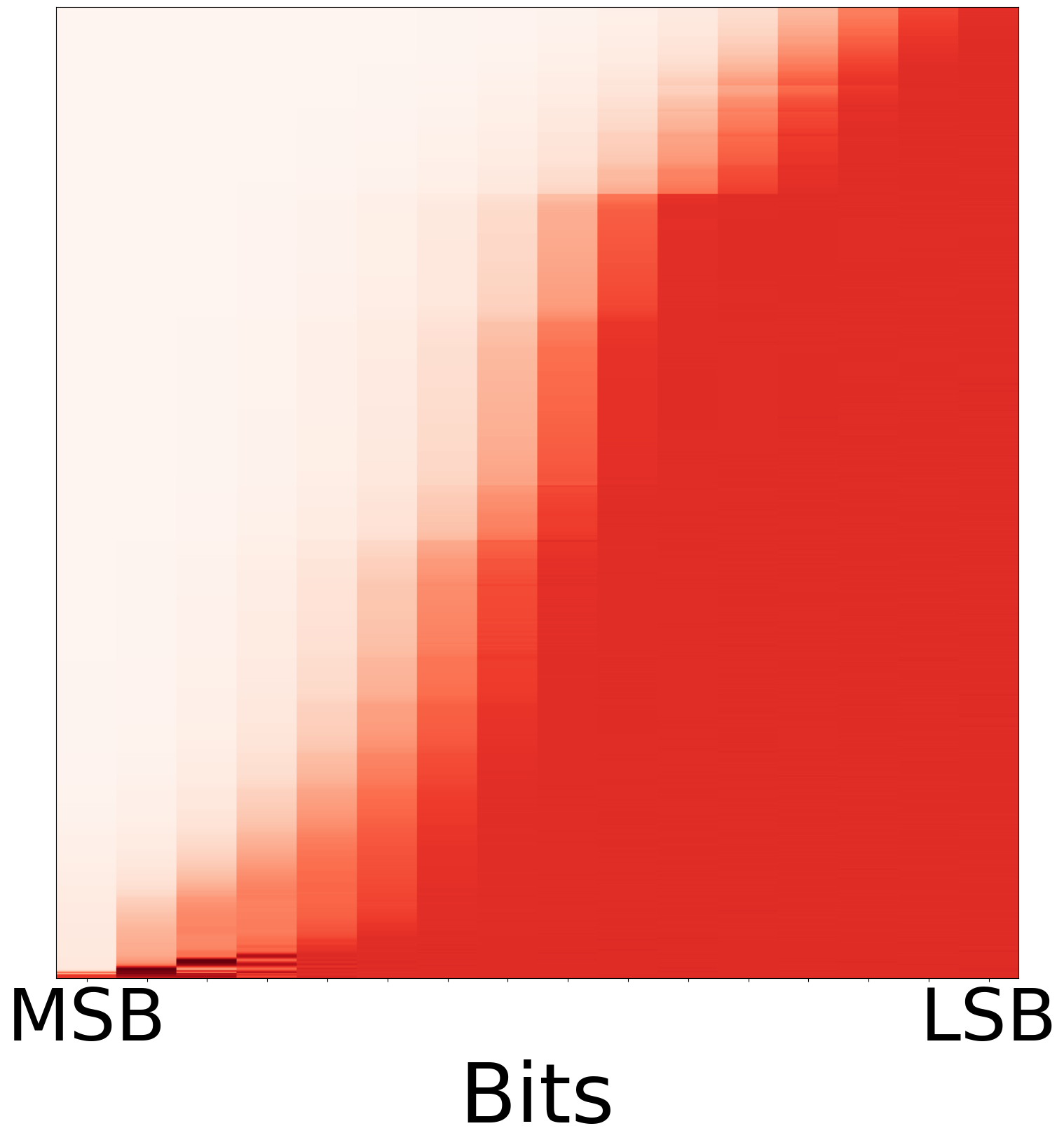}}{FINER \cite{liu2023finer}}
\stackunder[2pt]{\includegraphics[trim=0 0 0 0,clip,height = 1.06in,width = 1.06in]{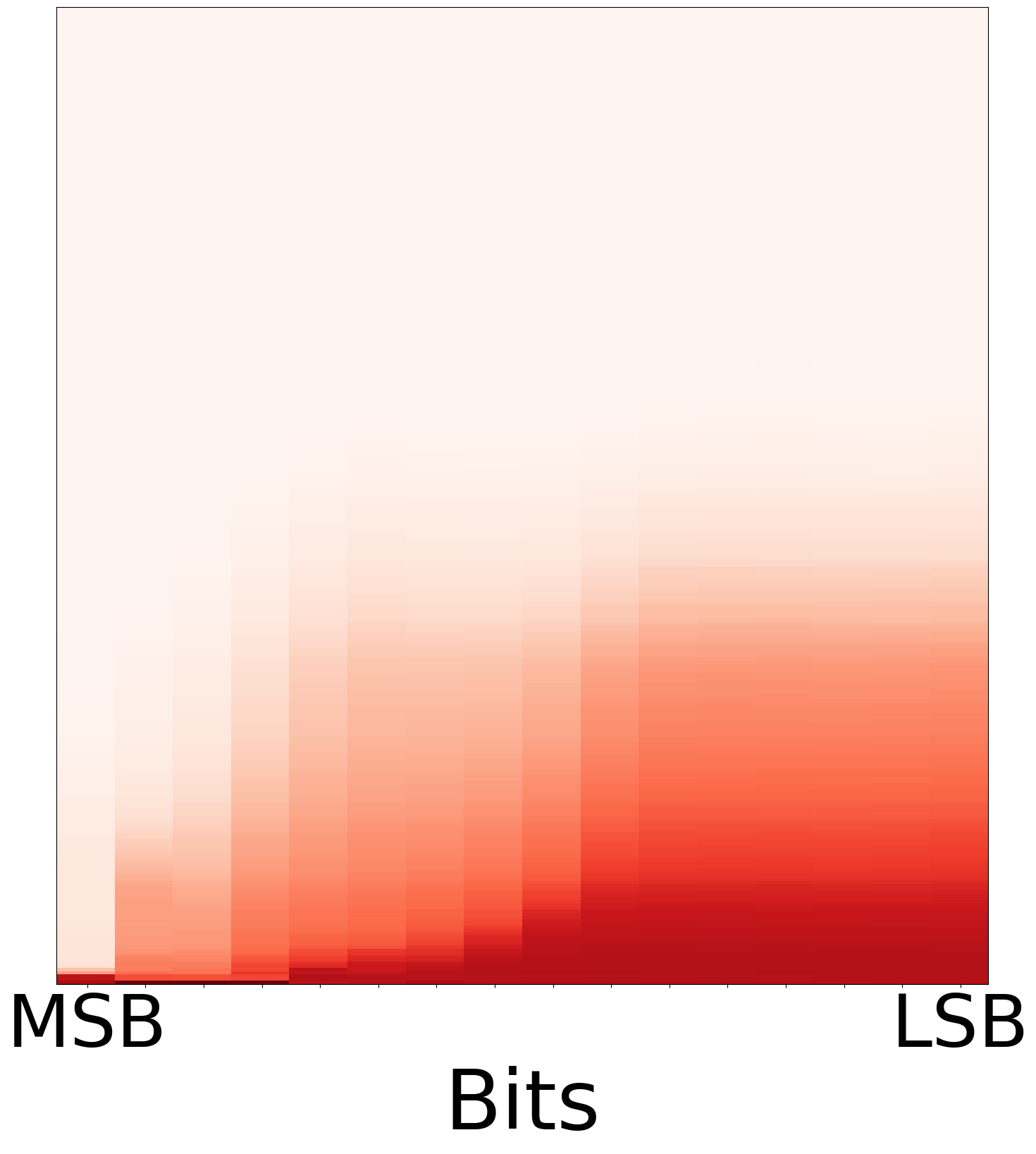}}{Ours}
\includegraphics[trim=0 0 0 0,clip,height = 1.06in]{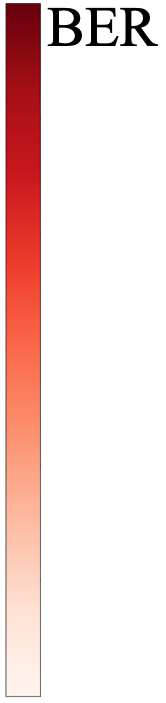}
\vspace{-5pt}
\caption{\textbf{Bit-Error-Rate} of each bit-plane on a TESTIMAGE \cite{asuni2013testimages}. The X-axis is for bit depth (MSB to LSB), and the Y-axis is for iteration.}
\vspace*{-15pt}
\label{fig:bit-bias}
\end{figure*}

\begin{figure*}
\footnotesize
\centering

\raisebox{0.2in}{\rotatebox{90}{Image :$\mathbf{I_\theta}$}}
\begin{tikzpicture}[x=6cm, y=6cm, spy using outlines={every spy on node/.append style={smallwindow_r}}]
\node[anchor=south] (FigA) at (0,0) {\includegraphics[trim=0 0 0
0 ,clip,height=0.88in]{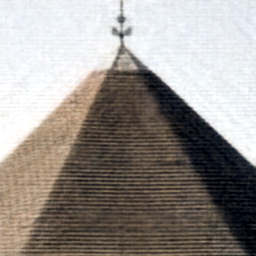}};
\spy [closeup_w_4,magnification=3] on ($(FigA)+( +0.155, +0.12)$) 
    in node[largewindow_r,anchor=east]       at ($(FigA.north) + (-0.016,-0.105)$);
\spy [closeup_w_4,magnification=3] on ($(FigA)+( +0.125, -0.03)$) 
    in node[largewindow_r,anchor=east]       at ($(FigA.north) + (-0.016,-0.305)$);
\end{tikzpicture}
\hspace{-2.5mm}
\begin{tikzpicture}[x=6cm, y=6cm, spy using outlines={every spy on node/.append style={smallwindow_r}}]
\node[anchor=south] (FigA) at (0,0) {\includegraphics[trim=0 0 0
0 ,clip,height=0.88in]{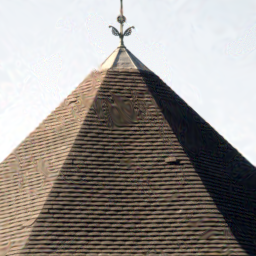}};
\spy [closeup_w_4,magnification=3] on ($(FigA)+( +0.155, +0.12)$) 
    in node[largewindow_r,anchor=east]       at ($(FigA.north) + (-0.016,-0.105)$);;
\spy [closeup_w_4,magnification=3] on ($(FigA)+( +0.125, -0.03)$) 
    in node[largewindow_r,anchor=east]       at ($(FigA.north) + (-0.016,-0.305)$);
\end{tikzpicture}
\hspace{-2.5mm}
\begin{tikzpicture}[x=6cm, y=6cm, spy using outlines={every spy on node/.append style={smallwindow_r}}]
\node[anchor=south] (FigA) at (0,0) {\includegraphics[trim=0 0 0
0 ,clip,height=0.88in]{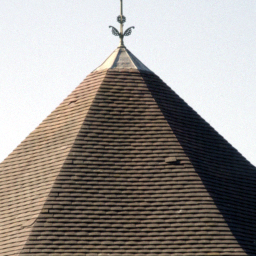}};
\spy [closeup_w_4,magnification=3] on ($(FigA)+( +0.155, +0.12)$) 
    in node[largewindow_r,anchor=east]       at ($(FigA.north) + (-0.016,-0.105)$);
\spy [closeup_w_4,magnification=3] on ($(FigA)+( +0.125, -0.03)$) 
    in node[largewindow_r,anchor=east]       at ($(FigA.north) + (-0.016,-0.305)$);
\end{tikzpicture}
\hspace{-2.5mm}
\begin{tikzpicture}[x=6cm, y=6cm, spy using outlines={every spy on node/.append style={smallwindow_r}}]
\node[anchor=south] (FigA) at (0,0) {\includegraphics[trim=0 0 0
0 ,clip,height=0.88in]{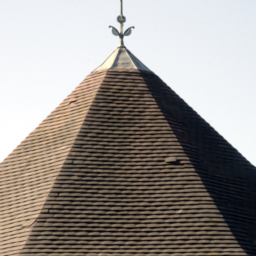}};
\spy [closeup_w_4,magnification=3] on ($(FigA)+( +0.155, +0.12)$) 
    in node[largewindow_r,anchor=east]       at ($(FigA.north) + (-0.016,-0.105)$);
\spy [closeup_w_4,magnification=3] on ($(FigA)+( +0.125, -0.03)$) 
    in node[largewindow_r,anchor=east]       at ($(FigA.north) + (-0.016,-0.305)$);
\end{tikzpicture}
\hspace{-2.5mm}
\begin{tikzpicture}[x=6cm, y=6cm, spy using outlines={every spy on node/.append style={smallwindow_r}}]
\node[anchor=south] (FigA) at (0,0) {\includegraphics[trim=0 0 0
0 ,clip,height=0.88in]{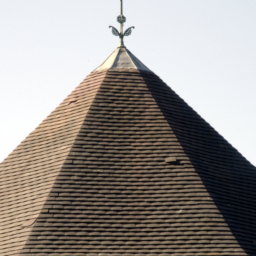}};
\spy [closeup_w_4,magnification=3] on ($(FigA)+( +0.155, +0.12)$) 
    in node[largewindow_r,anchor=east]       at ($(FigA.north) + (-0.016,-0.105)$);
\spy [closeup_w_4,magnification=3] on ($(FigA)+( +0.125, -0.03)$) 
    in node[largewindow_r,anchor=east]       at ($(FigA.north) + (-0.016,-0.305)$);
\end{tikzpicture}
\hspace{-2.5mm}
\begin{tikzpicture}[x=6cm, y=6cm, spy using outlines={every spy on node/.append style={smallwindow_r}}]
\node[anchor=south] (FigA) at (0,0) {\includegraphics[trim=0 0 0
0 ,clip,height=0.88in]{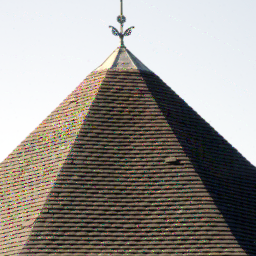}};
\spy [closeup_w_4,magnification=3] on ($(FigA)+( +0.155, +0.12)$) 
    in node[largewindow_r,anchor=east]       at ($(FigA.north) + (-0.016,-0.105)$);
\spy [closeup_w_4,magnification=3] on ($(FigA)+( +0.125, -0.03)$) 
    in node[largewindow_r,anchor=east]       at ($(FigA.north) + (-0.016,-0.305)$);
\end{tikzpicture}
\hspace{-2.5mm}
\begin{tikzpicture}[x=6cm, y=6cm, spy using outlines={every spy on node/.append style={smallwindow_r}}]
\node[anchor=south] (FigA) at (0,0) {\includegraphics[trim=0 0 0
0 ,clip,height=0.88in]{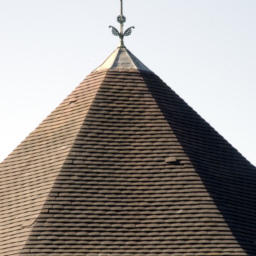}};
\spy [closeup_w_4,magnification=3] on ($(FigA)+( +0.155, +0.12)$) 
    in node[largewindow_r,anchor=east]       at ($(FigA.north) + (-0.016,-0.105)$);
\spy [closeup_w_4,magnification=3] on ($(FigA)+( +0.125, -0.03)$) 
    in node[largewindow_r,anchor=east]       at ($(FigA.north) + (-0.016,-0.305)$);
\end{tikzpicture}
\hspace{-2.5mm}

\raisebox{0.2in}{\rotatebox{90}{MSB-1 :$\mathbf{B}^{(15)}$}}
\begin{tikzpicture}[x=6cm, y=6cm, spy using outlines={every spy on node/.append style={smallwindow_r}}]
\node[anchor=south] (FigA) at (0,0) {\includegraphics[trim=0 0 0
0 ,clip,height=0.88in]{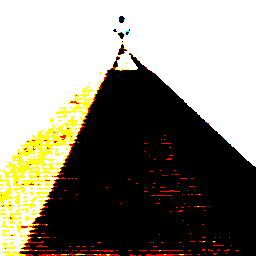}};
\end{tikzpicture}
\hspace{-2.5mm}
\begin{tikzpicture}[x=6cm, y=6cm, spy using outlines={every spy on node/.append style={smallwindow_r}}]
\node[anchor=south] (FigA) at (0,0) {\includegraphics[trim=0 0 0
0 ,clip,height=0.88in]{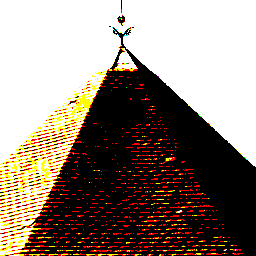}};
\end{tikzpicture}
\hspace{-2.5mm}
\begin{tikzpicture}[x=6cm, y=6cm, spy using outlines={every spy on node/.append style={smallwindow_r}}]
\node[anchor=south] (FigA) at (0,0) {\includegraphics[trim=0 0 0
0 ,clip,height=0.88in]{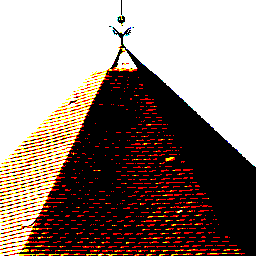}};
\end{tikzpicture}
\hspace{-2.5mm}
\begin{tikzpicture}[x=6cm, y=6cm, spy using outlines={every spy on node/.append style={smallwindow_r}}]
\node[anchor=south] (FigA) at (0,0) {\includegraphics[trim=0 0 0
0 ,clip,height=0.88in]{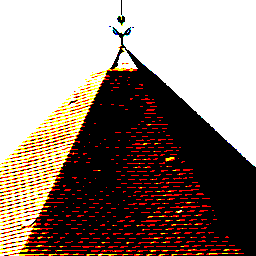}};
\end{tikzpicture}
\hspace{-2.5mm}
\begin{tikzpicture}[x=6cm, y=6cm, spy using outlines={every spy on node/.append style={smallwindow_r}}]
\node[anchor=south] (FigA) at (0,0) {\includegraphics[trim=0 0 0
0 ,clip,height=0.88in]{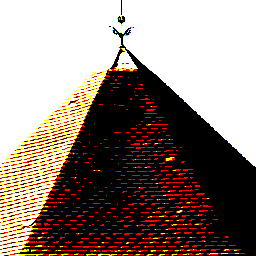}};
\end{tikzpicture}
\hspace{-2.5mm}
\begin{tikzpicture}[x=6cm, y=6cm, spy using outlines={every spy on node/.append style={smallwindow_r}}]
\node[anchor=south] (FigA) at (0,0) {\includegraphics[trim=0 0 0
0 ,clip,height=0.88in]{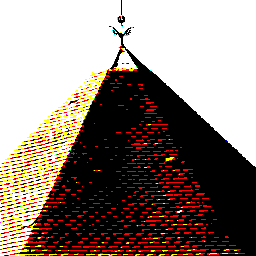}};
\end{tikzpicture}
\hspace{-2.5mm}
\begin{tikzpicture}[x=6cm, y=6cm, spy using outlines={every spy on node/.append style={smallwindow_r}}]
\node[anchor=south] (FigA) at (0,0) {\includegraphics[trim=0 0 0
0 ,clip,height=0.88in]{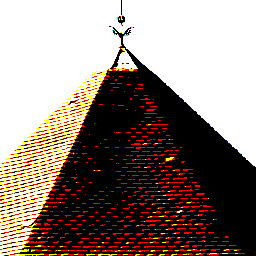}};
\end{tikzpicture}
\hspace{-2.5mm}

\vspace{-2mm}

\raisebox{0.2in}{\rotatebox{90}{MSB-4 :$\mathbf{B}^{(12)}$}}
\begin{tikzpicture}[x=6cm, y=6cm, spy using outlines={every spy on node/.append style={smallwindow_r}}]
\node[anchor=south] (FigA) at (0,0) {\includegraphics[trim=0 0 0
0 ,clip,height=0.88in]{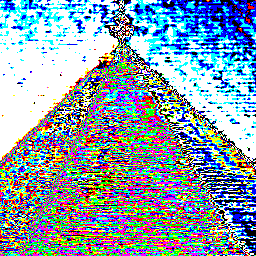}};
\spy [closeup_w_4,magnification=3] on ($(FigA)+( +0.125, -0.03)$) 
    in node[largewindow_r,anchor=east]       at ($(FigA.north) + (-0.016,-0.305)$);
\end{tikzpicture}
\hspace{-2.5mm}
\begin{tikzpicture}[x=6cm, y=6cm, spy using outlines={every spy on node/.append style={smallwindow_r}}]
\node[anchor=south] (FigA) at (0,0) {\includegraphics[trim=0 0 0
0 ,clip,height=0.88in]{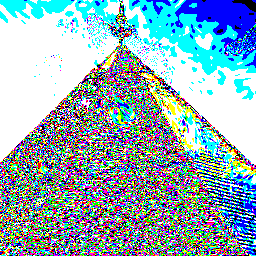}};
\spy [closeup_w_4,magnification=3] on ($(FigA)+( +0.125, -0.03)$) 
    in node[largewindow_r,anchor=east]       at ($(FigA.north) + (-0.016,-0.305)$);
\end{tikzpicture}
\hspace{-2.5mm}
\begin{tikzpicture}[x=6cm, y=6cm, spy using outlines={every spy on node/.append style={smallwindow_r}}]
\node[anchor=south] (FigA) at (0,0) {\includegraphics[trim=0 0 0
0 ,clip,height=0.88in]{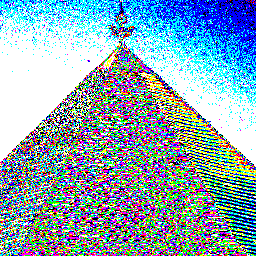}};
\spy [closeup_w_4,magnification=3] on ($(FigA)+( +0.125, -0.03)$) 
    in node[largewindow_r,anchor=east]       at ($(FigA.north) + (-0.016,-0.305)$);
\end{tikzpicture}
\hspace{-2.5mm}
\begin{tikzpicture}[x=6cm, y=6cm, spy using outlines={every spy on node/.append style={smallwindow_r}}]
\node[anchor=south] (FigA) at (0,0) {\includegraphics[trim=0 0 0
0 ,clip,height=0.88in]{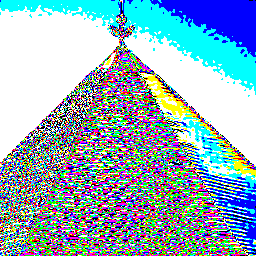}};
\spy [closeup_w_4,magnification=3] on ($(FigA)+( +0.125, -0.03)$) 
    in node[largewindow_r,anchor=east]       at ($(FigA.north) + (-0.016,-0.305)$);
\end{tikzpicture}
\hspace{-2.5mm}
\begin{tikzpicture}[x=6cm, y=6cm, spy using outlines={every spy on node/.append style={smallwindow_r}}]
\node[anchor=south] (FigA) at (0,0) {\includegraphics[trim=0 0 0
0 ,clip,height=0.88in]{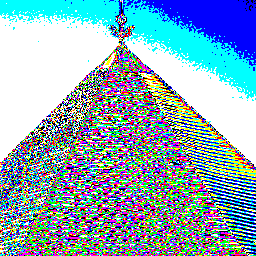}};
\spy [closeup_w_4,magnification=3] on ($(FigA)+( +0.125, -0.03)$) 
    in node[largewindow_r,anchor=east]       at ($(FigA.north) + (-0.016,-0.305)$);
\end{tikzpicture}
\hspace{-2.5mm}
\begin{tikzpicture}[x=6cm, y=6cm, spy using outlines={every spy on node/.append style={smallwindow_r}}]
\node[anchor=south] (FigA) at (0,0) {\includegraphics[trim=0 0 0
0 ,clip,height=0.88in]{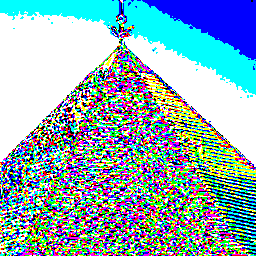}};
\spy [closeup_w_4,magnification=3] on ($(FigA)+( +0.125, -0.03)$) 
    in node[largewindow_r,anchor=east]       at ($(FigA.north) + (-0.016,-0.305)$);
\end{tikzpicture}
\hspace{-2.5mm}
\begin{tikzpicture}[x=6cm, y=6cm, spy using outlines={every spy on node/.append style={smallwindow_r}}]
\node[anchor=south] (FigA) at (0,0) {\includegraphics[trim=0 0 0
0 ,clip,height=0.88in]{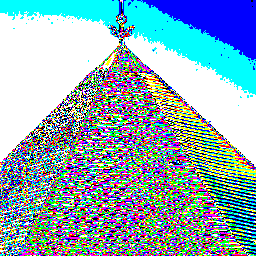}};
\spy [closeup_w_4,magnification=3] on ($(FigA)+( +0.125, -0.03)$) 
    in node[largewindow_r,anchor=east]       at ($(FigA.north) + (-0.016,-0.305)$);
\end{tikzpicture}
\hspace{-2.5mm}

\vspace{-2mm}

\raisebox{0.2in}{\rotatebox{90}{MSB-9 :$\mathbf{B}^{(7)}$}}
\stackunder[2pt]{\begin{tikzpicture}[x=6cm, y=6cm, spy using outlines={every spy on node/.append style={smallwindow_r}}]
\node[anchor=south] (FigA) at (0,0) {\includegraphics[trim=0 0 0
0 ,clip,height=0.88in]{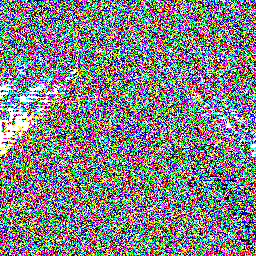}};
\spy [closeup_w_4,magnification=3] on ($(FigA)+( +0.155, +0.12)$) 
    in node[largewindow_r,anchor=east]       at ($(FigA.north) + (-0.016,-0.305)$);
\end{tikzpicture}}{ReLU+P.E \cite{mildenhall2020nerf}}
\hspace{-2.5mm}
\stackunder[2pt]{\begin{tikzpicture}[x=6cm, y=6cm, spy using outlines={every spy on node/.append style={smallwindow_r}}]
\node[anchor=south] (FigA) at (0,0) {\includegraphics[trim=0 0 0
0 ,clip,height=0.88in]{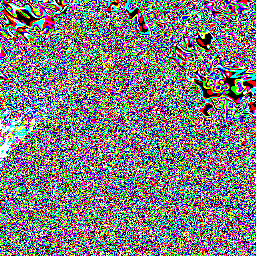}};
\spy [closeup_w_4,magnification=3] on ($(FigA)+( +0.155, +0.12)$) 
    in node[largewindow_r,anchor=east]       at ($(FigA.north) + (-0.016,-0.305)$);
\end{tikzpicture}}{WIRE \cite{wirewavelet2023}}
\hspace{-2.5mm}
\stackunder[2pt]{\begin{tikzpicture}[x=6cm, y=6cm, spy using outlines={every spy on node/.append style={smallwindow_r}}]
\node[anchor=south] (FigA) at (0,0) {\includegraphics[trim=0 0 0
0 ,clip,height=0.88in]{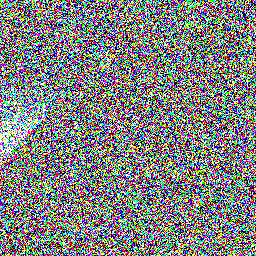}};
\spy [closeup_w_4,magnification=3] on ($(FigA)+( +0.155, +0.12)$) 
    in node[largewindow_r,anchor=east]       at ($(FigA.north) + (-0.016,-0.305)$);;
\end{tikzpicture}}{Gauss \cite{activationinr_gauss_etc}}
\hspace{-2.5mm}
\stackunder[2pt]{\begin{tikzpicture}[x=6cm, y=6cm, spy using outlines={every spy on node/.append style={smallwindow_r}}]
\node[anchor=south] (FigA) at (0,0) {\includegraphics[trim=0 0 0
0 ,clip,height=0.88in]{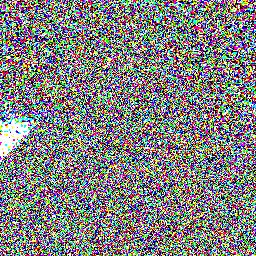}};
\spy [closeup_w_4,magnification=3] on ($(FigA)+( +0.155, +0.12)$) 
    in node[largewindow_r,anchor=east]       at ($(FigA.north) + (-0.016,-0.305)$);
\end{tikzpicture}}{SIREN \cite{sitzmann2019siren}}
\hspace{-2.5mm}
\stackunder[2pt]{\begin{tikzpicture}[x=6cm, y=6cm, spy using outlines={every spy on node/.append style={smallwindow_r}}]
\node[anchor=south] (FigA) at (0,0) {\includegraphics[trim=0 0 0
0 ,clip,height=0.88in]{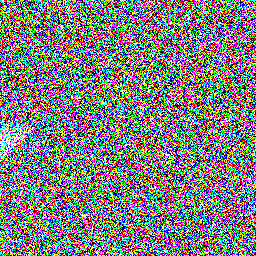}};
\spy [closeup_w_4,magnification=3] on ($(FigA)+( +0.155, +0.12)$) 
    in node[largewindow_r,anchor=east]       at ($(FigA.north) + (-0.016,-0.305)$);
\end{tikzpicture}}{FINER \cite{liu2023finer}}
\hspace{-2.5mm}
\stackunder[2pt]{\begin{tikzpicture}[x=6cm, y=6cm, spy using outlines={every spy on node/.append style={smallwindow_r}}]
\node[anchor=south] (FigA) at (0,0) {\includegraphics[trim=0 0 0
0 ,clip,height=0.88in]{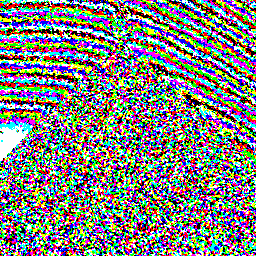}};
\spy [closeup_w_4,magnification=3] on ($(FigA)+( +0.155, +0.12)$) 
    in node[largewindow_r,anchor=east]       at ($(FigA.north) + (-0.016,-0.305)$);
\end{tikzpicture}}{Ours}
\hspace{-2.5mm}
\stackunder[2pt]{\begin{tikzpicture}[x=6cm, y=6cm, spy using outlines={every spy on node/.append style={smallwindow_r}}]
\node[anchor=south] (FigA) at (0,0) {\includegraphics[trim=0 0 0
0 ,clip,height=0.88in]{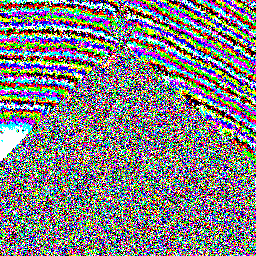}};
\spy [closeup_w_4,magnification=3] on ($(FigA)+( +0.155, +0.12)$) 
    in node[largewindow_r,anchor=east]       at ($(FigA.north) + (-0.016,-0.305)$);
\end{tikzpicture}}{GT}
\hspace{-2.5mm}

\vspace*{-6pt}
\caption{Qualitative comparison of an under-fitted image (\# of Iterations : 400) and its bit-plane. The experiment was conducted on the 16-bit image of TESTIMAGE \cite{asuni2013testimages}. MSB-$n$ indicates $n$th bit-plane from the MSB. }
\vspace*{-12pt}
\label{fig:bit-bias_qual}
\end{figure*}

\begin{table}[ht]
\footnotesize
\centering

\begin{tabular}{l|cc|cc}

\hline
\multirow{2}{*}{ }&  \multicolumn{2}{c|}{Kodak \cite{Kodakdataset}} & \multicolumn{2}{c}{TESTIMAGES \cite{asuni2013testimages}}\\
\cline{2-5}
 & \#Iter.($\downarrow$) & PSNR ($\uparrow$)& \#Iter.($\downarrow$) & PSNR ($\uparrow$)\\
 \hline 
Instant-NGP \cite{muller2022instantngp}&  2000   &  52.82      &  5000   &     54.92   \\
Instant-NGP + \textbf{Ours}& 1130    &      $\infty$      &   4668  &     $\infty$   \\
\hline
DINER \cite{xie2023diner}&  5000   &  39.59      &  5000   &     38.30   \\
DINER + \textbf{Ours}& 3347    &      $\infty$      &   3915  &     $\infty$   \\
\hline
 \hline 
Gauss \cite{activationinr_gauss_etc}&  15000   &  100.48      &  50000   &     74.88   \\
Gauss + \textbf{Ours}& 7931    &      $\infty$      &   29546  &     $\infty$   \\
\hline
FINER \cite{liu2023finer} &  500   &  48.52      &  2000   &     56.58   \\
FINER + \textbf{Ours} & 428    &      $\infty$      &   1464  &     $\infty$   \\
\hline
\end{tabular}
\vspace{-7pt}
\caption{Quantitative comparison results combining existing methods with ours. Coordinate encoding method (top) and activation modification method (bottom).}
\vspace{-20pt}
\label{tab:combination}
\end{table}

\section{Experiments}
\subsection{Implementation Details}\label{sec:details}
\noindent To validate our proposed method, we conduct experiments on MIT-fiveK \cite{fivek} and TESTIMAGES1200 \cite{asuni2013testimages} dataset that require high dynamic range (0-65,535). The TESTIMAGES dataset includes 40 natural images. We select the last 1,000 images (with indices from 4,001 to 5,000) labeled by expert E in the MIT-fiveK dataset. We also conducted the representation experiments on general 8-bit imaged datasets: validation set of DIV2K \cite{DIV2kdataset}, which includes 100 images, and Kodak \cite{Kodakdataset}, containing 24 images. All images are center-cropped and downsampled to a size of 256. Coordinates are normalized to $[-1,1]$ as per prior works.
Our method is compared with existing methods, including Tanh and ReLU activations with position encoding (+P.E) \cite{mildenhall2020nerf}, wavelet \cite{wirewavelet2023}, Gaussian \cite{activationinr_gauss_etc}, sine activation \cite{sitzmann2019siren}, and variable periodic activation \cite{liu2023finer}. All reported values, including baselines, are evaluated after the quantization (\cref{eq:quantization}). For a fair comparison, we take the average and standard deviation of the number of iterations and train baselines for a larger number than our average. We adopt the sine activation function for generality and use the BCE loss function unless otherwise stated. In \cref{sec:abl}, we conduct ablation studies on activation functions and loss functions. All networks have an identical number of parameters: 5 hidden layers, each with 512 dimensions, ensuring a fair comparison. We use NVIDIA RTX 3090 24GB for training and optimized all networks by Adam \cite{Adamoptimizer}, with a 1e-4 learning rate.
\begin{figure}[!h]
    \vspace{-5pt}
        \centering
   \includegraphics[trim=0 0 0 0 ,clip,width = 2.9in]{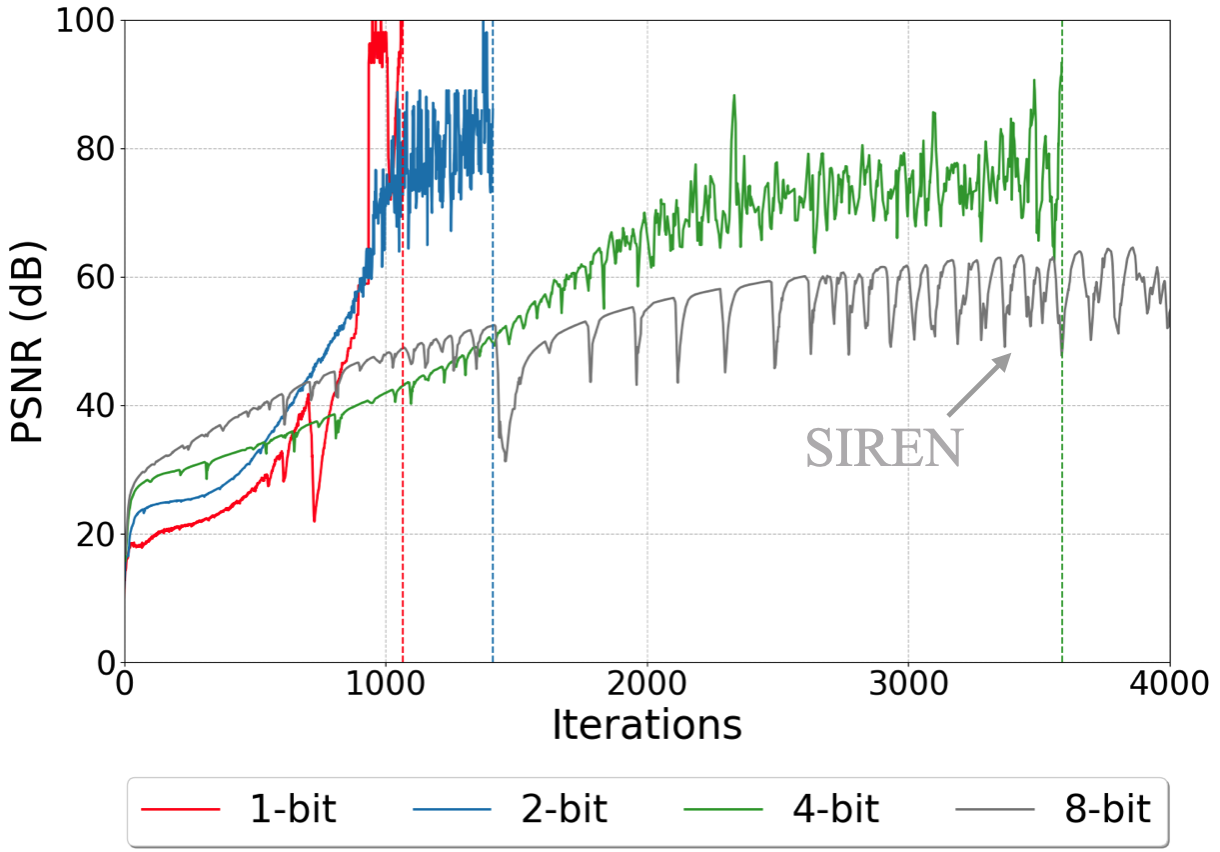}
   
    \vspace{-10pt}
\caption{Comparison of convergence curve for an 8-bit image based on bit precision. Vertical dashed lines indicate the iteration when the model achieves lossless. We show that when $\mathcal{P}(f_\theta)$ is constant and close to $\mathcal{U}_d(n)$, convergence occurs effectively, enabling lossless representation.}
\label{fig:valid_fg1b}
   \vspace{-15pt}
\end{figure}

\subsection{Image Representation}\label{sec:quali_results}
\textbf{Validation}
We quantify the theoretical upper bound $\mathcal{U}$ of INRs with a given bit precision. We provide experimental evidence supporting our hypothesis: if $\mathcal{P}(f_\theta)$ is close to the upper bound $\mathcal{U}_d$, then it is more efficient to achieve \cref{eq:def_lossless}. We set all networks with the same number of parameters and set bit-precision ($n$) as a variable. The detailed figure is in the supplementary. \cref{fig:valid_fg1b} shows the experiment results for our hypothesis. 
In \cref{tab:hypothesis_test}, the proposed method performs best against others regarding fast convergence. There are two reasons for fast convergence while the upper bound ($\mathcal{U}$) is higher than the second column of \cref{tab:hypothesis_test}. First, $\mathcal{L_{\text{BCE}}}$ converges faster than $\mathcal{L_{\text{MSE}}}$ as in \cref{fig:ablation_loss}. Second, the experimental group uses the same number of layers and hidden parameters for quantized images while the bit bias exists in the image, which is inefficient.

\begin{table}[t]
\footnotesize
\centering
\setlength{\tabcolsep}{1.2pt}
    \begin{tabular}{l|c|c|c|c|c}
    \hline
     Method& \makecell[c]{Bit \\ Precision ($n$)} &  \makecell[c]{$\mathcal{U}_d(n)$\\$(\cdot \times \mathfrak{C})$} & \makecell[c]{Loss\\Function} & \makecell[c]{PSNR\\(dB)} & \makecell[c]{\#Iter.\\ (Mean $\pm$ std)} \\ 
    \hline
      \multirow{3}{*}{ \makecell[l]{ Experiment \\ Group} }& 1 & $16$ & \multirow{3}{*}{$\mathcal{L}_{\text{MSE}}$}&\multirow{3}{*}{\makecell[c]{$\infty$\\(Lossless)}}  & $1233 \pm 241$ \\ 
       &2 & $1.30K$ &  & & $1288 \pm 235$ \\ 
      & 4 & $0.81M$ &  & & $3852 \pm 710$ \\
    \hline
    \hline
     SIREN &8& $67.7G$ & $\mathcal{L}_{\text{MSE}}$ & 102.8 & 5000 \\ 
    \hline
    \textbf{Proposed} & 1 & $64$ & $\mathcal{L}_{\text{BCE}}$ &\makecell[c]{$\infty$\\(Lossless)}  & \textbf{778} $\pm$ \textbf{83} \\ 
    \hline
    \end{tabular}
    \vspace{-5pt}
\caption{Quantitative result of our hypothesis test experiment on Kodak \cite{Kodakdataset}. The number of iterations is proportional to $\mathcal{U}_d(n)$.}
\vspace{-20pt}
\label{tab:hypothesis_test}
\end{table}


\noindent\textbf{Quantitative Results} In \cref{tab:quan_main}, we report peak signal-to-noise ratio (PSNR(dB)), structural similarity index measure (SSIM), root mean squared error (RMSE) and bit-error-rate (BER) for evaluation on 16-bit and 8-bit image datasets. We pick the best values of metrics for each image during the training. RMSE and SSIM are calculated using integer value.
Our method accomplishes lossless representation on all 16-bit and 8-bit images in experiments. Our method converges faster than all other baselines. \cref{fig:learning_curve} represents the results on a single 8-bit image. Note that minimizing BER is highly correlated to increasing PSNR but not equivalent. As a result, our proposed method has consistently low BER throughout the learning process, while the PSNR does not.
\cref{tab:combination} demonstrate that our proposed method can be applied to other existing INR approaches. We integrate our method with conventional approaches that use hash inputs \cite{muller2022instantngp,xie2023diner}, as well as with efficient activation functions \cite{activationinr_gauss_etc,liu2023finer}. For hash-based methods \cite{muller2022instantngp,xie2023diner}, we follow the settings specified in their respective papers. We measure the average number of iterations for each result, ensuring that each baseline was trained sufficiently for a fair comparison. Our method is compatible with existing methods while achieving lossless representation.

\noindent\textbf{Qualitative Results} We report a visual comparison of under-fitted images in \cref{fig:qual_convergence}. Different artifacts occur during training, such as blurry artifacts for WIRE \cite{wirewavelet2023} and SIREN \cite{sitzmann2019siren}, or noise-like artifacts for Gauss \cite{activationinr_gauss_etc}. Our method learns both high-frequency and low-frequency components faster than the others; however, salt-pepper impulse noise artifacts are present in the training stage. In \cref{fig:demo_qual}, we present converged images with all baselines. Since the converged images are not easily discernible to human eyes, we highlight the occurrence of dominant errors ($\geq\text{MSB-}4$). The error map indicates residual between the ground truth (GT). Our method makes INR represent no error in image representation.

\begin{figure}
\vspace{-10pt}
\footnotesize
\centering
\includegraphics[trim=0 0 0 0 ,clip,width = 3.2in]{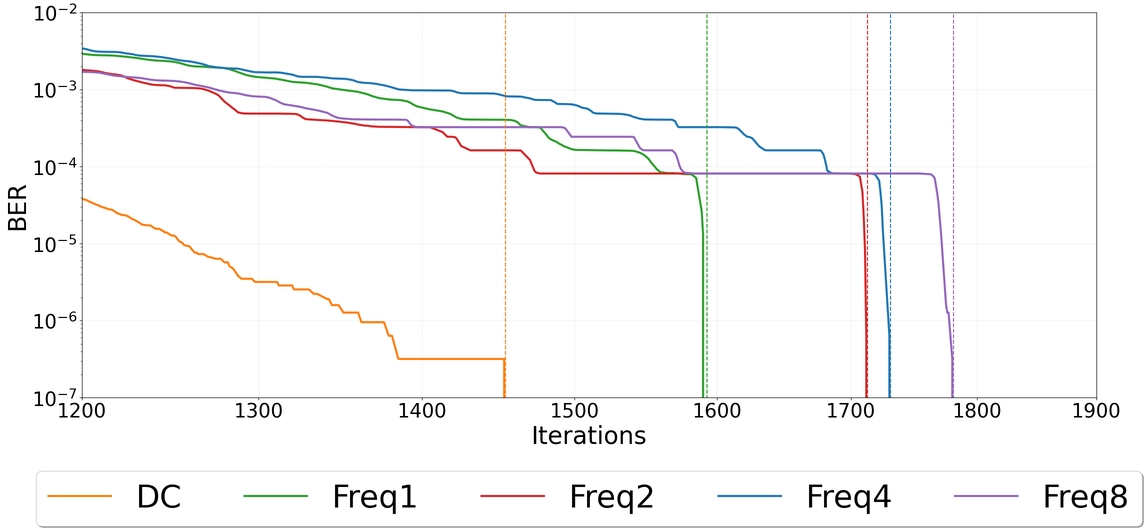}
\vspace{-7pt}
\caption{Quantitative comparison based on frequency to the bit axis. Vertical dashed lines indicate the iteration when the model achieves lossless.}
\vspace{-10pt}
\label{fig:quanti_bitwise_bias}

\end{figure}

\subsection{Bit \& Bit-Spectral Bias}

\textbf{Bit Bias} In this section, we examine two different biases that we discovered: 1) `\textit{bit bias}' and 2)`\textit{bit-spectral bias},' validating them through experiments. In \cref{fig:bit-bias}, conducted on a single 16-bit image, the experiment quantitatively demonstrates the presence of bit bias across all tested baselines. 
Whether weights are assigned or not to the MSBs \footnote{Baselines are equivalent to assigning weights to bits.}, there is a common challenge in representing the LSBs.
\cref{fig:bit-bias_qual} shows that fitting artifacts mentioned in \cref{sec:quali_results} are related to bit bias. According to \cref{fig:bit-bias_qual}, MSBs are nearly indiscernible between INRs and GT. However, significant differences were observed in LSBs $\mathbf{B}^{(i)} (i\leq 13)$. In conclusion, the proposed method effectively reduces bit bias resulting in representing the signal's LSBs. 


\noindent\textbf{Bit-spectral Bias} In the following, we study our method's bit axis and its bias. Spectral bias \cite{spectralbias} exists on the bit axis. Therefore, specific pixel values are difficult to represent in our structure. The experiment is conducted on the 16-bit synthetic CYMK-RGB image, where we set frequency along the bit axis as a variable.  \cref{fig:quanti_bitwise_bias} quantitatively shows the existence of bit-spectral bias. In conclusion, our method parameterizes specific values like 65,535 or 0 (DC) faster than high-frequency values. Implementation details and qualitative results are shown in the supplementary material.

\begin{table}[t]
\footnotesize
\begin{tabular}{
l|c|c|c|c|c
}
\hline
&Weight &\makecell[c]{Model Size\\(Byte$\downarrow$)} & \makecell[c]{PSNR\\(dB$\uparrow$)} &\#Iter.($\downarrow$)  & \makecell[c]{\#BitOps.\\(/Pixel)}\\
\hline

\textbf{Ours$^{-}$} &$\{-1,0,1\}$&656.18K & $\infty$  & 80K &50.66M \\
SIREN&FP32& 1.27M &119.7&200K& 405.2M\\
\hline

\end{tabular}
\vspace{-7pt}
\captionof{table}{Model size and weight \& performance comparison in the aspect of model quantization.}
    \label{tab:model_quant}
\vspace{-20pt}
\end{table}

\begin{figure}[t]
\vspace{-10pt}
        \centering
        \includegraphics[trim=0 0 0 0,clip,width=3.2in]{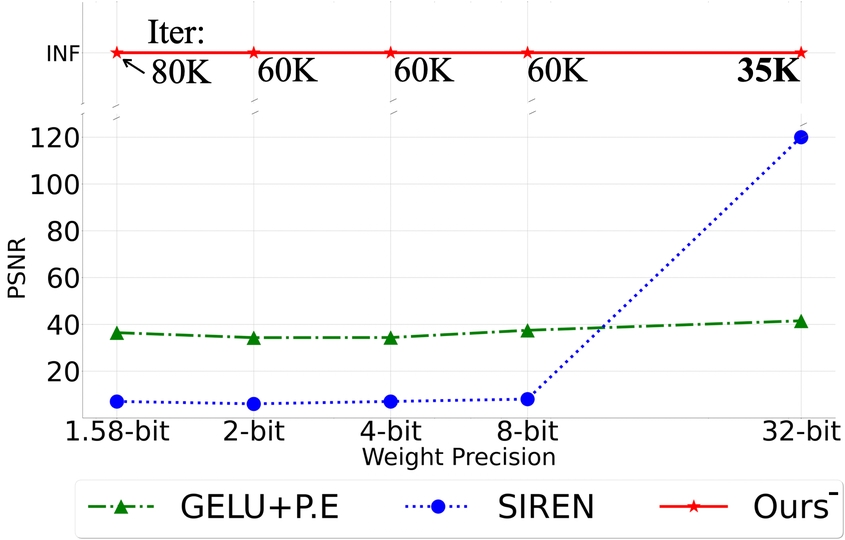}
        \vspace{-7pt}
        \caption{Quantitative comparison of performance (Y-axis) according to the parameter precision (X-axis)}
        \vspace{-10pt}
        \label{fig:Quanti_modelquant}
\end{figure}

\subsection{Applications}
We propose new applications by using our method. We introduce ternary INR with extreme weight quantization, bit-depth expansion using a bit axis, and lossless compression utilizing lossless representation. The implementation details are in the supplement material.

\noindent\textbf{Ternary Implicit Neural Representation}
The intuitive question is whether 32-bit floating precision (FP32) parameters are necessary when parameterizing outputs with 1-bit precision. To address our concern, we design a ternary-weighted (1.58-bit) implicit neural representation for an image fitting that employs the novel method proposed by \citet{1bitllm_bitnet} and \citet{158bitllm}.
Each fully connected layer has ternary weights and calculates its output as follows: 
\vspace{-5pt}
\begin{align}
        y = \beta \gamma \tilde{\mathcal{W}}\tilde{x}, \quad (\tilde{\mathcal{W}} \in \{ -1, 0 ,1 \}^{d_{\text{out}}\times d_{\text{in}}}),
\end{align}
where $\tilde{x}$ is layer normalized and quantized values of input $x$ and $\beta := \frac{1}{d_{\text{in}}d_{\text{out}}}||\mathcal{W}||_1$, $\gamma := ||x||_\infty$ as suggested in \cite{1bitllm_bitnet}. 


In \cref{tab:model_quant,fig:Quanti_modelquant}, we show the performance comparison of weight-quantized INRs.  As shown in \cref{fig:Quanti_modelquant}, our proposed method accomplishes lossless image representation until ternary weights. 
The activation for our ternary INR should be the GELU \cite{hendrycks2016GELU} function, and we denote it as `Ours$^-$'. Periodic activations must follow the strict weight initialization \cite{sitzmann2019siren,liu2023finer}; breaking such initialization by the quantization makes the network collapse. The numbers under `Ours$^-$' in \cref{fig:Quanti_modelquant} indicate the minimum iteration number for each model. 
In \cref{tab:model_quant}, We report model size, the number of parameters of $f_\theta$ (i.e. $\mathcal{P}(f_\theta)$) in bytes and the number of bit operations (BitOps). The proposed method requires less storage and BitOps than SIREN.

\begin{table}[!t]
    \centering
    \footnotesize

    \begin{tabular}{l|c|cc}
    \hline
     \multicolumn{2}{c|}{ PSNR(dB$\uparrow$)$|$SSIM($\uparrow$))}&  $8$-bit $\rightarrow$ 16-bit &  $8$-bit $\rightarrow$ 12-bit   \\
     \hline
     \hline
     ZP&\multirow{3}{*}{Rule-based}&52.92$|$0.9990 &53.31$|$0.9990 \\
     MIG& &\textcolor{black}{55.91}$|$\textcolor{black}{0.9991}& 55.93$|$0.9991 \\
  BR \cite{ulichney1998pixel_br}& &52.98$|$0.9991 & 53.32$|$0.9991 \\
     \hline
     BECNN \cite{su2019photo}&\multirow{3}{*}{Supervised} &53.14$|$0.9986  & N/A \\
     BitNet \cite{byun2018bitnet}& & 53.60$|$0.9970& N/A  \\
     ABCD \cite{abcd} & & \textcolor{red}{59.39}$|$\textcolor{red}{0.9997} &\textcolor{red}{59.37}$|$\textcolor{red}{0.9995}\\
     \hline
     \textbf{Ours}&Self-supervised&\textcolor{blue}{55.92}$|$\textcolor{blue}{0.9993}  & \textcolor{blue}{55.94$|$0.9997}\\
     \hline
    \end{tabular}
    \vspace{-7pt}
        \caption{Quantitative comparison in the bit-depth expansion on TESTIMAGES\cite{asuni2013testimages}. \textcolor{red}{Red} and \textcolor{blue}{blue} indicate the best and the second-best performance, respectively. `N/A' indicates not applicable.}
    \vspace{-18pt}
        \label{tab:bde_appli}
\end{table}



\noindent\textbf{Bit Depth Expansion} Our method conducts bit depth expansion by extrapolating the bit-axis. In \cref{tab:bde_appli}, we conduct a quantitative comparison with existing methods. We train on the 8 MSBs of a 16-bit image and predict the lower 8 bits without using the 16-bit ground truth.
To the best of our knowledge, our method is the first attempt at a self-supervised learning approach for bit depth expansion.
Our method performs superior than existing rule-based algorithms or learning-based methods (BitNet \cite{byun2018bitnet} and BECNN\cite{su2019photo}). 

\begin{table}[!t]
    \centering
    \footnotesize
    \begin{tabular}{l|cc}
    \hline
     Bits Per Pixel (bpp)($\downarrow$)&  MNIST & Fashion MNIST   \\
     \hline
     \hline
     PNG \cite{roelofs1999png} & 3.52\textcolor{black}{(+36\%)} & \textcolor{red}{5.78}\textcolor{black}{(-5\%)} \\
     JPEG2000 \cite{taubman2002jpeg2000} &6.75\textcolor{black}{(+162\%)} & 7.74\textcolor{black}{(+27\%)}\\
     WebP \cite{si2016researchwebp}&  \textcolor{red}{2.11}\textcolor{black}{(-18\%)}& 6.60\textcolor{black}{(+8\%)} \\
     TIFF \cite{poynton1992overview_tiff}&  3.93\textcolor{black}{(+52\%)}&  6.76\textcolor{black}{(+11\%)}\\
     \hline
     RECOMBINER\cite{he2023recombiner}+\textbf{Ours}&\textcolor{blue}{2.58}  & \textcolor{blue}{6.11}\\
     \hline
    \end{tabular}
    \vspace{-7pt}
     \caption{Quantitative comparison for lossless compression. \textcolor{red}{Red} and \textcolor{blue}{blue} indicate the best and second-best performance, respectively.}
     \vspace{-7pt}
     \label{tab:compression}
\end{table}

\begin{table}[!t]
    \footnotesize
    \centering
    \vspace{-5pt}
    \label{tab:sub2}
    \setlength{\tabcolsep}{1.2pt}
    \begin{tabular}{l|cc|c|c|cc|c|c}
    \hline
     & \multicolumn{4}{c|}{MNIST} & \multicolumn{4}{c}{Fashion MNIST}\\
     \cline{2-9}
      & bpp&PSNR&SSIM&RMSE&  bpp  &PSNR&SSIM&RMSE  \\
     \hline
     \hline
      \scriptsize{RECOMBINER \cite{he2023recombiner}} & 4.20&48.60&0.994&0.945& 9.06 & 56.64&0.996&0.375 \\
      \scriptsize{RECOMBINER+\textbf{Ours}} &\textbf{2.58} & $\infty$&1.000&0.000&\textbf{6.11}& $\infty$&1.000&0.000\\
    \hline
        \end{tabular}
        \vspace{-10pt}
            \caption{Quantitative comparison between RECOMBINER \cite{he2023recombiner} and RECOMBINER with our method.}
            \label{tab:compress_futrher}
        \vspace{-10pt}
\end{table}

\noindent\textbf{Lossless Compression} We conduct lossless compression experiments by applying our method to the state-of-the-art INR compression method \cite{he2023recombiner}. In \cref{tab:compression}, a simple combination of \cite{he2023recombiner} and ours shows superior results compared to existing lossless image codecs such as PNG \cite{roelofs1999png}, JPEG2000 \cite{taubman2002jpeg2000}, WebP \cite{si2016researchwebp}, and TIFF \cite{si2016researchwebp}. 
In \cref{tab:compress_futrher}, \cite{he2023recombiner} cannot achieve lossless representation, even if the bpp is significantly increased.


\section{Discussion}\label{sec:abl}

\begin{figure}[!t]
\centering
    \centering
            \includegraphics[trim= 0 0 0 0, clip,width = 3.2in]{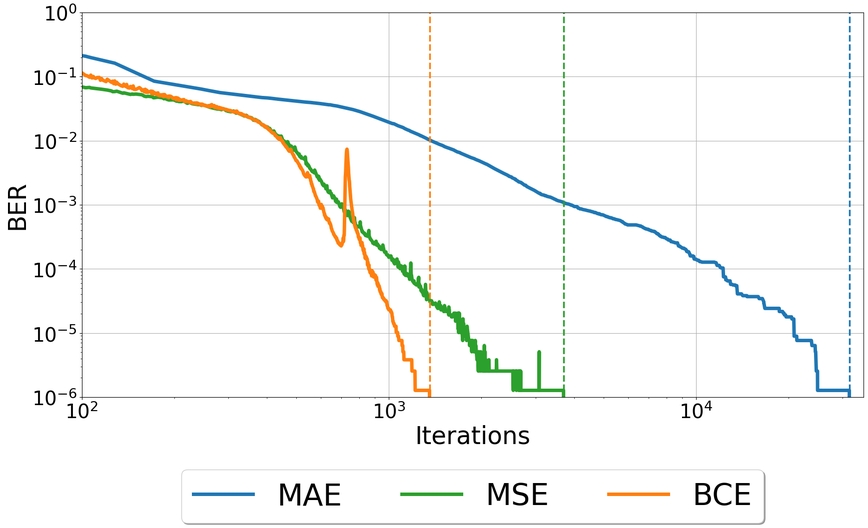}
        \vspace{-10pt}
    \caption{Quantitative ablation study on loss function of our method on TESTIMAGE\cite{asuni2013testimages}. Vertical dashed lines indicate the iteration when the model achieves lossless representations.}
    \label{fig:ablation_loss}
    \vspace{-20pt}
\end{figure}

\noindent \textbf{Ablation Study} We conduct ablation studies for the loss function of the proposed method. We utilize a 16-bit sample image in the TESTIMAGE dataset \cite{asuni2013testimages} and reduce the model size to observe convergence speed. 
In \cref{fig:ablation_loss}, the performance of MSE is close to that of BCE, while MAE exhibits a slow convergence speed. Although achieving lossless representation through MSE or MAE is possible, BCE shows the fastest convergence speed.

 We conduct an ablation study on the input dimension $d$. In \cref{tab:abl_dim}, we extend a 3-dimensional coordinate to a 4-dimensional one by incorporating color as a coordinate. As \cref{eq:boundfunction}, increasing a dimension increases the upper bound and makes INR converge slower than the proposed method.

\begin{table}[!t]
    \centering
    \footnotesize
    \begin{tabular}{l|cc}
                    \hline
              Method   & Proposed   &  Tested         \\
            Coord. & $\mathbf{x} = (h,w,i)$ & $\mathbf{ x} = (h,w,i,c)$\\
            \hline
                \hline
                \#Iter.($\downarrow$) &  \textbf{790} &  1438 \\
            \hline
    \end{tabular}
    \vspace{-7pt}
     \caption{Quantitative ablation study of our method on Kodak \cite{Kodakdataset} as the input dimension $d$ increases ($3\rightarrow4$).}
     \vspace{-8pt}
     \label{tab:abl_dim}
\end{table}

\begin{table}[t]
    \centering
    \scriptsize
    \begin{tabular}{l|cc}
    \hline
     Text &\multicolumn{2}{c}{Jack would become Eva's happy husband} \\

    \hline
    \hline
    Method    &  PSNR(dB)($\uparrow$) & Prediction from \cite{SileroModels_stt}   \\
    \hline
  GT Audio &- &Jack would become \textcolor{red}{even} happy \textcolor{red}{ashon} \\
     SIREN \cite{sitzmann2019siren} & 68.54&\textcolor{red}{Jark will} become \textcolor{red}{evil's haring ho}\\
     DINER \cite{xie2023diner}&85.19 & \textcolor{red}{Jar} would become  \textcolor{red}{even hary ashon}  \\
     Ours &$\infty$ &Jack would become \textcolor{red}{even} happy \textcolor{red}{ashon} \\
     \hline
    \end{tabular}
    \vspace{-7pt}
     \caption{Qualitative comparison on speech to text (STT) results of the represented audio using a pre-trained STT network \cite{SileroModels_stt}.}
     \vspace{-10pt}
     \label{tab:audio_classify}
\end{table}

 \begin{figure}[t]
    \centering
   \hspace{-1pt}
   \includegraphics[trim=0 0 0 0,clip,width = 3.2in]{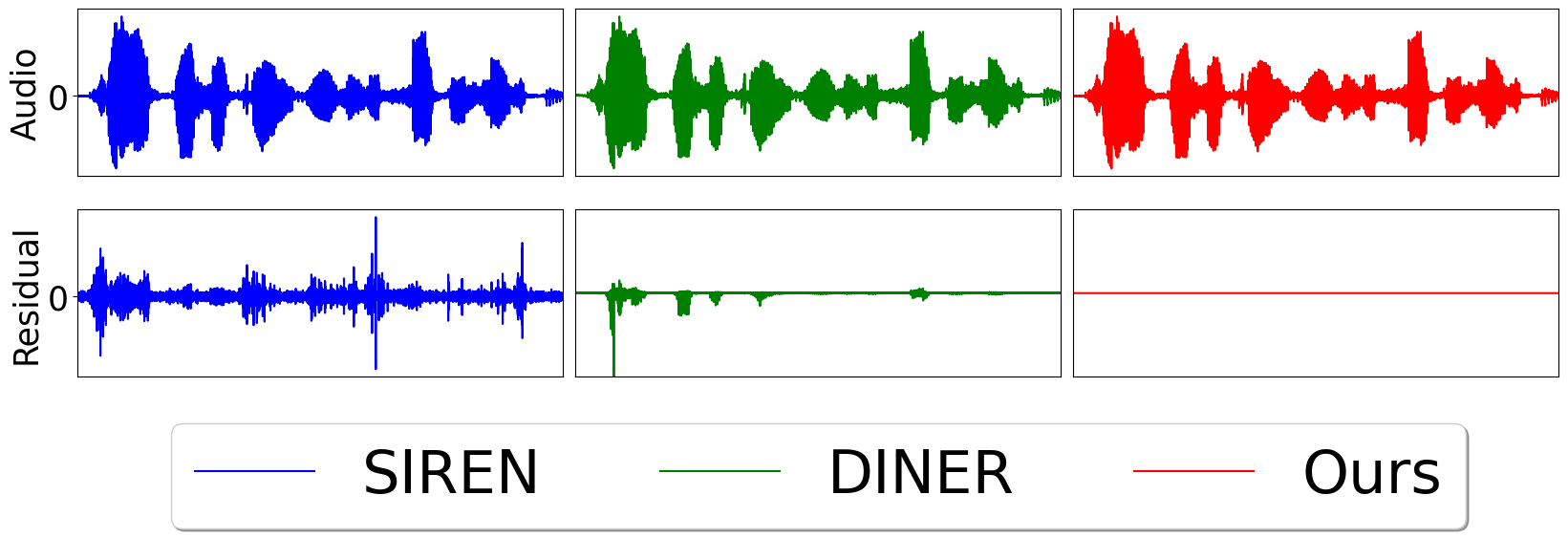}
      \vspace{-7pt}
       \caption{Qualitative comparison on representing the Librispeech \cite{panayotov2015librispeech} data with a floating point precision (FP32) and its residual.}
     \vspace{-18pt}
     \label{fig:audio}
  \end{figure}

\noindent\textbf{Floating Point Representation} Our approach has been discussed in the context of fixed precisions. We also verified whether our method can be applied to floating-point data, such as audio. The detailed formulation is in the supplement. In \cref{fig:audio}, we show the representation result of Librispeech data \cite{panayotov2015librispeech} and demonstrate that the FP32 data format is fitted by our method losslessly. In \cref{tab:audio_classify}, we report the speech-to-text (STT) results predicted by pre-trained model \cite{SileroModels_stt} and compare our method with other methods \cite{sitzmann2019siren,xie2023diner}.

\noindent \textbf{Limitation} As mentioned by \citet{mathdeeplearningtheory}, when $d$ exceeds 5, such as in radiance fields, the suggested upper bounds $\mathcal{U}_d$ increase extremely high ($\mathcal{U}_{5}(8)\simeq 1.23\mathfrak{C}\times10^{27}$). Although our proposed approach performs better in representing low-dimensional data, the main drawback lies in predicting high-dimensional data.  Further research is needed to explore parameter-efficient learning; thus, we demonstrate the use of recent techniques in the supplement material. 

\section{Conclusion}
We quantify the upper bound of the size of INRs based on the given bit precision. Through bit-plane decomposition, we achieve lossless representation, which was previously unachievable. With experiments, we validate our hypothesis that "lowering the upper bound accelerates the achievement of lossless representation in INR."
Furthermore, we reinterpret the concept of spectral bias from a digital computing perspective and explain new notions of `bit bias.' 
Our method mitigates the bit bias and makes INR represent true LSBs, resulting in lossless representation.
We demonstrate that our method enables true lossless representation in followed applications: ternary networks, lossless compression, and bit-depth expansion.

\vspace{8pt}
\noindent
\section*{Acknowledgments}
\small{
This work was partly supported by the National Research Foundation of Korea (NRF) grant funded by the Korea government (MSIT) (RS-2024-00335741) and (RS-2024-00413303).
}
\noindent
{
    \small
    \bibliographystyle{ieeenat_fullname}
    \bibliography{main}
}

\clearpage
\setcounter{equation}{1}

\setcounter{page}{1}
\setcounter{table}{0}
\setcounter{section}{0}
\setcounter{figure}{0}
\maketitlesupplementary
\normalsize
\section{Theory} \label{sup_sec:theory}
\begin{table*}[ht!]
\footnotesize
    \centering
    \setlength{\tabcolsep}{1.2pt}

    \begin{tabular}{c|l|l|l}
        \hline
        \textbf{Symbol}&\textbf{Definition} & \textbf{Description} & \textbf{Example/Meaning/Note} \\ \hline
        $d$     &  $\in \mathbb{N}$    & Dimension of Function (Signal) or Vector & $d=2$ for an Image \\
        $n$     &  $\in \mathbb{N}$    & Bit Precision a Ground Truth Function & $n=8$ for an 8-bit (uint 8) Image\\ 
        $k$     &  $\in \text{div}^{+}{(n)}$    & Bit Precision a Represented Function & Control Variable in \cref{tab:hypothesis_test}  \\ 
        $i$     &  $\in \mathbb{N}\cap(0,\frac{n}{k}]$    & Index of a Quantized Function& $i=\{0,1,2,3\}$ in case of $n=8$ and $k=2$\\ 
        $H,W,C$     &  $\in \mathbb{N}$    & Height, Width, Channels of Function&   \\ 
        $L$     &  $\in \mathbb{R}$    & Lipschitz Constant & Details in \cref{sup_sec:theory}  \\ 
        $\epsilon$     &  $\in \mathbb{R}$    & Error (or distance) & Quantization Error in our paper\\ 

        $\mathbf{x}$ &  $\in (\mathcal{X} \subseteq \mathbb{R}^d)$    & Input Vector of a Function & \\
        $\mathbf{I}$ &  $\in \mathbb{R}^{H\times W\times C}$    & Analog $C$-channel Image & \\
        $\mathbf{I}_n, \mathbf{Q}_k$ &  $\in {Q}_{n}^{H\times W\times C},{Q}_{k}^{H\times W\times C}$    & Digital n-bit (or $k$-bit) $C$-channel Image & $n=8,16$ for images\\
        $\mathbf{B}$ &  $\in \{0,1\}^{H\times W\times C}$    &Bit-plane of an image& In case of $\mathbf{Q}_{k=1}$ \\
        $\mathcal{W}$     &  $\in \mathbb{R}^{d_{\text{out}}\times d_{\text{in}}}$    &Weight Matrix (Trainable Parameters)&  \verb|nn.Linear.weight| (Pytorch) \\ 
        $\mathbf{b}$     &  $\in \mathbb{R}^{ d_{\text{out}}}$    & Bias Vector (Trainable Parameters)&  \verb|nn.Linear.bias| (Pytorch) \\ 
        $\theta$     &  $\in \{\varprod_{i=0}^{M-1}(\mathbb{R}^{d_{\text{out}}\times d_{\text{in}}},\mathbb{R}^{ d_{\text{out}}})\} (:= \Theta)$    &Trainable Parameters of $M$-Layer MLP and its Set& \\ 
        $\beta$     &  $:=\frac{1}{d_{\text{in}}d_{\text{out}}} ||\mathcal{W}||_1$    &Scaling Factor for a Weight Quantization&  \\ 
        $\gamma$     &  $:=||x||_\infty$    &Scaling Factor for an Activation Quantization&  \\ 
        $\mathfrak{C}$     &  $:= 9 \cdot (3d \max\{L(b-a), 1\})^{2d} \cdot d^2$    &Coefficient of $\mathcal{U}_d(n)$ &   Details in \cref{sup_sec:theory}\\ 

\hline
        $[a,b]$     &  $:=\{x\in\mathbb{R} | a\leq x \leq b \}$    &Domain of a function&  $a,b$ satisfy $(a,b\in\mathbb{R}, a\leq b)$\\ 
        $Q_n$     &  $\subset \mathbb{Q}\cap [0,1]$    &Codomain (or Range) of a Digital Function& $Q_n = \{0,\frac{1}{3},\frac{2}{3} ,1\}$ in case of $n=2$  \\ 
        $[\mathbf{Q}^{(i)}_k]^{j}_{i=0}$ &$\varprod_i^j {Q}_{k}^{H\times W \times C}$     & Sequence a Quantized Function&LSBs to MSBs as $i$ increased \\ 
        $[\mathbf{B}^{(i)}_k]^{j}_{i=0}$ &$\varprod_i^j {\{0,1\}}_{k}^{H\times W \times C}$     & Sequence a Bit-Plane & LSB to MSB as $i$ increased  \\ 
\hline
        $f,h$     &  $\mathbb{R}^d \rightarrow \mathbb{R}^k$    & Analog function  &  $k=1$ in \cref{sec:prelim} of main paper.\\ 
        $f_n,h_n$     &  $\mathbb{R}^d \rightarrow {Q}^k_n$    & Digital function with $n$-bit precision  &  \\ 
        $f_\theta,h_\theta$     &  $\mathbb{R}^d \rightarrow \mathbb{R}^k$    & Function that parameterized with $\theta$  & Implicit Neural Representation (INR)  \\ 
            $\mathcal{Q}_n(\cdot)$     &  $  \mathbb{R}^d \rightarrow Q_n^d,\quad \hat{x}\mapsto \scriptsize{\arg\min_{x\in Q_n} ||x-\hat x||_1}$    &$n$-bit Quantization &  Element-wise operation for vector inputs\\ 
    $\mathcal{P}(\cdot)$     &  $ \Theta \rightarrow \mathbb{N}$    & Number of Parameters of a Neural Network &  \\ 
    $\epsilon(\cdot)$     &  $\mathbb{N} \rightarrow \mathbb{R}, \quad n \mapsto \frac{1}{2(2^n-1)}$    & Upper bound of a quantization error with given $n$&  \\ 
    $\mathcal{U}_d(\cdot)$     &  $\mathbb{N} \rightarrow \mathbb{N} \quad n \mapsto \mathfrak{C}(2^{n+1}-2)^{2d}$    & Upper bound of a $\mathcal{P}$ with given $n$ and $d$&  \\ 
    $\widehat{(\cdot)}$     & $\simeq(\cdot)$ & Prediction to $(\cdot)$ & Applied to elements or functions  \\ 
    \hline
    \end{tabular}
    \caption{Notation table for the main paper (Elements, Sets, and Functions (calculations), respectively)}
    \label{tab:notation}
\end{table*}

 The specific theorems presented in \cite{mathdeeplearningtheory} are as follows:
\begin{theorem*}
\label{theorem:main}
(The implicit ANN approximations with described error tolerance and explicit parameter bounds by \citet{mathdeeplearningtheory}, Proposition 4.3.8, Corollary 4.3.9) Let $d\in \mathbb{N}$, $L,a\in \mathbb{R}$, $b\in [a,\infty)$, $\epsilon \in (0,1]$ and function $f$ satisfy for $\forall x,y \in [a,b]^d$ that $ |f(x) - f(y)| \leq L \|x - y\|_1$, 

Then there exist $f_\theta$ that satisfy : 
\begin{enumerate}
    \item It holds the Upper Bound on Network Error:
    
    $\sup||f_\theta(\mathbf{x}) - f(\mathbf{x})||_1\leq \epsilon$
    \item Upper Bound on Number of Layers: 
    
    $d(\log_2(\max\{\frac{3dL(b-a)}{2},1\}) + \log_2(\epsilon^{-1})) + 2$
    \item Upper Bound on Number of Channels of each layer:
    
    $\epsilon ^{-d} d (3\frac{(3dL(b-a))^d}{2^i}+1)$
    \item Upper Bound on Network Parameters:
    
    $\epsilon ^{-2d} 9 (3d\max\{L(b-a),1 \})^{2d}d^2$
\end{enumerate}

\end{theorem*}
\noindent In this section, we provide a brief proof of the theory.  We also demonstrate $\mathfrak{C}$ in our setup. Note that \cref{tab:notation} shows all notations for the main paper and \cref{tab:notation_supple} shows notations for supplementary material. 

\subsection{Proof of \cref{theorem:main}}
 According to \citet{mathdeeplearningtheory}, the proof of the theorem is derived as follows. The proof proceeds by designing a function that satisfies the \cref{prop:_conv}, substituting the L1 distance and maximum value approximated by the ANN, and then generalizing a distance to an arbitrary number. For brevity, we provide a summarized outline of the proof. For a rigorous mathematical proof, please refer to the original document.
 \begin{proposition} 
 \label{prop:_conv}
     Let $(E,\delta)$ be a metric space and $L\in [0,\infty)$, $\varnothing \neq \mathcal{M} \subseteq E $, and $f:E\rightarrow \mathbb{R}$ that satisfy $\forall x\in E, y\in\mathcal{M}$ s.t. $|f(x)-f(y)| \leq L \delta(x,y)$. Let $F : E\rightarrow \mathbb{R}\cup \{\infty\}$ for all $x\in E$ that 
     \begin{equation*}
         F(x) = \sup_{y\in\mathcal{M}}[f(y)-L\delta(x,y)].
     \end{equation*}
     Then, it holds  $\forall x\in E$ that
      \begin{equation*}
         |F(x)-f(x)| \leq 2L[\inf_{y\in \mathcal{M}}\delta(x,y)].
     \end{equation*}
 \end{proposition}
\noindent Let the notation $\mathbf{A}_{\mathcal{W},\mathbf{b}}$ indicates an affine transforms with weight ($\mathcal{W}$) and bias ($\mathbf{b}$), and $\mathbb{T}_{d,K}$ indicates an ANN that satisfies $\mathbb{T}_{d,K}(\mathbf{x}) = \underbrace{[\mathbf{x}^{T},\mathbf{x}^{T},\cdots \mathbf{x}^{T}]^{T}}_{K}$ with $d,K \in\mathbb{N}$. With ReLU activation, L1 distance is represented by 2-layer MLP as below:
\begin{definition}
\label{def:l1}
    Let weights ($\mathcal{W}^{(1,2)}$) and bias ($\mathbf{b}^{(1,2)}$) of the affine transform be as follows:
    \begin{align}
    \mathcal{W}^{(1)} :=\begin{bmatrix}
        1 \\
        -1
    \end{bmatrix}     \mathbf{b}^{(1)} :=\begin{bmatrix}
        0 \\
        0
    \end{bmatrix} ,      \mathcal{W}^{(2)} :=\begin{bmatrix}
        1 & 1
    \end{bmatrix}
    \mathbf{b}^{(2)} :=\begin{bmatrix}
        0 
    \end{bmatrix}.
    \end{align}
 $\forall x\in \mathbb{R}$, the ANN $\mathbb{L}_1 (x) = |x|$ is defined as below:
    \begin{align}
        \mathbb{L}_1(\mathbf{x}) :=&  \mathbf{A}_{\mathcal{W}^{(2)},\mathbf{b}^{(2)}}(\sigma(\mathbf{A}_{\mathcal{W}^{(1)},\mathbf{b}^{(1)}}(\mathbf{x}))),
        \\
        \text{where, } \sigma(x) &:= \max(x,0).
    \end{align}
    Then, $\forall \mathbf{x}\in \mathbb{R}^d$ and $d\in\mathbb{N}$ , the ANN $\mathbb{L}_d (x) = ||\mathbf{x}||_1$ is defined as below:
    \begin{align}
            \mathbb{L}_d(\mathbf{x}) :=&  \mathbf{A}_{\mathcal{W}^{(2)}_d,\mathbf{b}^{(2)}_d}(\sigma(\mathbf{A}_{\mathcal{W}^{(1)}_d,\mathbf{b}^{(1)}_d}(\mathbf{x}))),
    \end{align}
    where $\mathcal{W}^{(1)}_{d} \in \mathbb{R}^{2d\times d},\mathcal{W}^{(2)}_{d} \in \mathbb{R}^{1 \times 2d},\mathbf{b}^{(1)}_{d},$ and $\mathbf{b}^{(2)}_{d}$ are as below:
    \begin{align}
        \mathcal{W}^{(1)}_d &:= \mathbf{E}_d \otimes \mathcal{W}^{(1)} =\begin{bmatrix}
            \mathcal{W}^{(1)} & 0 & \cdots & 0 \\
             0 & \mathcal{W}^{(1)} &  \cdots & 0 \\
             \vdots & \vdots & \ddots &\vdots \\
             0 & 0 & \cdots &\mathcal{W}^{(1)}
        \end{bmatrix},\\
        \mathcal{W}^{(1)}_d &:= \underbrace{[1,1,\cdots ,1]}_{2d} \quad
        \mathbf{b}^{(1)}_{d} = \vec{0} \in \mathbb{R}^{2d} \quad \mathbf{b}^{(2)}_{d} = [0]
        \end{align}
    where $\mathbf{E}$ is an identity matrix and $\otimes$ is Kronecker product. 
\end{definition}
\noindent We denote $\mathbf{P}_d(\cdot,\cdot,... )$ as $d$-parallel of ANNs and $\bullet$ as sequential of ANNs. Likewise, the definition of the maximum value is defined as below:
\begin{definition}
\label{def:max}
        Let weights ($\mathcal{W}^{(1,2)}$) and bias ($\mathbf{b}^{(1,2)}$) of the affine transform be as follows:
    \begin{align}
    \mathcal{W}^{(1)} &:=\begin{bmatrix}
        1 & -1\\
        0 & 1 \\
        0 & -1 
    \end{bmatrix}     \mathbf{b}^{(1)} :=\begin{bmatrix}
        0 \\
        0 \\
        0
    \end{bmatrix} ,      \\ \mathcal{W}^{(2)} &:=\begin{bmatrix}
        1 & 1 & -1
    \end{bmatrix}
    \mathbf{b}^{(2)} :=\begin{bmatrix}
        0 
    \end{bmatrix}.
    \end{align}
     $\forall \mathbf{x} = [x_1,x_2]^{T}\in \mathbb{R}^2$, the ANN $\mathbb{M}_2 (x) = \max\{x_1,x_2\}$ is defined as below:
    \begin{align}
    \mathbb{M}_2(\mathbf{x}) :=&  \mathbf{A}_{\mathcal{W}^{(2)},\mathbf{b}^{(2)}}(\sigma(\mathbf{A}_{\mathcal{W}^{(1)},\mathbf{b}^{(1)}}(\mathbf{x}))),
    \end{align}

\noindent Then, $\mathbb{M}_d$ for $d\in \mathbb{N} \cap [3,\infty]$ is defined as follow:

    \begin{align}
            \mathbb{M}_d = \begin{cases}
            \mathbb{M}_k  \bullet \mathbf{P}_k(\mathbb{M}_2,\mathbb{M}_2,\cdots,\mathbb{M}_2) &d =2k\\
             \mathbb{M}_k  \bullet \mathbf{P}_k (\mathbb{M}_2,\mathbb{M}_2,\cdots,\mathbb{M}_2,\mathbf{E}_1) &d =2k-1
            \end{cases}
    \end{align}
$\mathbb{M}_d$ satisfy $\forall \mathbf{x} =[x_1,x_2 \cdots ,x_d]^{T}\in \mathbb{R}^d$,  $\mathbb{M}_d(\mathbf{x})= \max\{x_1, x_2, \cdots ,x_d \}$
\end{definition}
 \noindent 
 Then maximum convolution is represented with an ANN $\mathbf{\Phi}(\cdot)$ as follow: 
 \begin{proposition}
 \label{prop:ann_conv}
     Let $d,K\in \mathbb{N}$, $L\in[0,\infty)$, $\mathbf{x}_k\in\mathbb{R}^d$, and $\mathbf{y} =[y_1,y_2,\cdots y_K] \in \mathbb{R}^K$. Then the ANN $\mathbf{\Phi}$ defined as below:
     \begin{align*}
     \label{eq:supple_todo}
                  \mathbf{\Phi} =& \mathbb{M}_K \bullet \mathbf{A}_{-L \cdot \mathbf{E}_K, \mathbf{y}} \bullet \mathbf{P}_K(\mathbb{L}_d\bullet\mathbf{A}_{\mathbf{E}_d,-\mathbf{x}_1},
         \mathbb{L}_d\bullet\mathbf{A}_{\mathbf{E}_d,-\mathbf{x}_2},\\ &\cdots
         \mathbb{L}_d\bullet\mathbf{A}_{\mathbf{E}_d,-\mathbf{x}_K})\bullet\mathbb{T}_{d,K}.
     \end{align*}
\noindent $\mathbf{\Phi}$ holds $\forall \mathbf{x}\in \mathbb{R}^{d}$,
\begin{align}
    \mathbf{\Phi}(\mathbf{x}) = \max_{k\in\{1,2,\cdots K\}}(y_k - L||\mathbf{x}-\mathbf{x}_k||_1).
\end{align}

 \end{proposition}
\label{prop:conv_sub}
 Then with \cref{prop:_conv,prop:ann_conv}, the ANN approximation follows:
 \begin{proposition}
       Let $d,K\in \mathbb{N}$, $L\in[0,\infty)$, $\mathbf{x}_k \in E \subseteq \mathbb{R}^d$. Let $f:E\rightarrow \mathbb{R}$ satisfies $\forall \mathbf{x}_{1,2}\in E$, $|f(\mathbf{x}_1) -f(\mathbf{x})|\leq L||\mathbf{x}_1-\mathbf{x}_2||_1$. Let $\mathbf{y} =[f(\mathbf{x}_1),f(\mathbf{x}_2),\cdots,f(\mathbf{x}_K)]^T$ and $\mathbf{\Phi}$ is defined as \cref{prop:ann_conv}. Then,
       \begin{align}
           \sup_{\mathbf{x}\in E} |\mathbf{\Phi}(\mathbf{x}) - f(\mathbf{x})| \leq 2L [\sup_{\mathbf{x}\in E}(\min_{k}|| \mathbf{x} -\mathbf{x}_k||_1)] 
           \label{eq:supple_conv_sub}
       \end{align}

\begin{table*}[ht!]
\footnotesize
    \centering
    \setlength{\tabcolsep}{1.2pt}

    \begin{tabular}{c|l|l|l}
        \hline
        \textbf{Symbol}&\textbf{Definition} & \textbf{Description} & \textbf{Example/Meaning/Note} \\ \hline
$m$ &$\in \mathbb{Q}_{24}$ & Mantissa& $x=m\times 2^{e}$\\
$e$ &$\in \mathbb{Q}_{8}$ & Exponent& \\
$\mathbf{m}$ &$\in \mathbb{Q}^{L}_{24}$ & Mantissa Tensor with $L$ length& \\
$\mathbf{e}$ &$\in \mathbb{Q}^{L}_{8}$ & Exponent Tensor with $L$ length& \\
$\mathbf{O}$ &$\in \mathbb{R}^{L}$ & Floating point audio signal with $L$ length&$\mathbf{O}=\mathbf{m}\times 2^\mathbf{e}$ \\
$\mathbf{E}_d$ & $\in \mathbb{R}^{d\times d}$ & Identity matrix &  $\mathbf{E}_2 = \begin{bmatrix}
    1 & 0 \\
    0 & 1
\end{bmatrix}$ \\
$\mathcal{W}_d$ &$\mathbb{R}^{d_{\text{out}}\times d_{\text{in}}}$ & Weights for ANNs & Used for \cref{def:l1} and \cref{def:max} \\
$\mathbf{b}_d$ &$\mathbb{R}^{d_{\text{out}}}$ & Bias for ANNs & Used for \cref{def:l1} and \cref{def:max} \\
$\mathcal{C}^{(E,\delta),r}$ & \makecell[c]{$\min(\{n\in \mathbb{N}_0:[\exists A\subset E: ((|A|\leq n)\land$ \\ $ (\forall x\in E: \exists a \in A: \delta(a,x) \leq r))]\}\cup \{\infty\})$} & Covering numbers & $r$-convering number of $(E,\delta)$ \\
\hline
$(E,\delta)$ &  - & Metric Space & Set $E$ and its metric $\delta$\\
$\mathcal{M}$ &$ \subseteq E$ & Subset of $E$ &  \\
$\mathbf{N}$ & $\forall {\mathbf{\Phi}}$  & A set of ANNs &  \\

\hline
$\delta$ & $E \times E \rightarrow [0,\infty)$ & Metric on $E$ &\makecell[l]{Satisfy positive definiteness, symmetry, \\ and triangle inequality}  \\
$\mathbf{A}_{\mathcal{W},\mathbf{b}}$ & $\mathbb{R}^{d_{\text{in}}}\rightarrow \mathbb{R}^{d_{\text{out}}}$, $\mathbf{x} \mapsto \mathcal{W}\mathbf{x} + \mathbf{b} $ & Affine transform &  \\
 $\sigma(\cdot)$ & $\mathbb{R}^d \rightarrow \mathbb{R}^d, \quad \mathbf{x} \mapsto \max\{ \mathbf{x},0\}$ & ReLU Activation function & Applied for each elements of $\mathbf{x}$\\
 $\mathbf{\Phi}$     &  $\mathbb{R}^{\text{in}} \rightarrow \mathbb{R}^{\text{out}}$    & Artificial Neural Networks  &  In \cref{sup_sec:theory}, range is constrained to $\mathbb{R}$\\
$\mathbb{L}_d$     &  $\mathbb{R}^d \rightarrow \mathbb{R}$    & $||\mathbf{x}||_1$ representation with ANN  &  (\cref{def:l1}) 2-layer\\
 $\mathbb{M}_d$     &  $\mathbb{R}^d \rightarrow \mathbb{R}$    & $\max\{\mathbf{x}\}$ representation with ANN  &  (\cref{def:max}) An unique\\
  $\mathbb{T}_{d,K}$     &  $\mathbb{R}^d \rightarrow \mathbb{R}^{Kd}$, $\mathbf{x} \mapsto {[\mathbf{x}^{T},\mathbf{x}^{T},\cdots \mathbf{x}^{T}]^{T}}$    & $K$-times repetition of $\mathbf{x}$ with ANN  & \\
    $\mathbf{P}( \cdot, ..., \cdot ) $ &$\varprod_{i=0}^{K-1}\mathbf{N} \rightarrow \mathbf{N}$ &  Parallel of $K$ ANNs  &  \\
  $(\cdot) \bullet (\cdot) $ &$\mathbf{N}\times \mathbf{N} \rightarrow \mathbf{N}$ & Sequence of ANNs & $\mathbf{\Phi}_2 \bullet \mathbf{\Phi}_1 (\mathbf{x}) = \mathbf{\Phi}_2(\mathbf{\Phi}_1(\mathbf{x}))$ \\
    $(\cdot)_{n=k}$ &- & Functions with $k$-bit precision&  $f_{\theta,n=2}$ indicates an INR with 2-bit precision \\
        $(\cdot)^{*}$ &  & INRs satisfy \cref{eq:def_lossless}& $f^{*}_{\theta,n=4}$ indicates lossless INR with $4$-bit precision \\

    \hline
    \end{tabular}
    \caption{Notation table for the supplement material (Elements, Sets, and Functions (calculations), respectively)}
    \vspace{-15pt}
    \label{tab:notation_supple}
\end{table*}

\end{proposition}
\noindent The proof of the \cref{prop:conv_sub} accomplished by substitution of \cref{prop:_conv} to \cref{prop:ann_conv}. Generalizing \cref{eq:supple_conv_sub} complete the proof. Let $\mathcal{C}^{(E,\delta),r}$ is $r$-covering number of $(E,\delta)$. Then,
\begin{equation}
\footnotesize
    \mathcal{C}^{([a,b]^d,||\cdot||_p),r} \leq (\lceil \frac{d^{1/p} (b-a)}{2r} \rceil)^{d} \leq \begin{cases}
        \frac{d(b-a)}{r})^d  &(r< \frac{d(b-a)}{2})\\
        1 & (r \geq \frac{d(b-a)}{2})
    \end{cases}
\end{equation}
\begin{lemma}
    Let $d,K\in \mathbb{N}$, $L\in[0,\infty)$, $a\in\mathbb{R}, b\in(a,\infty)$, $f:[a,b]^d\rightarrow \mathbb{R}$ satisfies $\forall \mathbf{x}_{1,2}\in [a,b]^d$, $|f(\mathbf{x}_1) -f(\mathbf{x})|\leq L||\mathbf{x}_1-\mathbf{x}_2||_1$. And let $\mathbf{F} =\mathbf{A}_{0,f([(a+b)/2]^d)}$ Then,
    \begin{align}
        \sup_{\mathbf{x}\in[a,b]^d}|\mathbf{F}(\mathbf{x})-f(\mathbf{x})| \leq \frac{dL(b-a)}{2}.
    \end{align}
    The inequality is derived by substituting $\mathbf{x}_1 = [(a+b)/2,(a+b)/2,\cdots,(a+b)/2]^{T}$ in $|f(\mathbf{x}_1) -f(\mathbf{x})|\leq L||\mathbf{x}_1-\mathbf{x}_2||_1$.
\end{lemma}
\begin{proposition}
     Let $d\in \mathbb{N}$, $L\in[0,\infty)$, $a\in\mathbb{R}, b\in(a,\infty), r\in(0,d/4)$, $f:[a,b]^d\rightarrow \mathbb{R}$ satisfies $\forall \mathbf{x}_{1,2}\in [a,b]^d$, $|f(\mathbf{x}_1) -f(\mathbf{x})|\leq L||\mathbf{x}_1-\mathbf{x}_2||_1$. Let $\mathbf{x}_k\in\mathbb{R}^d$, and $\mathbf{y} =[y_1,y_2,\cdots y_K] \in \mathbb{R}^K$ and let $K$ satisfy $K=\mathcal{C}^{([a,b],||\cdot||_1),(b-a)r}, \sup_{\mathbf{x}}[\min_{k}||\mathbf{x}-\mathbf{x}_k||_1] \leq (b-a)r$ and $\mathbf{y} =[f(\mathbf{x}_1),f(\mathbf{x}_2),\cdots,f(\mathbf{x}_K)]^T$ and $\mathbf{\Phi}$ is defined as \cref{eq:supple_todo}. Then it holds
     \begin{align}
         \sup_{\mathbf{x}}|\mathbf{\Phi}(\mathbf{x})- f(\mathbf{x})| \leq 2L(b-a)r
     \end{align}
     This is derived by \cref{eq:supple_conv_sub} and assumption.
\end{proposition}
Then generalizing the function with the proposition as follows:
\begin{proposition}
\label{prop:last}
     Let $d\in \mathbb{N}$, $L\in[0,\infty)$, $a\in\mathbb{R}, b\in(a,\infty), r\in(0,\infty)$, $f:[a,b]^d\rightarrow \mathbb{R}$ satisfies $\forall \mathbf{x}_{1,2}\in [a,b]^d$, $|f(\mathbf{x}_1) -f(\mathbf{x})|\leq L||\mathbf{x}_1-\mathbf{x}_2||_1$. Then there exists an ANN $\mathbf{\Phi}$ s.t. 
     \begin{align}
         \sup_{\mathbf{x}}|\mathbf{\Phi}(\mathbf{x})- f(\mathbf{x})| \leq 2L(b-a)r
     \end{align}
    The definition of covering number and $K=\mathcal{C}^{([a,b],||\cdot||_1),(b-a)r} <\infty$ ensure that there exist $\mathbf{x}_k \in [a,b]^d$ s.t. 
    \begin{equation}
        \sup_{\mathbf{x}}[\min_{k} ||\mathbf{x} -\mathbf{x}_k||_1] \leq (b-a)r
    \end{equation}
\end{proposition}
\noindent Without loss of generality, $L(b-a) \neq 0$ the main theorem is thus complete by \cref{prop:last} by adjusting $r$. In conclusion, reducing the bit-precision of a digital signal is equivalent to increasing $r$, i.e., reducing $\mathcal{C}$. The number of layers and parameters is then derived by calculating the number of parameters in \cref{def:l1,def:max}.



 
\subsection{Hyperparameter}
The most important factor that determines $\mathfrak{C}$ is the Lipschitz constant $L$. The constant $L$ represents how `smooth' the signal is in a discrete setting. In discrete spaces, computing $L$ is known to be an NP-hard problem. However, it can be estimated under various assumptions. Specifically, since $L$ satisfies the inequality below, where 1) $x, y \in [a,b]^d$ are fixed-size discrete domains, and 2) $f(x) \in [0,1]$, it is possible to estimate its upper bound.

Therefore, the term $\mathfrak{C}$ in the main text is given by
\begin{equation}
    \mathfrak{C} = 9 \cdot (3d \max\{L(b-a), 1\})^{2d} \cdot d^2,
\end{equation}
where $a$ and $b$ are the same as in previous studies, i.e., -1 and 1, $d$ varies depending on the shape of the signal (Audio, Image or Video, etc.). The Lipschitz constant $L$ changes according to the domain size and the signal derivative.
For the $256\times256$ images used in the experiments, with a range of [0,1], the Lipschitz constant must satisfy $256 \leq L$ for all arbitrary signals.

\section{Quantized Representation}
\noindent\textbf{Details for hypothesis Validation}
In this section, we provide a detailed schematic diagram of \textbf{Validation} of the experiment section to avoid confusion and provide additional analysis. The quantized representation is a generalized form of our main paper's bit-plane decomposition. We use a model without a bit axis for a fair comparison with SIREN. 

\begin{figure}[t]
       \centering
       \stackunder[2pt]{\includegraphics[width=0.75in]{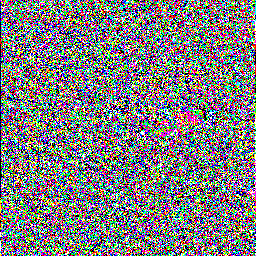}}{$\mathbf{Q}^{(0)}_1(=\mathbf{B}^{(0)})$}
       \stackunder[2pt]{\includegraphics[width=0.75in]{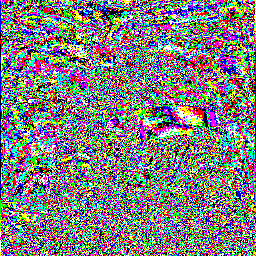}}{$\mathbf{Q}^{(1)}_1$}
       \stackunder[2pt]{\includegraphics[width=0.75in]{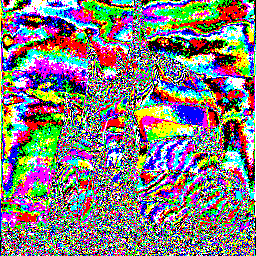}}{$\mathbf{Q}^{(2)}_1$}
    \stackunder[2pt]{\includegraphics[width=0.75in]{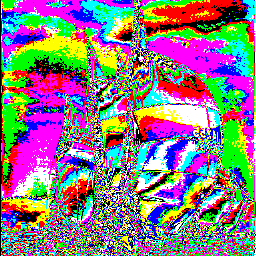}}{$\mathbf{Q}^{(3)}_1$}
    \\
           \stackunder[2pt]{\includegraphics[width=0.75in]{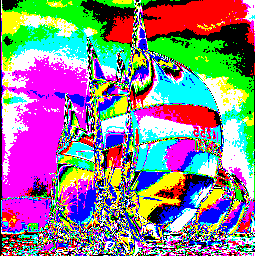}}{$\mathbf{Q}^{(4)}_1$}
       \stackunder[2pt]{\includegraphics[width=0.75in]{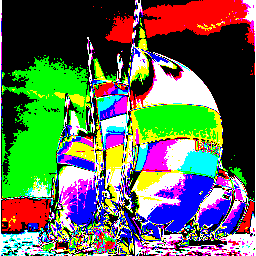}}{$\mathbf{Q}^{(5)}_1$}
       \stackunder[2pt]{\includegraphics[width=0.75in]{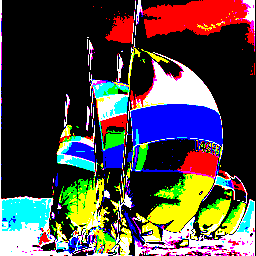}}{$\mathbf{Q}^{(6)}_1$}
    \stackunder[2pt]{\includegraphics[width=0.75in]{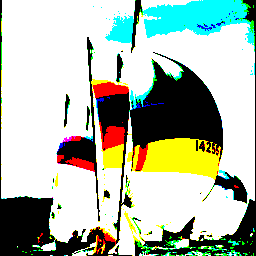}}{$\mathbf{Q}^{(7)}_1$}
    \\
       \stackunder[2pt]{\includegraphics[width=0.75in]{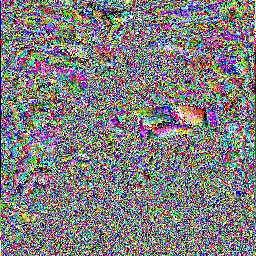}}{$\mathbf{Q}^{(0)}_2$}
       \stackunder[2pt]{\includegraphics[width=0.75in]{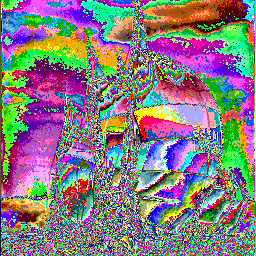}}{$\mathbf{Q}^{(1)}_2$}
       \stackunder[2pt]{\includegraphics[width=0.75in]{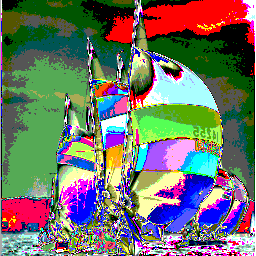}}{$\mathbf{Q}^{(2)}_2$}
    \stackunder[2pt]{\includegraphics[width=0.75in]{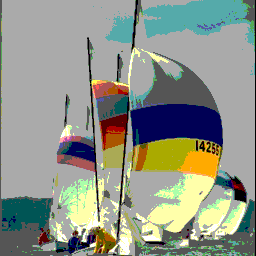}}{$\mathbf{Q}^{(3)}_2$}
    \\
   \stackunder[2pt]{\includegraphics[width=0.75in]{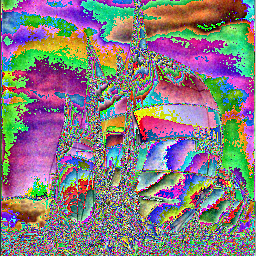}}{$\mathbf{Q}^{(0)}_4$}
       \stackunder[2pt]{\includegraphics[width=0.75in]{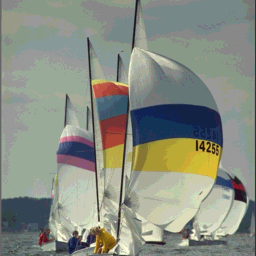}}{$\mathbf{Q}^{(1)}_4$}
       \stackunder[2pt]{\includegraphics[width=0.75in]{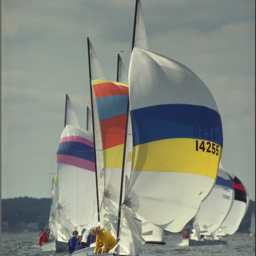}}{$\mathbf{Q}^{(0)}_8(= \mathbf{I}_8)$}
    \stackunder[2pt]{\includegraphics[width=0.75in]{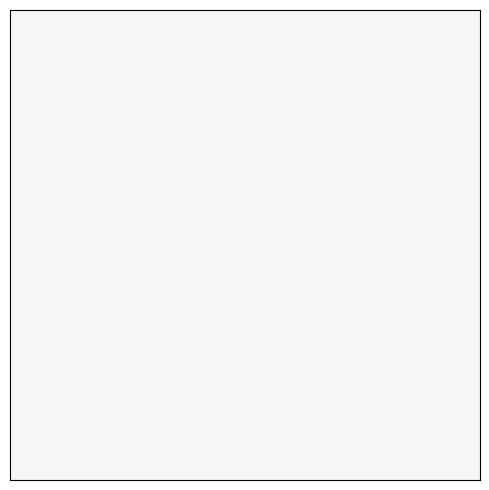}}{}
       \caption{Quantized representations depending on $k$-bit precision.}
       \label{fig:example_quantied}
   \end{figure}

Let $n$-bit images be $\mathbf{I}_n = \mathbf{Q}_n$, where $\mathbf{Q}^{(i)}_k \in Q_k^{H\times W \times 3}$.  Images are represented as $f_{n=8}: \mathbb{R}^2\rightarrow \mathbf{I}_8$ or $f_{n=16}:\mathbb{R}^2\rightarrow \mathbf{I}_{16}$ for 8-bit and 16-bit, respectively. 
A bit-plane decomposition method reduces $d$ to its divisor $k$, i.e., $k\in \text{div}^+(n)$ , thereby reducing $\mathcal{U}_d(n)$. Instead of $n$-bit images, we parameterize a quantized set of images. Then quantized images ($[\mathbf{Q}_k^{(i)}]_{i=0}^{\frac{n}{k} -1} := [\mathbf{Q}_k^{(0)},\cdots,\mathbf{Q}_k^{(\frac{n}{k}-1)}]$) is a sequence that satisfies: 
 \begin{align}
     \mathbf{I}_n = \frac{1}{2^n-1}\sum_{i=0}^{ \frac{n}{k} -1} (2^k)^i \mathbf{Q}_k^{(i)}.\label{eq:bitdecomp_generalize}
 \end{align}
   Specifically, when $\mathbf{Q}^{(i)}_{k=1}$, it is bit-plane $\mathbf{B}$ and it is the method of our main paper. We present example images of $\mathbf{Q,B,}$ and $\mathbf{I}$ in \cref{fig:example_quantied}.
  
\begin{figure}[t]
    \centering
    \includegraphics[width = 3.0in]{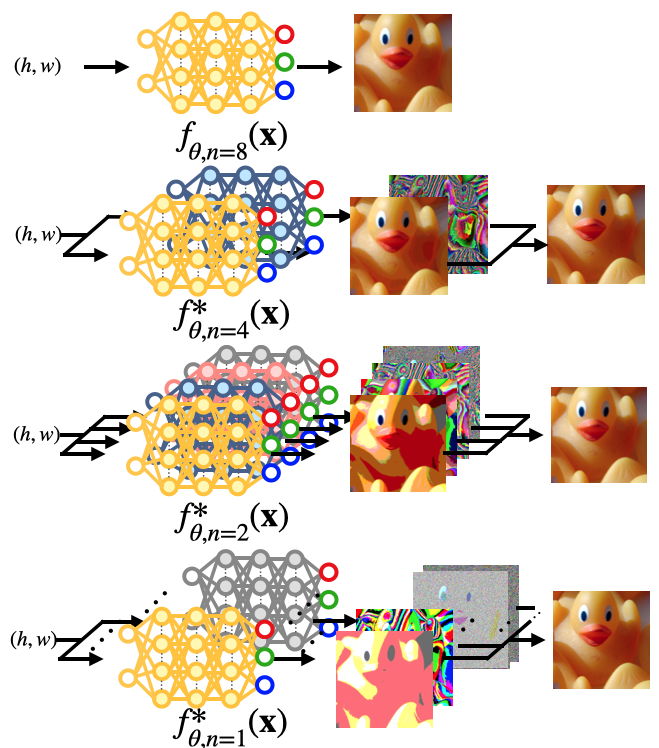}
    \caption{Schematic diagram of the parallel model used for the validation in the experiment section. $f_{\theta,n=k}$ indicates INRs that require $k$-bit precision with a given parameter $\theta$. }
    \label{fig:allschemetic}
\end{figure} 
  The validation experiment for our hypothesis is representing an $n$-bit signal is employing $\frac{n}{k}$ parallel sequence of INRs i.e. :
  \begin{align}
      [\mathbf{Q}_k^{(0)},\cdots, \mathbf{Q}_k^{(\frac{n}{k}-1)}] &\simeq [f^{(0)}_{\theta,k}, \cdots, f^{(\frac{n}{k}-1)}_{\theta,k}] \\
      \mathbf{Q}^{(i)}_k(\mathbf{x}) &\simeq f^{(i)}_{\theta,k}(\mathbf{x})
  \end{align}
We denote each INR as $f_{n=k}$, meaning the INR with $k$-bit precision. Further, $f^{\ast}_k$ indicates an INR that satisfies the required $k$-bit precision.
Note that $f_{\theta,n=8}$ indicates the baseline SIREN model. Since we set all parameters to have identical numbers, increasing the threshold of error ($\epsilon(n)$) is identical to bringing closer to the upper bound $\mathcal{U}_d(n)$.
  For example, the second row of Fig. \ref{fig:allschemetic} indicates two models that require 4-bit precision and predicting 4-MSBs and 4-LSB ($[\mathbf{Q}^{(0)}_4, \mathbf{Q}^{(1)}_4]$). 
 All evaluation follows the equation below:
 \begin{align}
  \mathbf{I}_n(\mathbf{x};{\theta}|k) = \frac{1}{2^n-1}\sum_{i=0}^{\frac{n}{k}-1} (2^k)^i \mathcal{Q}_k(\hat{f}^{(i)}_{\theta,n=k}(\mathbf{x})).
  \label{eq:summarize_method_general} 
\end{align}
\cref{eq:summarize_method_general} is a generalized form of the equation of the main paper. We provide pseudocode, \cref{alg:bitplane} and \cref{alg:quantized} for each method bit-plane decomposition and quantized representation, respectively.


\begin{algorithm}[t]
\caption{Bit-Plane Decomposition Algorithm}
\begin{algorithmic}[1]
\footnotesize
    \State \textbf{Input:} \texttt{image} (tensor), \texttt{bits} (integer)
    \State \textbf{Output:} \texttt{bit\_planes} (list) 

    \Function{bit\_decomposition}{image, bits}
        \State \texttt{bit\_planes} $\gets$ [ ] \Comment{Initialize an empty list}
        \For{$i = 0$ to \texttt{bits} - 1}
            \State \texttt{bit\_planes.append(image \% 2)}
            \State \texttt{image} $\gets$ \texttt{image} // 2 \Comment{Integer division by 2}
        \EndFor
        \State \Return \texttt{bit\_planes} \Comment{Return bit-planes}
    \EndFunction
    \end{algorithmic}
    \label{alg:bitplane}
\end{algorithm}

\begin{algorithm}[t]
\caption{Quantized Representation Algorithm}
\begin{algorithmic}[1]
\footnotesize
        \State \textbf{Input:} \texttt{bit\_planes} (list), \texttt{bits} (integer)
    \State \textbf{Output:} Quantized Representations
    
    \Function{partial\_composition}{bit\_planes, bits}
        \State \texttt{basis} $\gets$ $2^{\texttt{torch.arange}(0, \texttt{bits})}$ \Comment{Calculate basis}
        \State \texttt{n} $\gets$ $2^{(\texttt{bits})}-1$ \Comment{Normalize term}
        \State \texttt{iters} $\gets$ \texttt{len(bit\_planes)//bits} - 1 

        \State \texttt{res} $\gets$ [ ] \Comment{Initialize an empty list}

        \For{$i = 0$ to \texttt{iters}}
            \State \texttt{part $\gets$ bit\_planes[i:i+bits]}
            \State \texttt{part $\gets$ part * basis} \Comment{Multiply Bit Weight}
            \State \texttt{part $\gets$ part / n} \Comment{Normalize to [0,1]}
            \State \texttt{res.append(part)}
        \EndFor
        \State \Return \texttt{res} \Comment{Return computed value}
    \EndFunction
\end{algorithmic}
\label{alg:quantized}
\end{algorithm}

\begin{figure*}[ht]
    \centering
  \normalsize
  \vspace{0pt}
  \centering
  \begin{subfigure}[t]{1.00\textwidth}
  \centering
  \raisebox{0.15in}{\rotatebox{90}{Images}}
    \hspace{0pt}
    \stackunder[2pt]{\begin{tikzpicture}
    \node[anchor=south west, inner sep=0pt]{\includegraphics[trim={0 0 0 0},clip,height=0.8in]{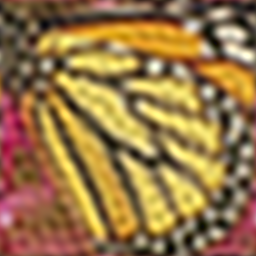}};
    \end{tikzpicture}}{}
    \hspace{-2pt}
    \stackunder[2pt]{\begin{tikzpicture}
    \node[anchor=south west, inner sep=0pt]{\includegraphics[trim={0 0 0 0},clip,height=0.8in]{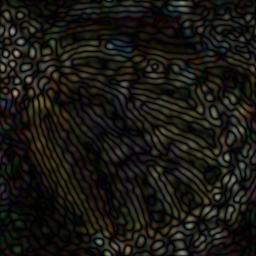}};
    \end{tikzpicture}}{}
    \hspace{-2pt}
    \stackunder[2pt]{\begin{tikzpicture}
    \node[anchor=south west, inner sep=0pt]{\includegraphics[trim={0 0 0 0},clip,height=0.8in]{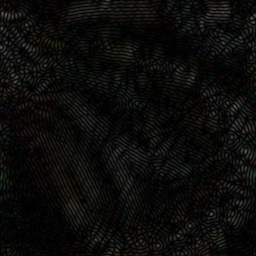}};
    \end{tikzpicture}}{}
    \hspace{-2pt}
    \stackunder[2pt]{\begin{tikzpicture}
    \node[anchor=south west, inner sep=0pt]{\includegraphics[trim={0 0 0 0},clip,height=0.8in]{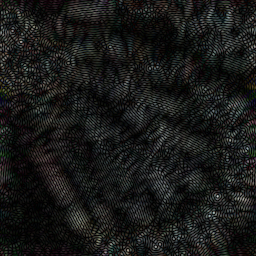}};
    \end{tikzpicture}}{}
    \hspace{-2pt}
    \stackunder[2pt]{\begin{tikzpicture}
    \node[anchor=south west, inner sep=0pt]{\includegraphics[trim={0 0 0 0},clip,height=0.8in]{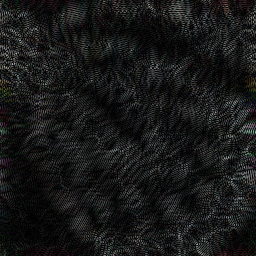}};
    \end{tikzpicture}}{}
    \hspace{-2pt}
    \stackunder[2pt]{\begin{tikzpicture}
    \node[anchor=south west, inner sep=0pt]{\includegraphics[trim={0 0 0 0},clip,height=0.8in]{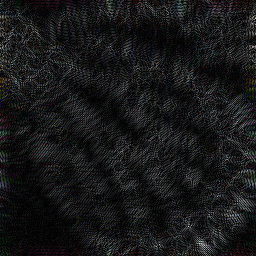}};
    \end{tikzpicture}}{}
    \hspace{-2pt}
    \stackunder[2pt]{\begin{tikzpicture}
    \node[anchor=south west, inner sep=0pt]{\includegraphics[trim={0 0 0 0},clip,height=0.8in]{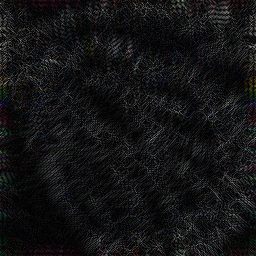}};
    \end{tikzpicture}}{}
    \hspace{-2pt}
    \stackunder[2pt]{\begin{tikzpicture}
    \node[anchor=south west, inner sep=0pt]{\includegraphics[trim={0 0 0 0},clip,height=0.8in]{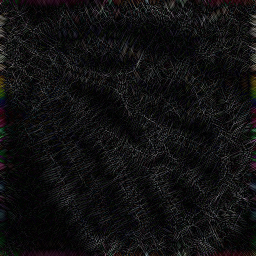}};
    \end{tikzpicture}}{{}}
    \\
    \raisebox{0.15in}{\rotatebox{90}{Spectra}}
    \hspace{-3pt}
    \stackunder[2pt]{\begin{tikzpicture}
    \node[anchor=south west, inner sep=0pt]{\includegraphics[trim={0 0 0 0},clip,height=0.8in]{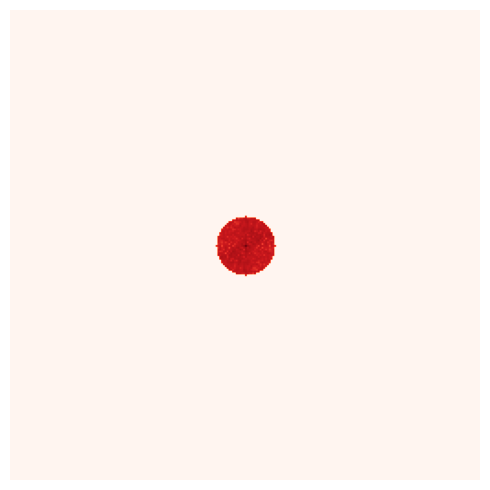}};
    \end{tikzpicture}}{{Low Frequency}}
    \hspace{-3pt}
    \stackunder[2pt]{\begin{tikzpicture}
    \node[anchor=south west, inner sep=0pt]{\includegraphics[trim={0 0 0 0},clip,height=0.8in]{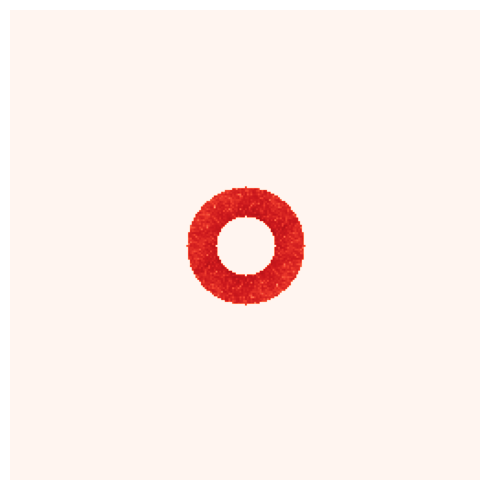}};
    \end{tikzpicture}}{}
    \hspace{-3pt}
    \stackunder[2pt]{\begin{tikzpicture}
    \node[anchor=south west, inner sep=0pt]{\includegraphics[trim={0 0 0 0},clip,height=0.8in]{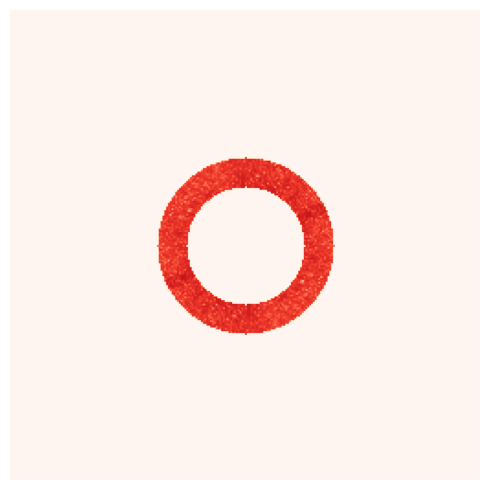}};
    \end{tikzpicture}}{}
    \hspace{-3pt}
    \stackunder[2pt]{\begin{tikzpicture}
    \node[anchor=south west, inner sep=0pt]{\includegraphics[trim={0 0 0 0},clip,height=0.8in]{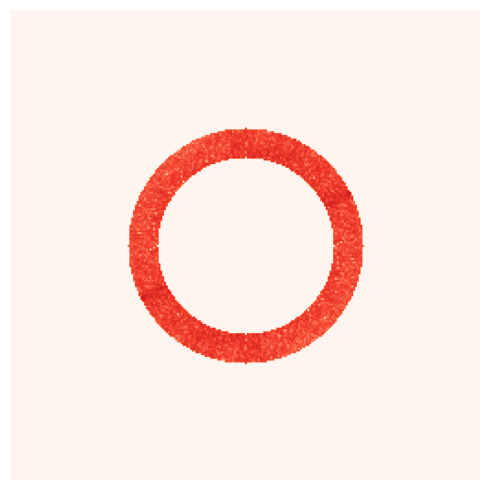}};
    \end{tikzpicture}}{}
    \hspace{-3pt}
    \stackunder[2pt]{\begin{tikzpicture}
    \node[anchor=south west, inner sep=0pt]{\includegraphics[trim={0 0 0 0},clip,height=0.8in]{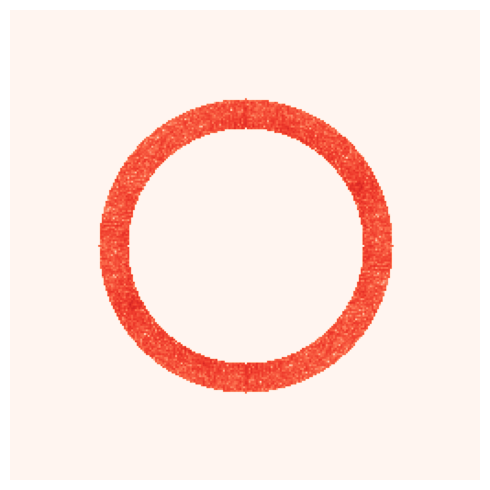}};
    \end{tikzpicture}}{}
    \hspace{-3pt}
    \stackunder[2pt]{\begin{tikzpicture}
    \node[anchor=south west, inner sep=0pt]{\includegraphics[trim={0 0 0 0},clip,height=0.8in]{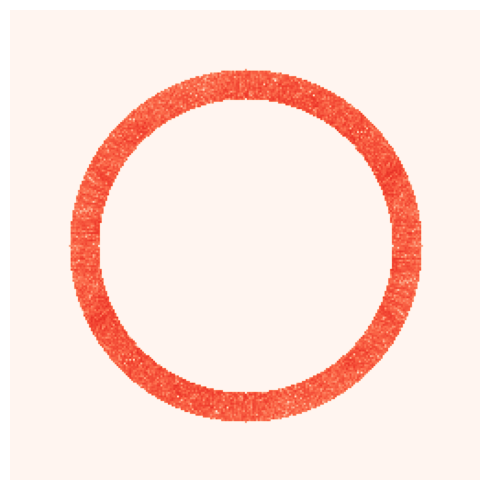}};
    \end{tikzpicture}}{}
    \hspace{-3pt}
    \stackunder[2pt]{\begin{tikzpicture}
    \node[anchor=south west, inner sep=0pt]{\includegraphics[trim={0 0 0 0},clip,height=0.8in]{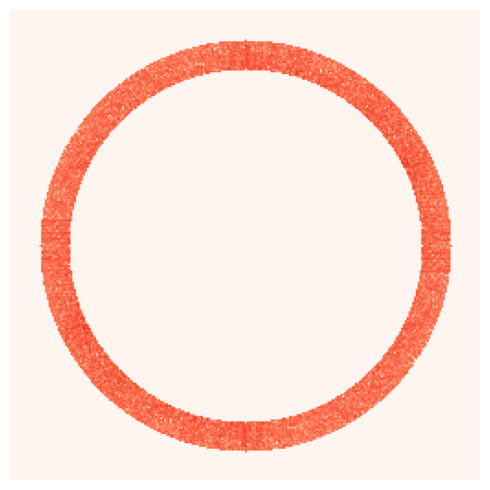}};
    \end{tikzpicture}}{}
    \hspace{-3pt}
    \stackunder[2pt]{\begin{tikzpicture}
    \node[anchor=south west, inner sep=0pt]{\includegraphics[trim={0 0 0 0},clip,height=0.8in]{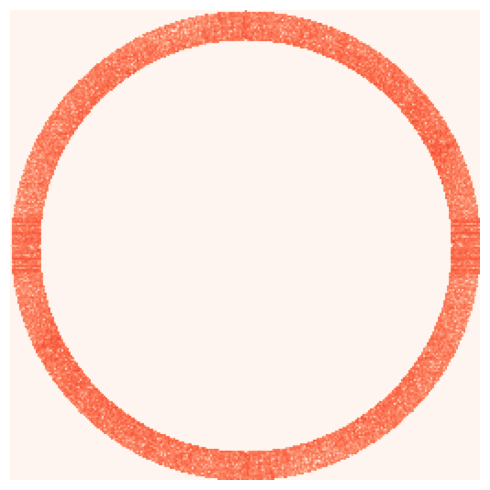}};
    \end{tikzpicture}}{{High Frequency}}
    \caption{Frequency-based image decomposition and its spectrum}
    \label{fig:spec_decomp}
  \end{subfigure}
    \\
\begin{subfigure}[t]{1.00\textwidth}
\centering
\raisebox{0.15in}{\rotatebox{90}{Images}}
    \hspace{-3pt}
    \stackunder[2pt]{\begin{tikzpicture}
    \node[anchor=south west, inner sep=0pt]{\includegraphics[trim={0 0 0 0},clip,height=0.8in]{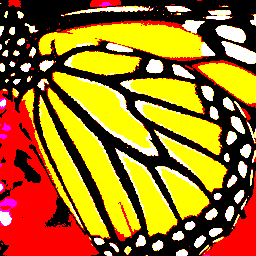}};
    \end{tikzpicture}}{}
    \hspace{-3pt}
    \stackunder[2pt]{\begin{tikzpicture}
    \node[anchor=south west, inner sep=0pt]{\includegraphics[trim={0 0 0 0},clip,height=0.8in]{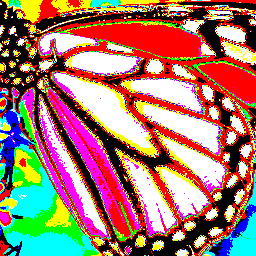}};
    \end{tikzpicture}}{}
    \hspace{-3pt}
    \stackunder[2pt]{\begin{tikzpicture}
    \node[anchor=south west, inner sep=0pt]{\includegraphics[trim={0 0 0 0},clip,height=0.8in]{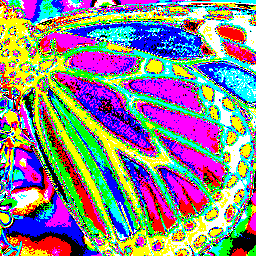}};
    \end{tikzpicture}}{}
    \hspace{-3pt}
    \stackunder[2pt]{\begin{tikzpicture}
    \node[anchor=south west, inner sep=0pt]{\includegraphics[trim={0 0 0 0},clip,height=0.8in]{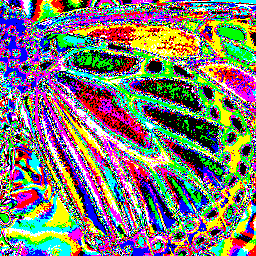}};
    \end{tikzpicture}}{}
    \hspace{-3pt}
    \stackunder[2pt]{\begin{tikzpicture}
    \node[anchor=south west, inner sep=0pt]{\includegraphics[trim={0 0 0 0},clip,height=0.8in]{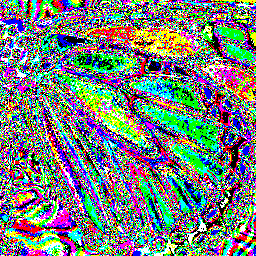}};
    \end{tikzpicture}}{}
    \hspace{-3pt}
    \stackunder[2pt]{\begin{tikzpicture}
    \node[anchor=south west, inner sep=0pt]{\includegraphics[trim={0 0 0 0},clip,height=0.8in]{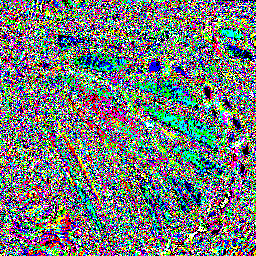}};
    \end{tikzpicture}}{}
    \hspace{-3pt}
    \stackunder[2pt]{\begin{tikzpicture}
    \node[anchor=south west, inner sep=0pt]{\includegraphics[trim={0 0 0 0},clip,height=0.8in]{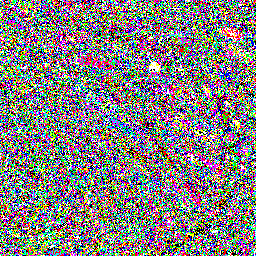}};
    \end{tikzpicture}}{}
    \hspace{-3pt}
    \stackunder[2pt]{\begin{tikzpicture}
    \node[anchor=south west, inner sep=0pt]{\includegraphics[trim={0 0 0 0},clip,height=0.8in]{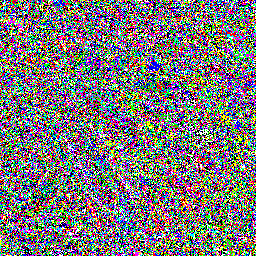}};
    \end{tikzpicture}}{}
    \\
    \raisebox{0.15in}{\rotatebox{90}{Spectra}}
    \hspace{-3pt}
    \stackunder[2pt]{\begin{tikzpicture}
    \node[anchor=south west, inner sep=0pt]{\includegraphics[trim={0 0 0 0},clip,height=0.8in]{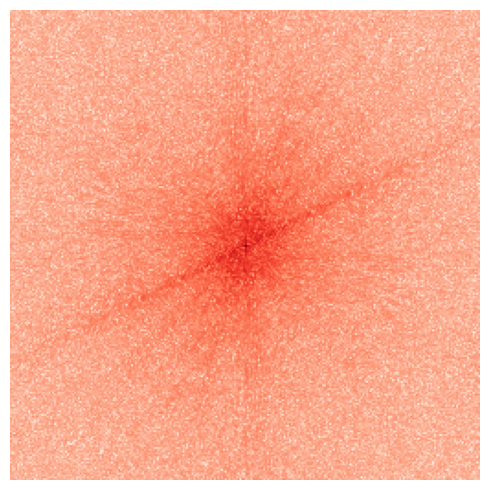}};
    \end{tikzpicture}}{MSB}
    \hspace{-3pt}
    \stackunder[2pt]{\begin{tikzpicture}
    \node[anchor=south west, inner sep=0pt]{\includegraphics[trim={0 0 0 0},clip,height=0.8in]{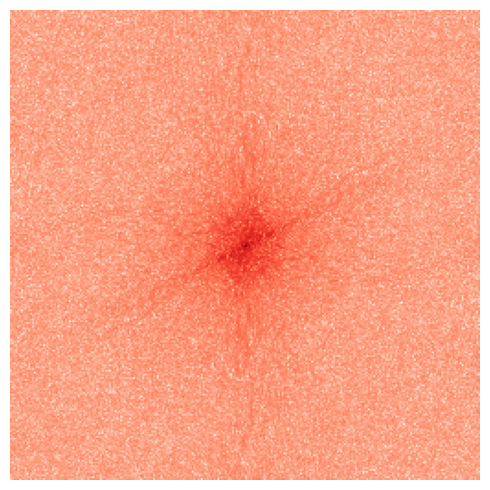}};
    \end{tikzpicture}}{}
    \hspace{-3pt}
    \stackunder[2pt]{\begin{tikzpicture}
    \node[anchor=south west, inner sep=0pt]{\includegraphics[trim={0 0 0 0},clip,height=0.8in]{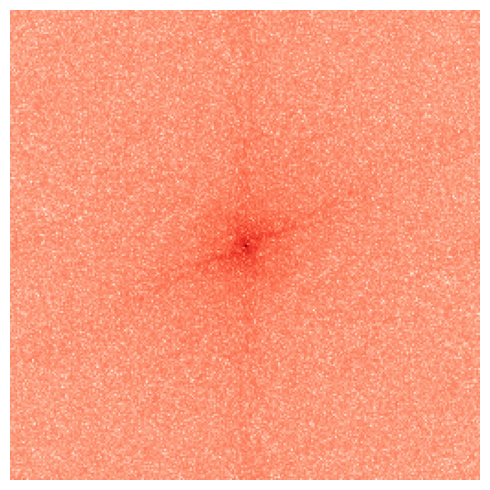}};
    \end{tikzpicture}}{}
    \hspace{-3pt}
    \stackunder[2pt]{\begin{tikzpicture}
    \node[anchor=south west, inner sep=0pt]{\includegraphics[trim={0 0 0 0},clip,height=0.8in]{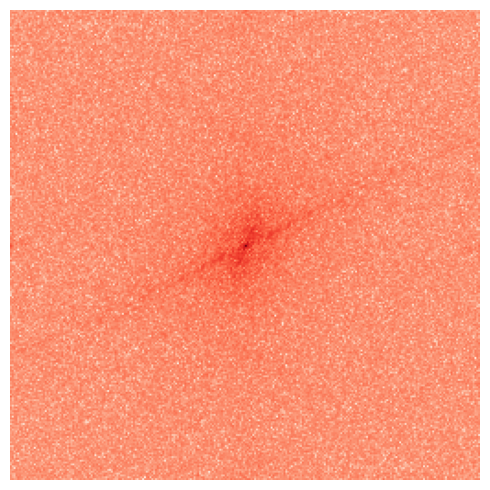}};
    \end{tikzpicture}}{}
    \hspace{-3pt}
    \stackunder[2pt]{\begin{tikzpicture}
    \node[anchor=south west, inner sep=0pt]{\includegraphics[trim={0 0 0 0},clip,height=0.8in]{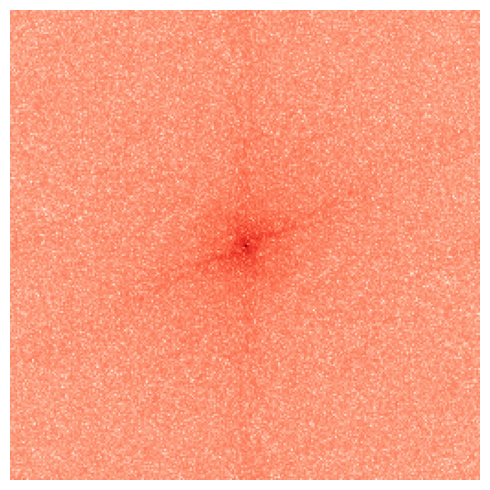}};
    \end{tikzpicture}}{}
    \hspace{-3pt}
    \stackunder[2pt]{\begin{tikzpicture}
    \node[anchor=south west, inner sep=0pt]{\includegraphics[trim={0 0 0 0},clip,height=0.8in]{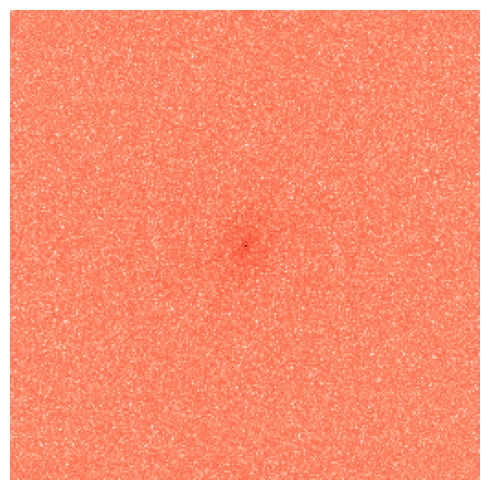}};
    \end{tikzpicture}}{}
    \hspace{-3pt}
    \stackunder[2pt]{\begin{tikzpicture}
    \node[anchor=south west, inner sep=0pt]{\includegraphics[trim={0 0 0 0},clip,height=0.8in]{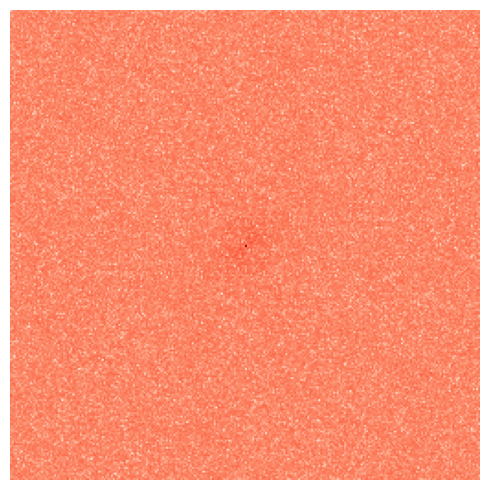}};
    \end{tikzpicture}}{}
    \hspace{-3pt}
    \stackunder[2pt]{\begin{tikzpicture}
    \node[anchor=south west, inner sep=0pt]{\includegraphics[trim={0 0 0 0},clip,height=0.8in]{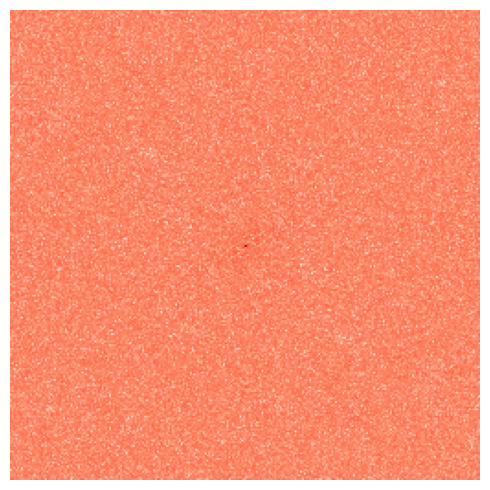}};
    \end{tikzpicture}}{LSB}
    \vspace{-0pt}
       \caption{Bit-plane image decomposition and its spectrum}
       \label{fig:bit_decomp}
  \vspace{-5pt}
    \end{subfigure}
    \caption{Frequency-based decomposition (\cref{fig:spec_decomp}) and bit-plane-based decomposition (\cref{fig:bit_decomp}). }
    \label{fig:bias_difference}
\end{figure*}

 \begin{figure*}[ht]
  \scriptsize
    \centering
    \begin{subfigure}[ht]{0.3\textwidth}
    \centering
    \raisebox{0.2in}{\rotatebox{90}{Frequecy}}
    \hspace{-20pt}
     {\includegraphics[trim=310 0 0 0,clip,height = 1.4in]{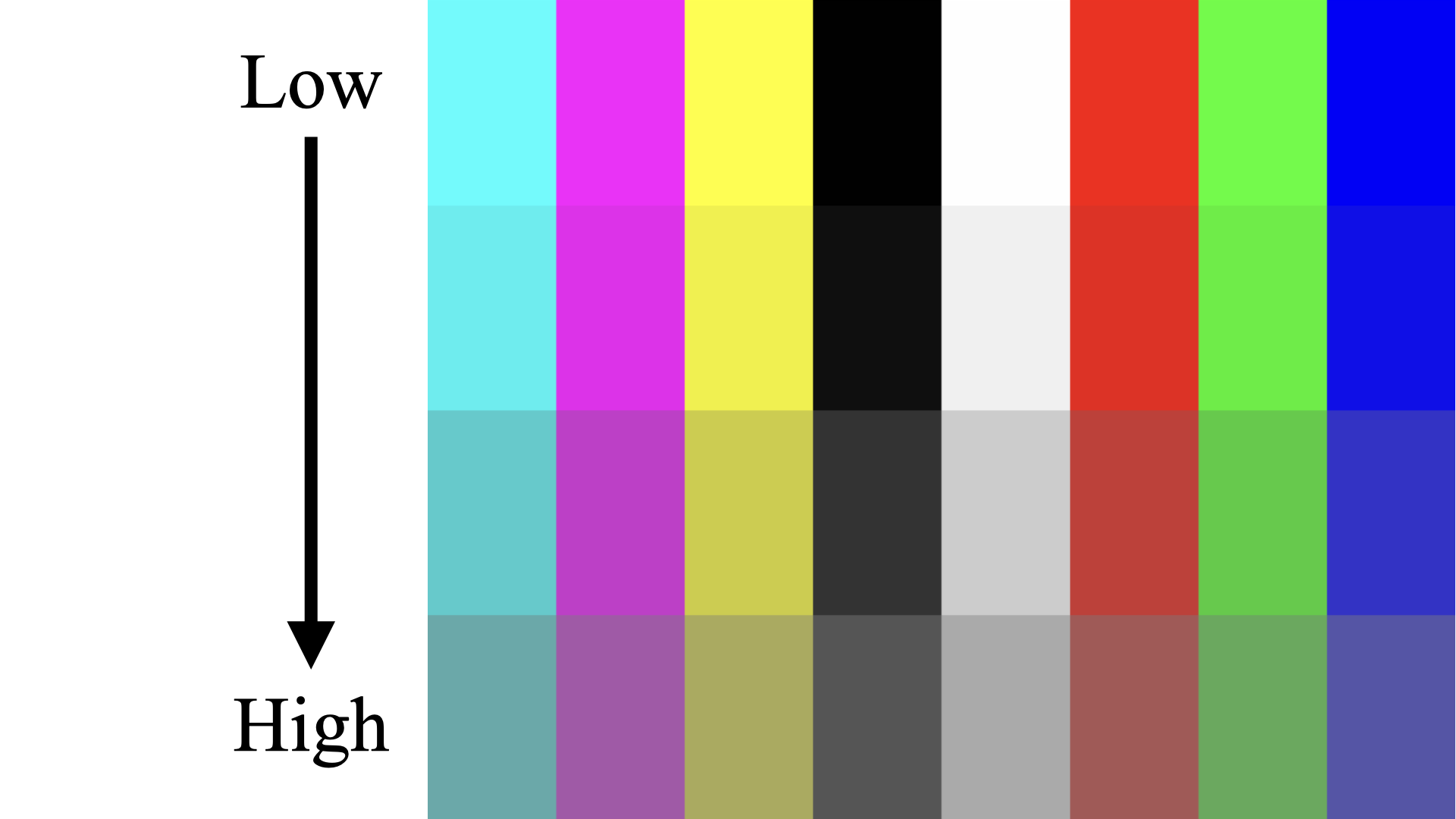}}
    \caption{16bit bit-wise bias test image}
    \label{fig:Testimage}
    \end{subfigure}
    \hspace{-10pt}
    \begin{subfigure}[ht]{0.3\textwidth}
    \centering
         {\includegraphics[trim=555 0 0 0,clip,height = 1.4in]{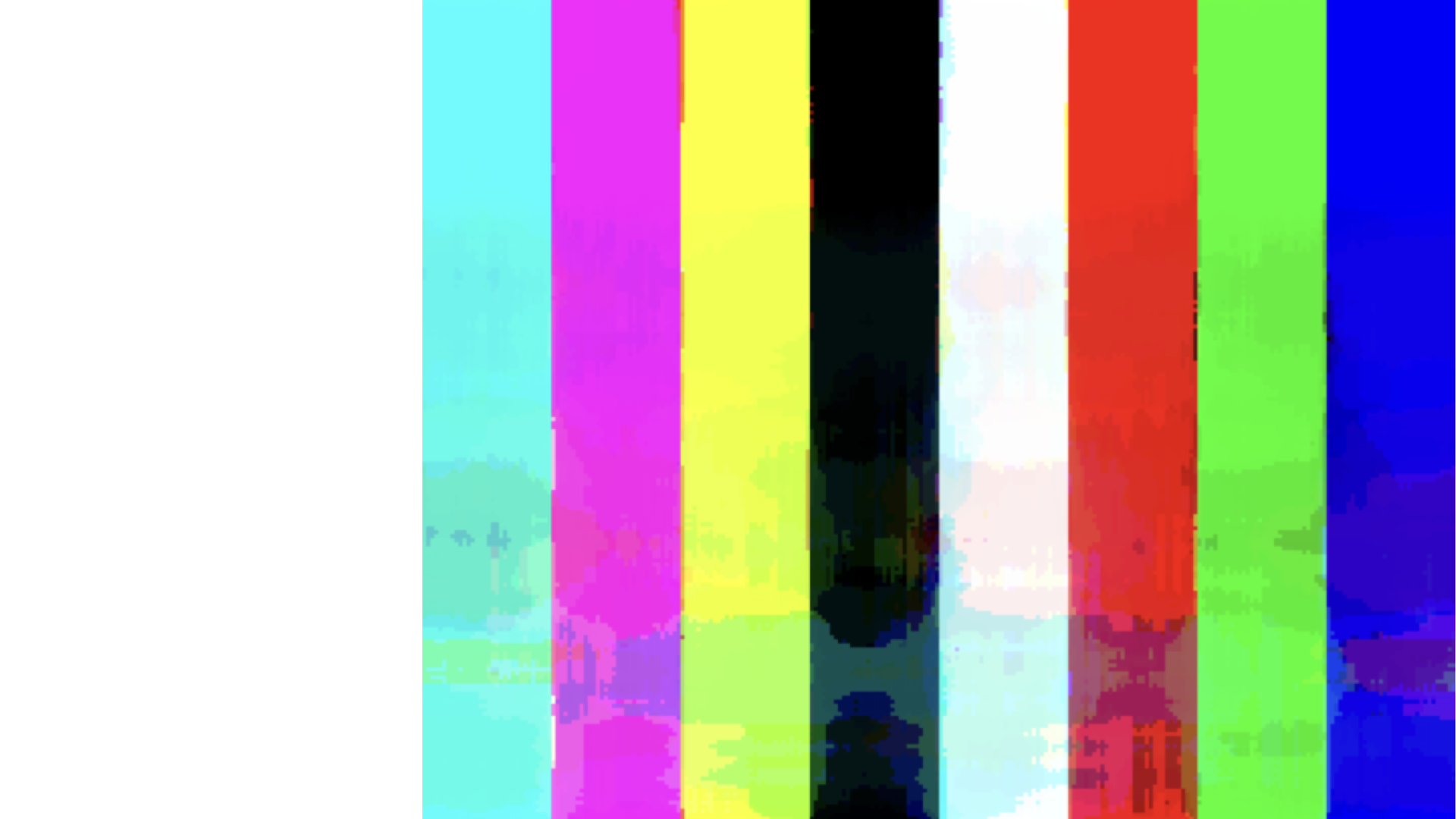}}
        \caption{Underfitted Image of Fig.\ref{fig:Testimage}}
        \label{fig:underfitted}
    \end{subfigure}
     \hspace{-10pt}
    \begin{subfigure}[ht]{0.3\textwidth}
    \centering
     {\includegraphics[trim=255 0 0 0,clip,height = 1.4in]{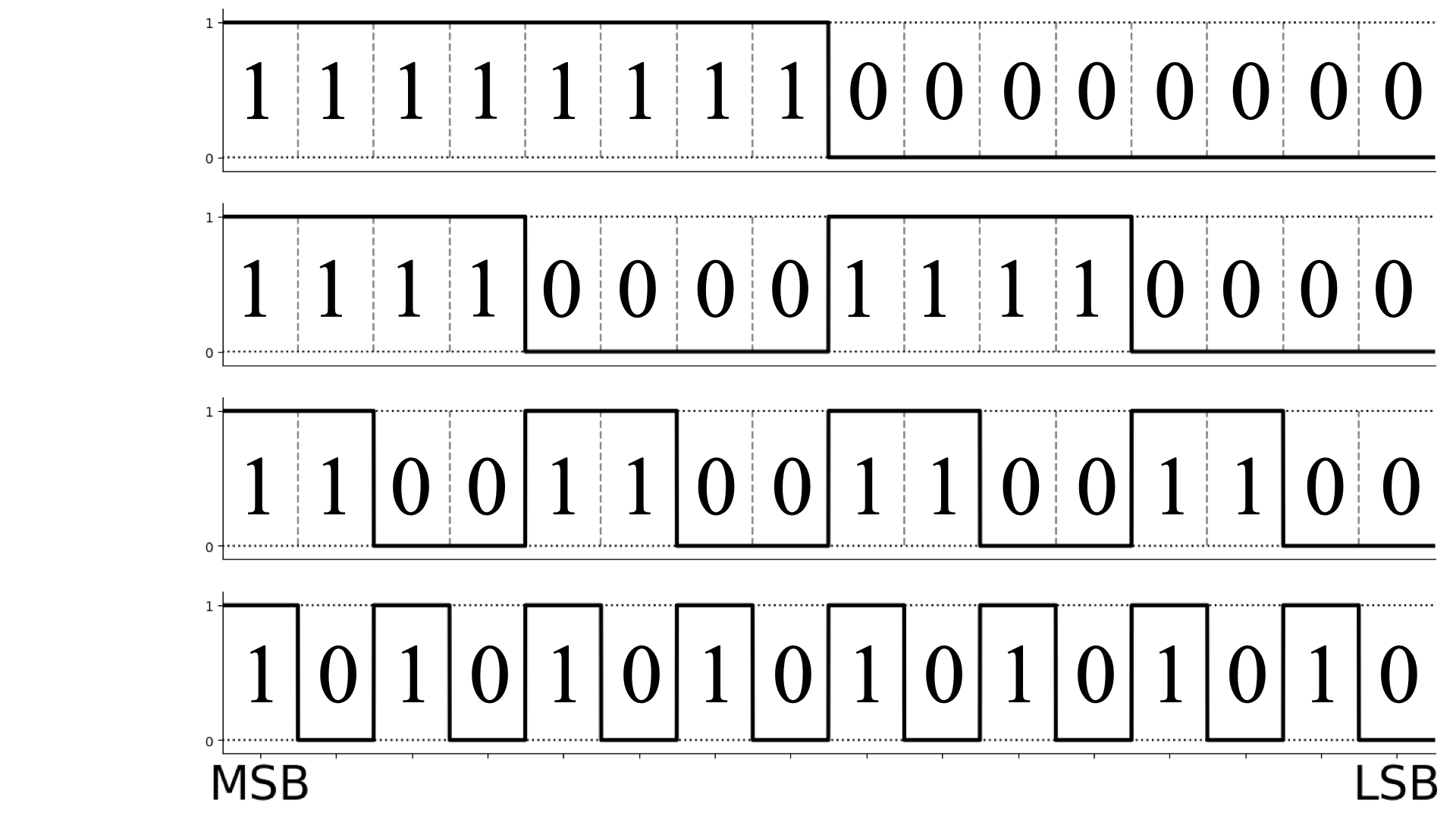}}
    \caption{Waveform in the bit-axis}
    \label{fig:waveform-fig}
    \end{subfigure}
    \caption{The sample image for the \textbf{Bit-wise bias} experiment (Fig. \ref{fig:Testimage}). The image includes different bit-frequency. The Fig. \ref{fig:waveform-fig} indicates waveforms of signals with different bit-frequency.}
    \label{fig:bitwise-bias_figure}
    \vspace{-10pt}
\end{figure*}

\begin{figure}[ht]
    \centering
    \includegraphics[width=3.1in]{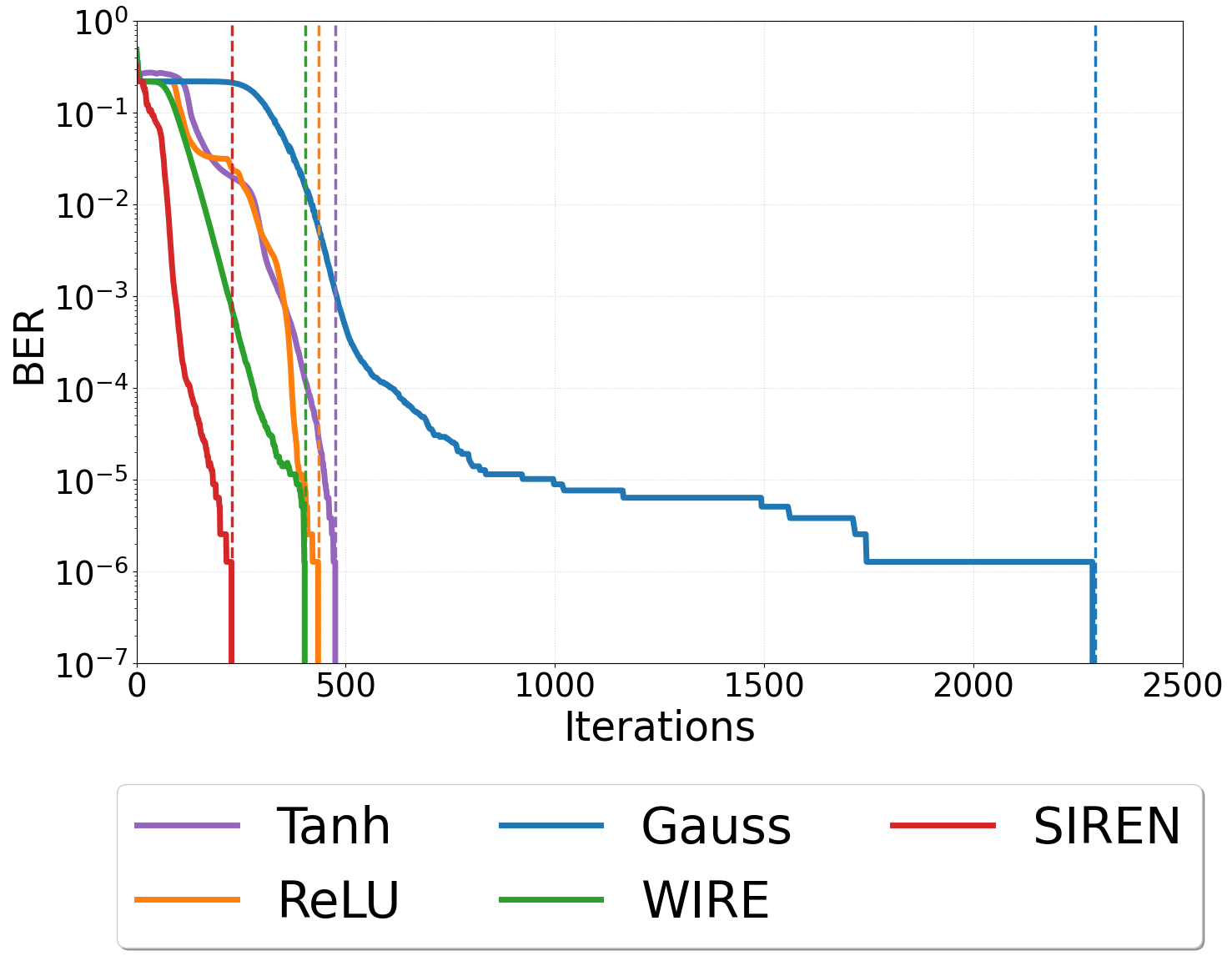}
    \caption{Quantitative ablation study on activation function. Vertical dashed lines indicate the iterations when each model achieves lossless.}
        \vspace{-22pt}
    \label{fig:abl_act}
\end{figure}

\section{Bit Bias \& Spectral Bias}
Our observation, \textit{Bit Bias}, is highly correlated to the \textit{Spectral Bias}; however, it is not identical. \cref{fig:bias_difference} provides visual information about the difference between \textit{Bit-Bias} and \textit{Spectral-Bias}.
In \cref{fig:spec_decomp}, we perform a Fourier transform on a single image and divide the frequencies into 8 bins, i.e., masking. After applying masking, we perform an inverse transform to obtain the resulting images. It shows the distribution of high-frequency components in the spatial domain, showing that high frequencies are concentrated in the wing's patterns.

In \cref{fig:bit_decomp}, we present bit-planes. For example, determining LSB possesses high-frequency components. This corresponds to a problem of determining whether each pixel value is even or odd, which is equivalent to a Bernoulli distribution with a probability of 0.5. High-frequency components are indeed present. However, these high-frequency components do not always exist in the LSB alone. The spectrum in \cref{fig:bit_decomp} shows that high-frequency components are also significantly present in the MSBs.

\begin{figure*}[!ht]
\footnotesize
\centering
\hspace{-1mm}
\raisebox{0.3in}{\rotatebox{90}{TESTIMAGES \cite{asuni2013testimages}}}%
\hspace{-1mm}
\begin{tikzpicture}[x=6cm, y=6cm, spy using outlines={every spy on node/.append style={smallwindow_w}}]
\node[anchor=south] (FigA) at (0,0) {\includegraphics[trim=0 0 0 0 ,clip,width=1.25in]{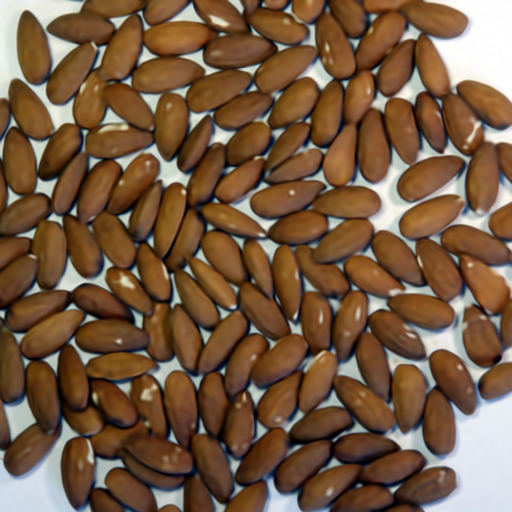}};
\spy [closeup_w_3,magnification=3] on ($(FigA)+( +0.203, -0.13)$) 
    in node[largewindow_w,anchor=east]       at ($(FigA.north) + (-0.040,-0.128)$);
\end{tikzpicture}
\hspace{-1mm}
\begin{tikzpicture}[x=6cm, y=6cm, spy using outlines={every spy on node/.append style={smallwindow_w}}]
\node[anchor=south] (FigA) at (0,0) {\includegraphics[trim=0 0 0 0 ,clip,width=1.25in]{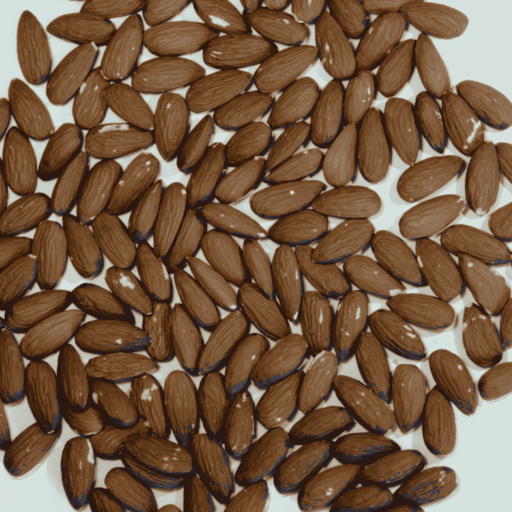}};
\spy [closeup_w_3,magnification=3] on ($(FigA)+( +0.203, -0.13)$) 
    in node[largewindow_w,anchor=east]       at ($(FigA.north) + (-0.040,-0.128)$);
\end{tikzpicture}
\hspace{-1mm}
\begin{tikzpicture}[x=6cm, y=6cm, spy using outlines={every spy on node/.append style={smallwindow_w}}]
\node[anchor=south] (FigA) at (0,0) {\includegraphics[trim=0 0 0 0 ,clip,width=1.25in]{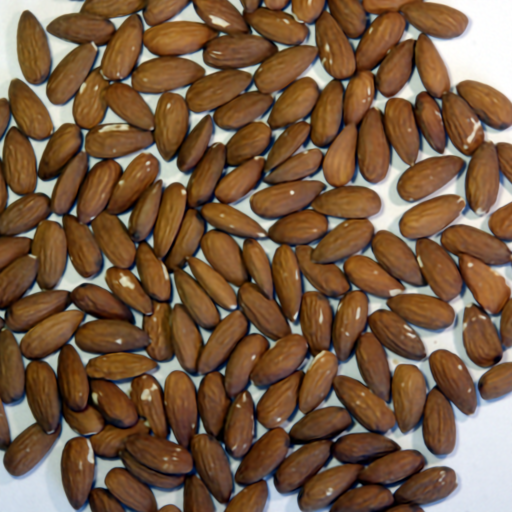}};
\spy [closeup_w_3,magnification=3] on ($(FigA)+( +0.203, -0.13)$) 
    in node[largewindow_w,anchor=east]       at ($(FigA.north) + (-0.040,-0.128)$);
\end{tikzpicture}
\hspace{-1mm}
\begin{tikzpicture}[x=6cm, y=6cm, spy using outlines={every spy on node/.append style={smallwindow_w}}]
\node[anchor=south] (FigA) at (0,0) {\includegraphics[trim=0 0 0 0 ,clip,width=1.25in]{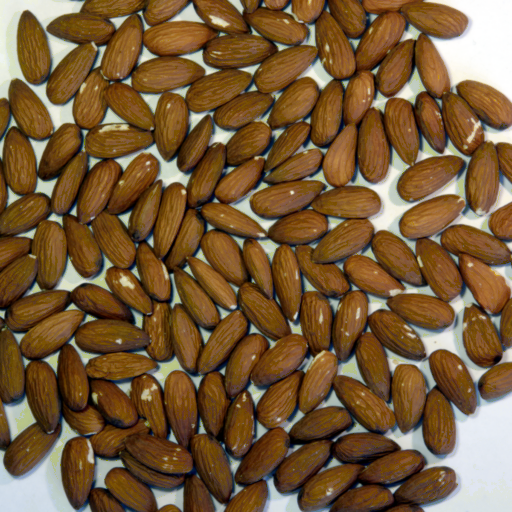}};
\spy [closeup_w_3,magnification=3] on ($(FigA)+( +0.203, -0.13)$) 
    in node[largewindow_w,anchor=east]       at ($(FigA.north) + (-0.040,-0.128)$);
\end{tikzpicture}
\hspace{-1mm}
\begin{tikzpicture}[x=6cm, y=6cm, spy using outlines={every spy on node/.append style={smallwindow_w}}]
\node[anchor=south] (FigA) at (0,0) {\includegraphics[trim=0 0 0 0 ,clip,width=1.25in]{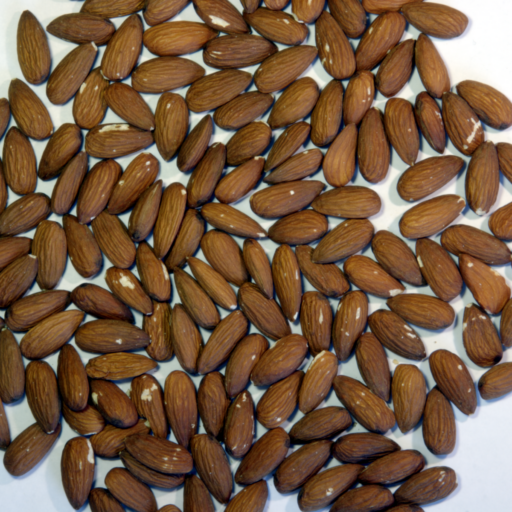}};
\spy [closeup_w_3,magnification=3] on ($(FigA)+( +0.203, -0.13)$) 
    in node[largewindow_w,anchor=east]       at ($(FigA.north) + (-0.040,-0.128)$);
\end{tikzpicture}

\raisebox{0.3in}{\rotatebox{90}{TESTIMAGES \cite{asuni2013testimages}}}%
\hspace{-1mm}
\begin{tikzpicture}[x=6cm, y=6cm, spy using outlines={every spy on node/.append style={smallwindow_w}}]
\node[anchor=south] (FigA) at (0,0) {\includegraphics[trim=0 0 0 0 ,clip,width=1.25in]{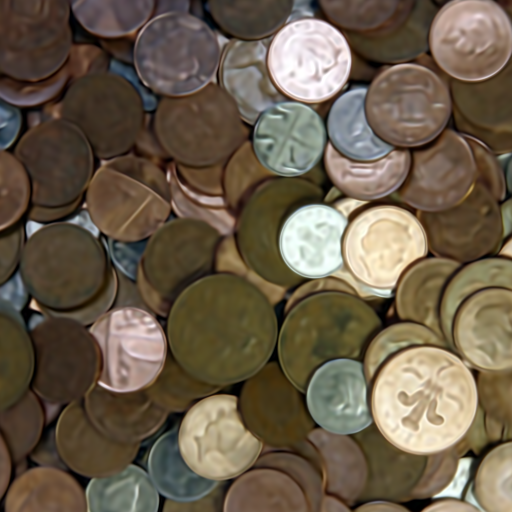}};
\spy [closeup_w_3,magnification=3] on ($(FigA)+( +0.203, -0.13)$) 
    in node[largewindow_w,anchor=east]       at ($(FigA.north) + (-0.040,-0.128)$);
\end{tikzpicture}
\hspace{-1mm}
\begin{tikzpicture}[x=6cm, y=6cm, spy using outlines={every spy on node/.append style={smallwindow_w}}]
\node[anchor=south] (FigA) at (0,0) {\includegraphics[trim=0 0 0 0 ,clip,width=1.25in]{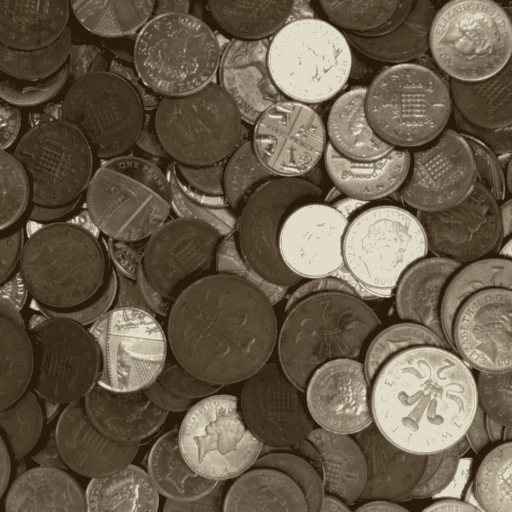}};
\spy [closeup_w_3,magnification=3] on ($(FigA)+( +0.203, -0.13)$) 
    in node[largewindow_w,anchor=east]       at ($(FigA.north) + (-0.040,-0.128)$);
\end{tikzpicture}
\hspace{-1mm}
\begin{tikzpicture}[x=6cm, y=6cm, spy using outlines={every spy on node/.append style={smallwindow_w}}]
\node[anchor=south] (FigA) at (0,0) {\includegraphics[trim=0 0 0 0 ,clip,width=1.25in]{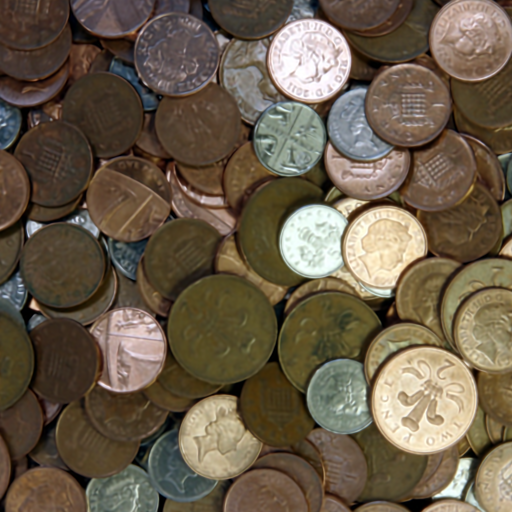}};
\spy [closeup_w_3,magnification=3] on ($(FigA)+( +0.203, -0.13)$) 
    in node[largewindow_w,anchor=east]       at ($(FigA.north) + (-0.040,-0.128)$);
\end{tikzpicture}
\hspace{-1mm}
\begin{tikzpicture}[x=6cm, y=6cm, spy using outlines={every spy on node/.append style={smallwindow_w}}]
\node[anchor=south] (FigA) at (0,0) {\includegraphics[trim=0 0 0 0 ,clip,width=1.25in]{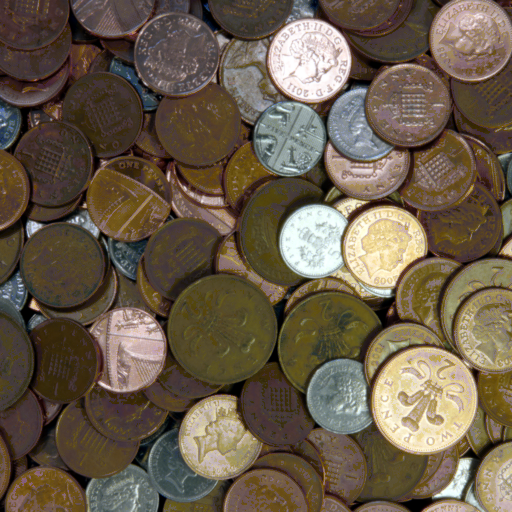}};
\spy [closeup_w_3,magnification=3] on ($(FigA)+( +0.203, -0.13)$) 
    in node[largewindow_w,anchor=east]       at ($(FigA.north) + (-0.040,-0.128)$);
\end{tikzpicture}
\hspace{-1mm}
\begin{tikzpicture}[x=6cm, y=6cm, spy using outlines={every spy on node/.append style={smallwindow_w}}]
\node[anchor=south] (FigA) at (0,0) {\includegraphics[trim=0 0 0 0 ,clip,width=1.25in]{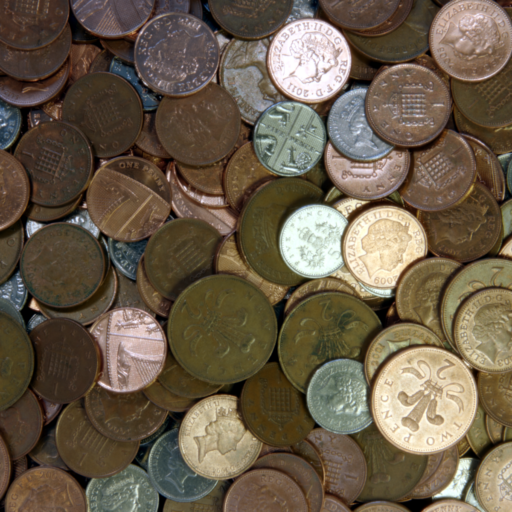}};
\spy [closeup_w_3,magnification=3] on ($(FigA)+( +0.203, -0.13)$) 
    in node[largewindow_w,anchor=east]       at ($(FigA.north) + (-0.040,-0.128)$);
\end{tikzpicture}

\hspace{-1mm}
\raisebox{0.3in}{\rotatebox{90}{Kodak \cite{Kodakdataset}}}
\hspace{-1mm}
\begin{tikzpicture}[x=6cm, y=6cm, spy using outlines={every spy on node/.append style={smallwindow_w}}]
\node[anchor=south] (FigA) at (0,0) {\includegraphics[trim=0 0 0 0 ,clip,width=1.25in]{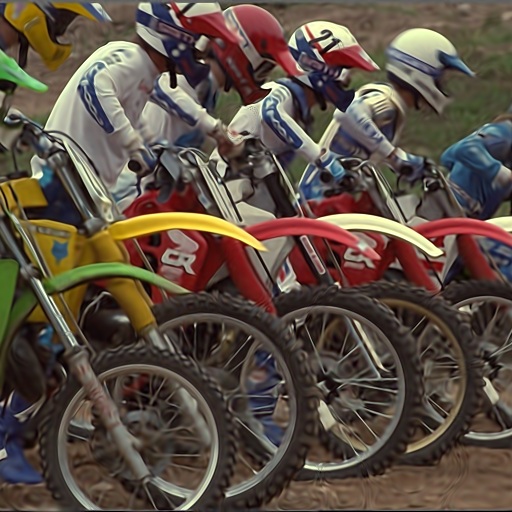}};
\spy [closeup_w_3,magnification=3] on ($(FigA)+( +0.163, 0.0)$) 
    in node[largewindow_w,anchor=east]       at ($(FigA.north) + (-0.040,-0.128)$);
\end{tikzpicture}
\hspace{-1mm}
\begin{tikzpicture}[x=6cm, y=6cm, spy using outlines={every spy on node/.append style={smallwindow_w}}]
\node[anchor=south] (FigA) at (0,0) {\includegraphics[trim=0 0 0 0 ,clip,width=1.25in]{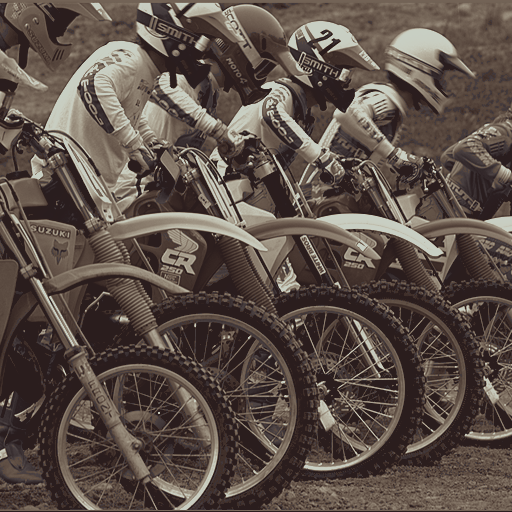}};
\spy [closeup_w_3,magnification=3] on ($(FigA)+( +0.163, 0.0)$) 
    in node[largewindow_w,anchor=east]       at ($(FigA.north) + (-0.040,-0.128)$);
\end{tikzpicture}
\hspace{-1mm}
\begin{tikzpicture}[x=6cm, y=6cm, spy using outlines={every spy on node/.append style={smallwindow_w}}]
\node[anchor=south] (FigA) at (0,0) {\includegraphics[trim=0 0 0 0 ,clip,width=1.25in]{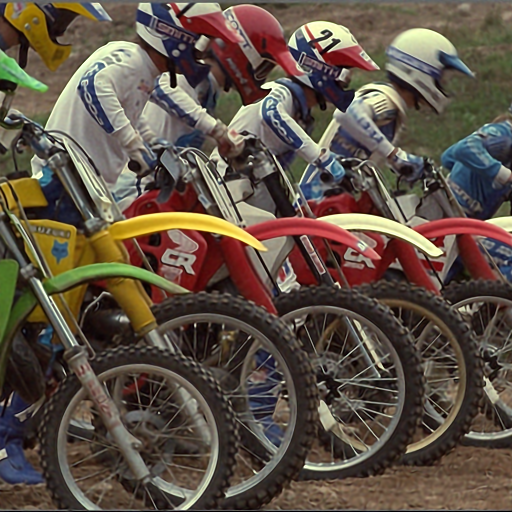}};
\spy [closeup_w_3,magnification=3] on ($(FigA)+( +0.163, 0.0)$) 
    in node[largewindow_w,anchor=east]       at ($(FigA.north) + (-0.040,-0.128)$);
\end{tikzpicture}
\hspace{-1mm}
\begin{tikzpicture}[x=6cm, y=6cm, spy using outlines={every spy on node/.append style={smallwindow_w}}]
\node[anchor=south] (FigA) at (0,0) {\includegraphics[trim=0 0 0 0 ,clip,width=1.25in]{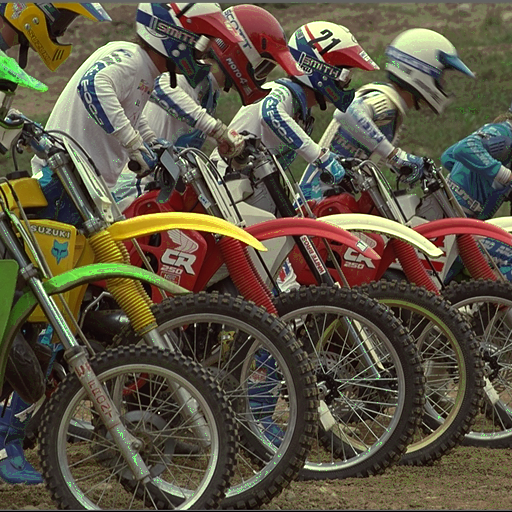}};
\spy [closeup_w_3,magnification=3] on ($(FigA)+( +0.163, 0.0)$) 
    in node[largewindow_w,anchor=east]       at ($(FigA.north) + (-0.040,-0.128)$);
\end{tikzpicture}
\hspace{-1mm}
\begin{tikzpicture}[x=6cm, y=6cm, spy using outlines={every spy on node/.append style={smallwindow_w}}]
\node[anchor=south] (FigA) at (0,0) {\includegraphics[trim=0 0 0 0 ,clip,width=1.25in]{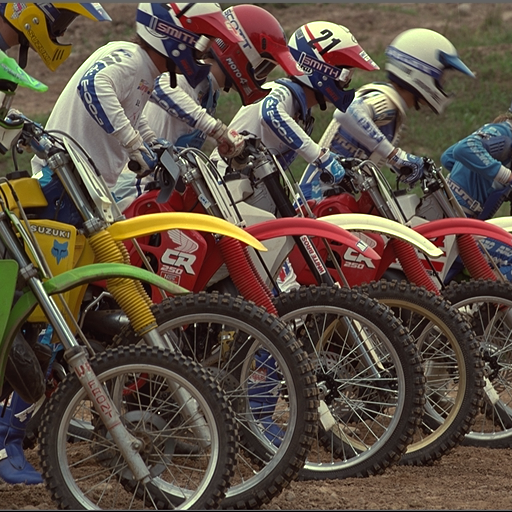}};
\spy [closeup_w_3,magnification=3] on ($(FigA)+( +0.163, 0.0)$) 
    in node[largewindow_w,anchor=east]       at ($(FigA.north) + (-0.040,-0.128)$);
\end{tikzpicture}

\hspace{-1mm}
\raisebox{0.3in}{\rotatebox{90}{Kodak \cite{Kodakdataset}}}
\hspace{-2mm}
\stackunder[2pt]{
\begin{tikzpicture}[x=6cm, y=6cm, spy using outlines={every spy on node/.append style={smallwindow_w}}]
\node[anchor=south] (FigA) at (0,0) {\includegraphics[trim=0 0 0 0 ,clip,width=1.25in]{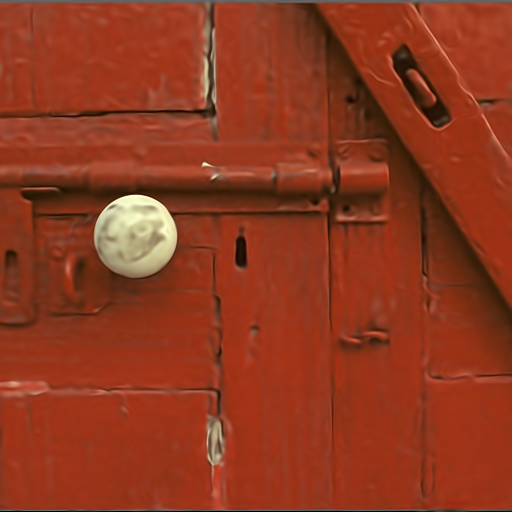}};
\spy [closeup_w_2,magnification=3] on ($(FigA)+( -0.12, +0.025)$) 
    in node[largewindow_w,anchor=east]       at ($(FigA.north) + (0.262,-0.408)$);
\end{tikzpicture}
}{SIREN \cite{sitzmann2019siren}}
\hspace{-2.5mm}
\stackunder[2pt]{
\begin{tikzpicture}[x=6cm, y=6cm, spy using outlines={every spy on node/.append style={smallwindow_w}}]
\node[anchor=south] (FigA) at (0,0) {\includegraphics[trim=0 0 0 0 ,clip,width=1.25in]{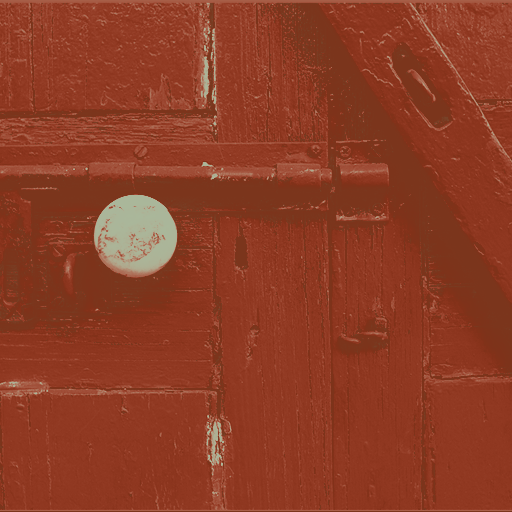}};
\spy [closeup_w_2,magnification=3] on ($(FigA)+( -0.12, +0.025)$) 
    in node[largewindow_w,anchor=east]       at ($(FigA.north) + (0.262,-0.408)$);
\end{tikzpicture}
}{{DINER} \cite{xie2023diner}}
\hspace{-2.5mm}
\stackunder[2pt]{
\begin{tikzpicture}[x=6cm, y=6cm, spy using outlines={every spy on node/.append style={smallwindow_w}}]
\node[anchor=south] (FigA) at (0,0) {\includegraphics[trim=0 0 0 0 ,clip,width=1.25in]{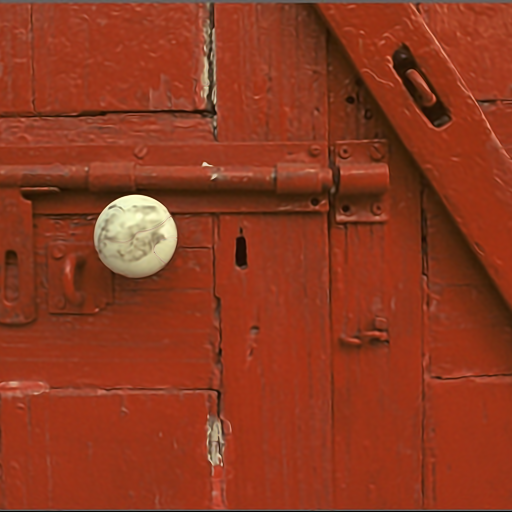}};
\spy [closeup_w_2,magnification=3] on ($(FigA)+( -0.12, +0.025)$) 
    in node[largewindow_w,anchor=east]       at ($(FigA.north) + (0.262,-0.408)$);
\end{tikzpicture}
}{{FINER} \cite{liu2023finer}}
\hspace{-2.5mm}
\stackunder[2pt]{
\begin{tikzpicture}[x=6cm, y=6cm, spy using outlines={every spy on node/.append style={smallwindow_w}}]
\node[anchor=south] (FigA) at (0,0) {\includegraphics[trim=0 0 0 0 ,clip,width=1.25in]{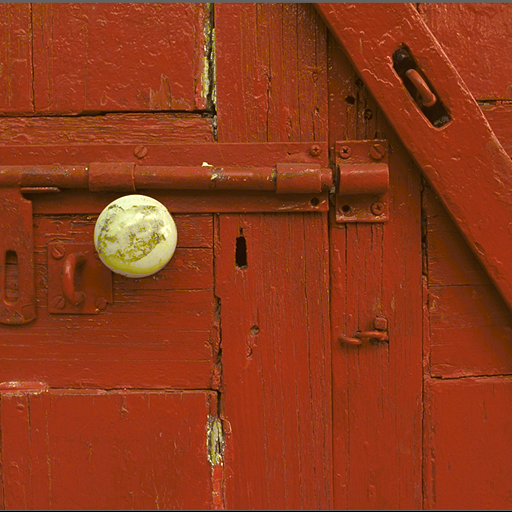}};
\spy [closeup_w_2,magnification=3] on ($(FigA)+( -0.12, +0.025)$) 
    in node[largewindow_w,anchor=east]       at ($(FigA.north) + (0.262,-0.408)$);
\end{tikzpicture}
}{\textbf{Ours$^{++}$}}
\hspace{-2.5mm}
\stackunder[2pt]{
\begin{tikzpicture}[x=6cm, y=6cm, spy using outlines={every spy on node/.append style={smallwindow_w}}]
\node[anchor=south] (FigA) at (0,0) {\includegraphics[trim=0 0 0 0 ,clip,width=1.25in]{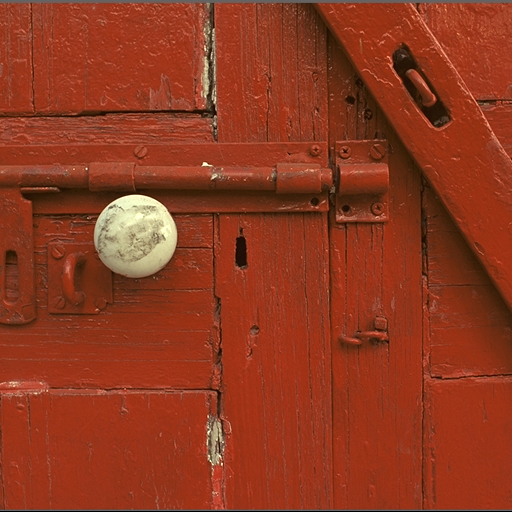}};
\spy [closeup_w_2,magnification=3] on ($(FigA)+( -0.12, +0.025)$) 
    in node[largewindow_w,anchor=east]       at ($(FigA.north) + (0.262,-0.408)$);
\end{tikzpicture}
}{{GT}}

\vspace*{-6pt}
\caption{Qualitative comparison of under-fitted images (\# of Iterations : 200) in \textbf{512$\times$512} images.}
\vspace*{-12pt}
\label{fig:suppleadd}
\end{figure*}

\section{Details for Bit-Spectral bias experiment}
 \cref{fig:bitwise-bias_figure} includes the image and its under-fitted prediction for the bit-spectral bias experiment. Extracting and comparing values with different bit frequencies from natural images is unsuitable because there are many variables, such as the spatial frequency of the image or surrounding pixels. Therefore, we control variables using the \cref{fig:Testimage} and experiment with different bit frequencies for each part as an experiment variable. \cref{fig:waveform-fig} illustrates the waveform and binary representation of each pixel value based on bit-spectral frequency. \cref{fig:underfitted} shows the qualitative result of bit-spectral bias that high-frequency values, such as 43,690 (=1010101010101010$_2$) or 21,845 (=0101010101010101$_2$), are hard to fit.

\begin{figure*}[ht!]
    \centering
    \begin{subfigure}[t]{1.\textwidth}
        \centering
            \includegraphics[width=3.1in]{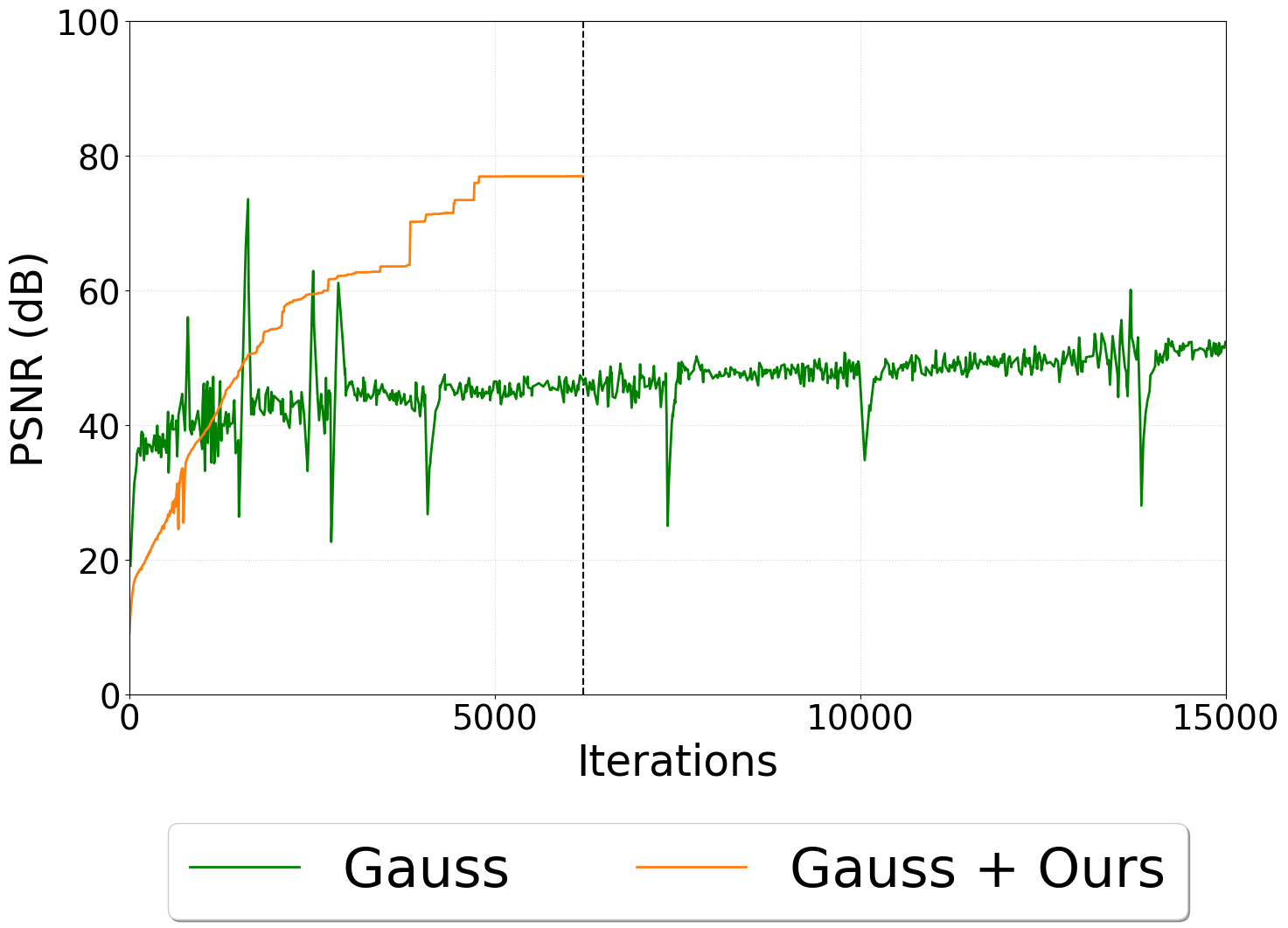}
            \includegraphics[width=3.1in]{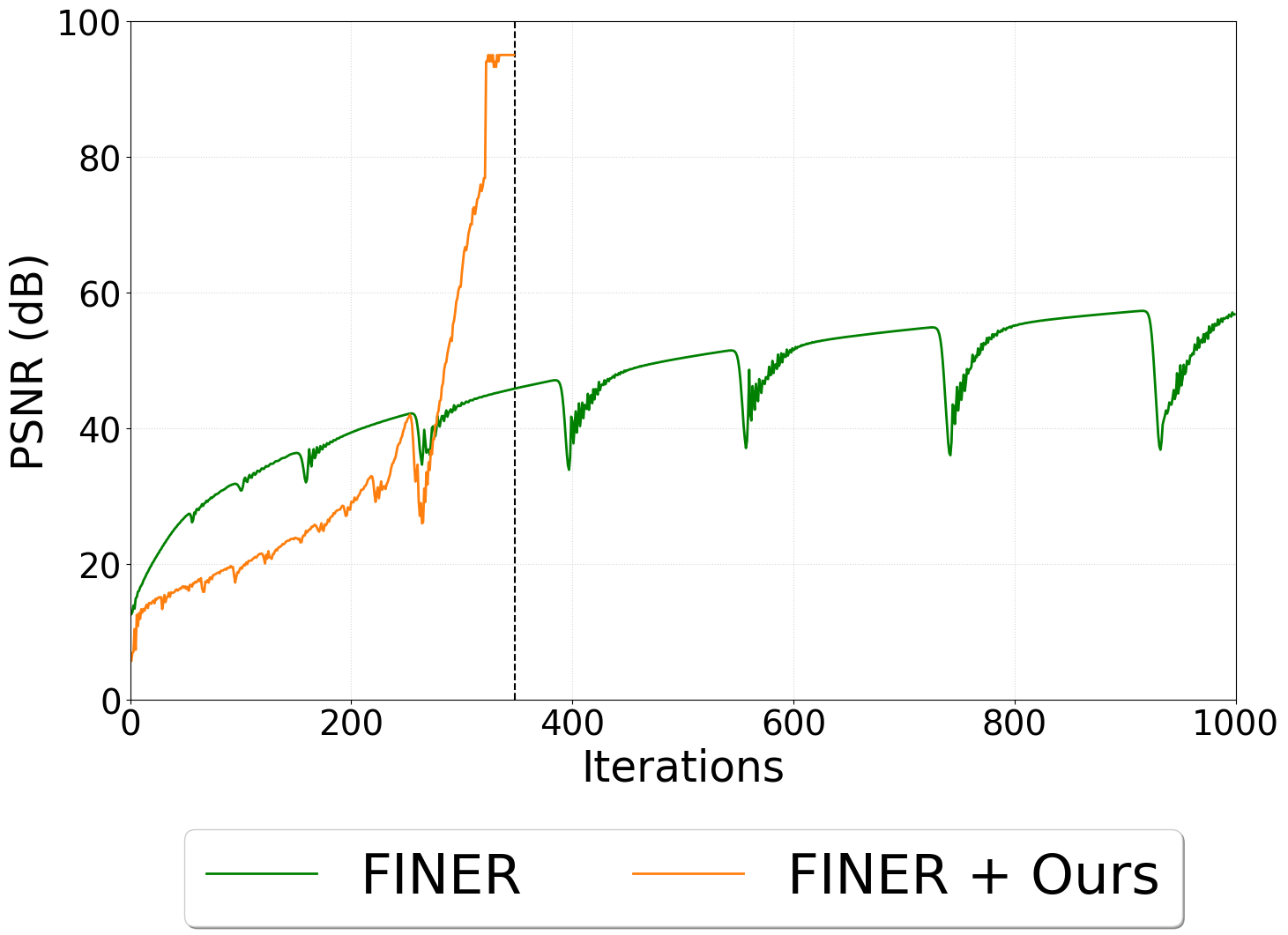}
    \caption{Activations}
    \label{fig:combinations_act}
    \end{subfigure}
    \\
    \begin{subfigure}[t]{1.\textwidth}
    \centering
            \includegraphics[width=3.1in]{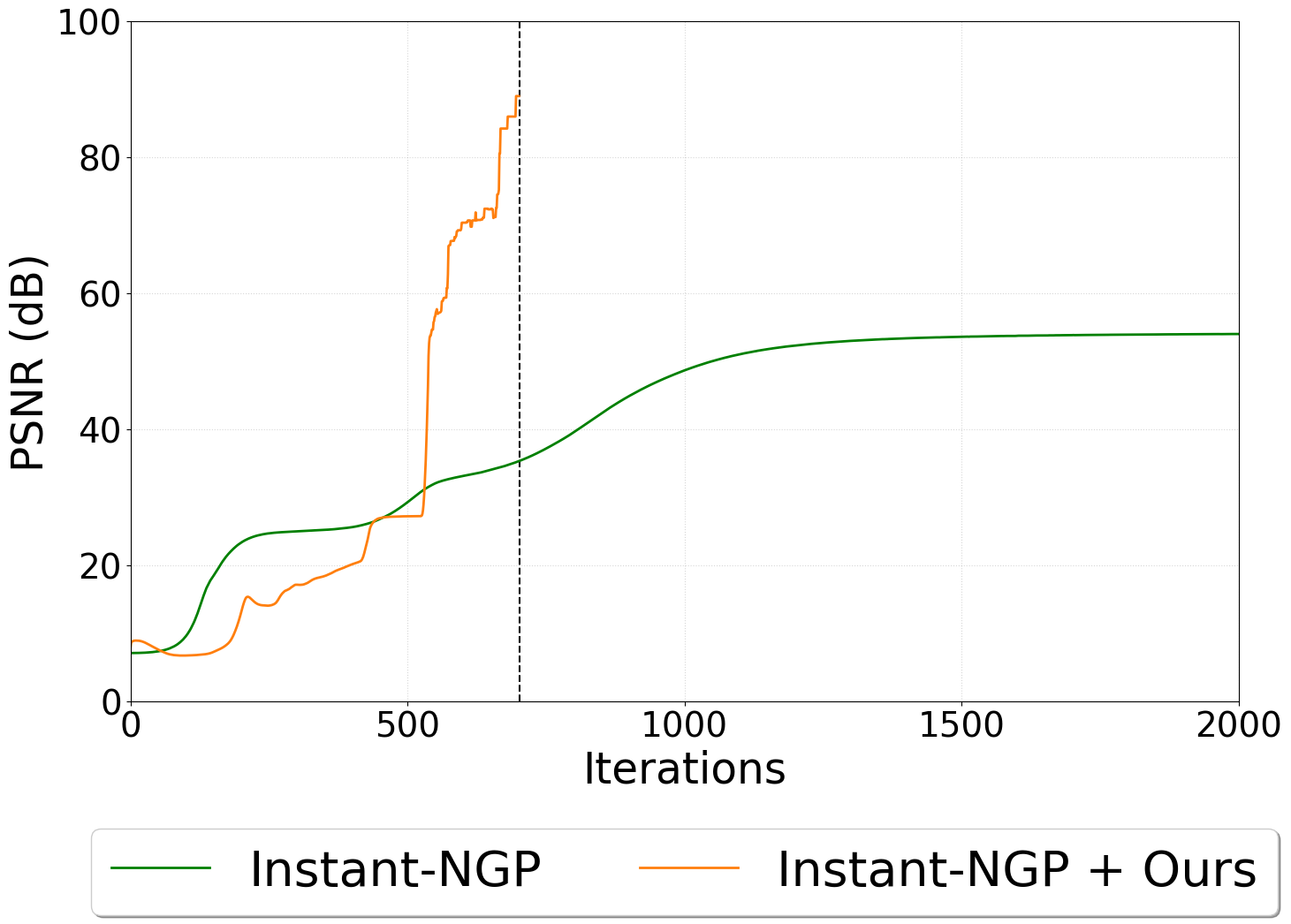}
            \includegraphics[width=3.1in]{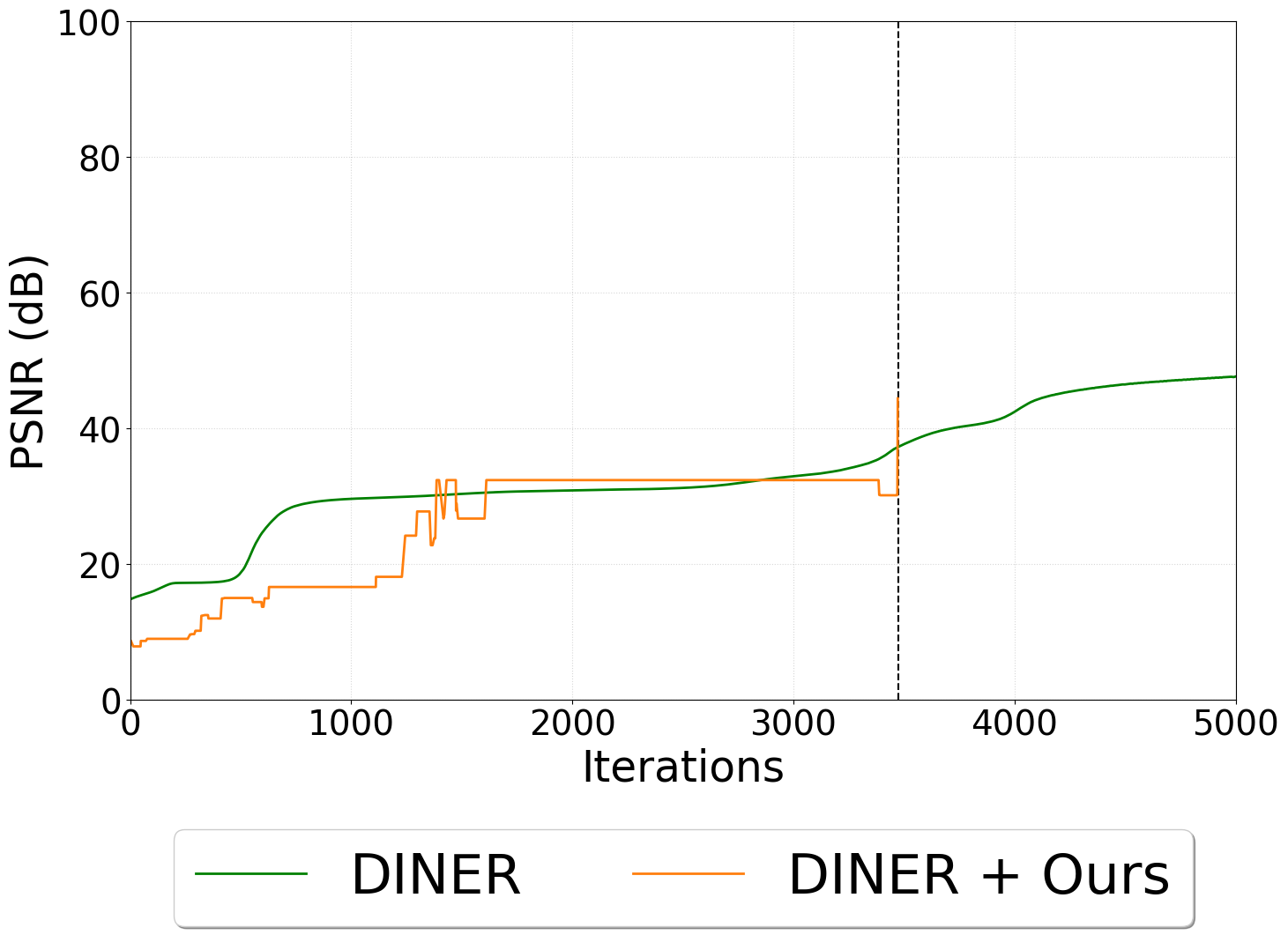}
    \caption{High-dimensional input}
     \label{fig:combinations_hash}
    \end{subfigure}

        \vspace{-10pt}
        \caption{Comparison of training curve. Vertical dashed lines indicate the iterations when our models achieve lossless.}
                \vspace{-10pt}
    \label{fig:combinations}
\end{figure*}

\section{Activation}

We conduct ablation studies for the activation and loss function of the proposed method, as shown in \cref{fig:abl_act}.
We utilized a 16-bit sample image in the TESTIMAGE dataset \cite{asuni2013testimages} and reduced the parameter count to observe convergence speed. 
 We adopt periodic activation function \cite{sitzmann2019siren} for fast conversion.
To support this, we set the activation function as the controlled variable in \cref{fig:abl_act}. In \cref{fig:abl_act}, the Gauss activation \cite{activationinr_gauss_etc} converged slower than other baselines. \cref{fig:abl_act} indicates that our method achieved lossless implicit representation with a sufficient number of iterations and parameters, regardless of the activation function used.


\section{Application Implementation Details}
\subsection{Ternary Implicit Neural Representation}
We detach bias terms in each affine linear layer for a lighter INR. Representing lossless complex images with networks consisting only of sums and differences is challenging. Unlike the image representation with full-precision parameters, the parallel network has been implemented.
We train each 1.58-bit INR from scratch per bit plane. The upper bound $\mathcal{U}_d(n)$ remains identical for 16-bit images. However, networks with limited precision have challenges due to the 3-dimensional complexity. The network to represent each bit-plane includes a 5-layer with 256 hidden channels. We replace sinusoidal activations with Gaussian Error Linear Unit (GELU) activation \cite{hendrycks2016GELU} following the prior works \cite{158bitllm, 1bitllm_bitnet}. The total number of iterations is 200K, with a learning rate scheduler decayed by a factor of 0.01 every 20K steps.
\subsection{Lossless Compression}
We conduct experiments using the MNIST and Fashion MNIST datasets. We selected 1,000 images for training and 100 images for testing. The network architecture follows RECOMBINER, with two main differences: it uses 3D coordinates as input and outputs the result using BCE Loss. The network consists of 3-layer MLP with 64 channels and sine activations.

\section{Floating Point Representation}
Our method is concentrated on presenting signals with a fixed bit-precision. However, following the standard format of the floating point representation, we expand our method to represent floating point (FP) representation. The straightforward approach is converting the numbers into binary numbers directly and aligning them to the longest bit length. We utilize the definition of floating-point representation.
The floating number is formulated as below:
\begin{align}
    x = m \times 2^e,
\end{align}
where $m\in Q_{24}$ indicates a mantissa and $e\in Q_{8}$ indicates an exponent including a sign. Note that the range of $m$ depends on the normalization method. The audio fitting experiment further supports the robustness of our approach. Audio has lower spatial complexity than an image but demands more bits. 
We serialize information, including signs for estimation and 
 recombine them as below:
 \begin{align}
     \mathbf{O}^{*}_{\theta}(\mathbf{x}) = \sum_{i=8}^{31}\mathcal{Q}_1(\mathbf{m}_{\theta,n=1}(\mathbf{x},i))\times 2^{\sum_{i=0}^{7}\mathcal{Q}_1(\mathbf{e}_{\theta,n=1}(\mathbf{x},i))},
 \end{align}
 where $\mathbf{m}_{\theta,n=1}$ and $\mathbf{e}_{\theta,n=1}$ are predicted mantissa and exponent, respectively. $\mathbf{O}$ indicates a FP32 audio signal.  We estimate each part using a single network; however, separate notations are needed to avoid confusion, i.e., $f_\theta(\mathbf{x},i) = [\mathbf{m}_{\theta}(\mathbf{x},i);\mathbf{e}_{\theta}(\mathbf{x},i)]$.


\section{Additional Results}

\noindent\textbf{Extended Model} In the main paper, we employed sinusoidal activations for generality; however, INRs with enhanced expressiveness perform more efficient results. Inspired by recent methods, we present a more efficient approach for accelerating the convergence of our INRs than using sinusoidal activations alone, as in the main paper. We utilize a hash-table \cite{xie2023diner} and modified sinusoidal activations \cite{liu2023finer}. We notate the method as `\textbf{Ours$^{++}$}'. This offers the following advantages: 1) faster convergence, 2) increased capacity for representable samples.
\begin{table}[t]
\centering
\setlength{\tabcolsep}{1.2pt}
\footnotesize
\begin{tabular}{c|c|c|c|c}
\hline
{Method} & SIREN \cite{sitzmann2019siren} & DINER \cite{xie2023diner}& FINER \cite{liu2023finer} &\textbf{Ours}\\
\hline
{Iterations ($\downarrow$)} &\multicolumn{3}{c|}{ 400}& \textbf{193}$(\pm 72)$\\
\hline
{\tiny{TESTIMAGES} \cite{asuni2013testimages}}& 36.51 &30.98 &38.71 & $\infty$\\
{Kodak \cite{Kodakdataset}}&32.91 & 32.85 &34.31 & $\infty$\\
\hline
\end{tabular}
\caption{\textbf{Quantitative comparison} on $512\times512$ image fitting with existing INR methods. The iteration number of our methods indicates `$mean(\pm std)$' for the total dataset.}
   \vspace{-5pt}
   \label{tab:supple_quan_1}
\end{table}
\begin{table}[t]
\footnotesize
\centering
\begin{tabular}{l|cc|cc}
\hline
\multirow{2}{*}{ }&  \multicolumn{2}{c|}{Kodak \cite{Kodakdataset}} & \multicolumn{2}{c}{TESTIMAGES \cite{asuni2013testimages}}\\
\cline{2-5}
 & \#Iter.($\downarrow$) & PSNR ($\uparrow$)& \#Iter.($\downarrow$) & PSNR ($\uparrow$)\\
 \hline 
SIREN + {Ours}& 790    &      \multirow{2}{*}{$\infty$}      &   3450  &      \multirow{2}{*}{$\infty$}   \\
\textbf{Ours$^{++}$}& \textbf{180}    &         &   \textbf{214}  &   \\
\hline
\end{tabular}
\vspace{-7pt}
\caption{Quantitative comparison results of \textbf{Ours$^{++}$} with our method in the main paper. The experiment has been conduct on $256\times 256$ resolutions}
\vspace{-20pt}
\label{tab:supple_quan_2}
\end{table}
\begin{table}[!t]
    \centering
    \footnotesize
    \begin{tabular}{l|c}
    \hline
     Bits Per sub-Pixel (bpsp)($\downarrow$)&  TESTIMAGES\cite{asuni2013testimages}   \\
     \hline
     \hline
           TIFF \cite{poynton1992overview_tiff} & 16.0017 \\
      JPEG2000 \cite{taubman2002jpeg2000} & 12.4021 \\
     PNG \cite{roelofs1999png} & 14.0001 \\
     \hline
     \textbf{Ours$^{++}$}& 10.4411 \\
     \hline
    \end{tabular}
    \vspace{-7pt}
     \caption{Quantitative Comparison for lossless compression on 16-bit images. }
     \vspace{-5pt}
     \label{tab:sup_compression}
\end{table}

\begin{table}[h]
\centering
\vspace{-10pt}
\footnotesize
\begin{tabular}{l|ccccc}
\hline
 & \multicolumn{3}{c}{Experiment Group}& SIREN \cite{sitzmann2019siren}& Ours\\
\hline
Bit-precision ($n$)&1 &2 & 4 & 8 & 1\\
\hline
\hline
\#Params. (M) &  \textbf{1.311} & 1.322 & 1.318 & 1.316 & 1.316 \\
Mem. (MB) & 14.17 & 14.17 & 14.16 & \textbf{14.15} & 14.18 \\
FLOPs (M) &  \textbf{1.303} & 1.316 &1.314 & 1.313 & 10.51 \\
Time (ms)&3.116 & 1.823 &0.922 & 0.771 & \textbf{0.761}\\
\hline
\end{tabular}
\vspace{-10pt}

\caption{Comparison of computational resource usage (parameters, memory, FLOPs, and time) among SIREN \cite{sitzmann2019siren} and our method for bit-precision settings.}
\vspace{-5pt}
      \label{tab:comp_time_base}

\end{table}

\begin{table}[!h]
    \scriptsize
    \centering
    \vspace{-5pt}
    \begin{tabular}{l|c}
    \hline
      &  Time(ms)  \\
     \hline
     \hline
      RECOMBINER \cite{he2023recombiner} &  0.455 \\
      RECOMBINER + \textbf{Ours}&0.876\\
    \hline
        \end{tabular}
        \vspace{-10pt}
        \caption{Decoding time for a single image used in the compression.}
        \vspace{-15pt}
        \label{tab:comp_time_compress}
\end{table}

We conduct 2D image fitting experiments on 512$\times$512 images which is a larger resolution than our main paper. \cref{fig:suppleadd} show that our method converges faster than other approaches while preserving details. \cref{tab:supple_quan_1} demonstrates that the applied method converges much faster while achieving lossless representation. Additionally, \cref{tab:supple_quan_2} shows that `Ours$^{++}$' converges faster than the method in the main paper.

\noindent\textbf{Training Curve} In \cref{fig:combinations}, we provide additional training curves that could not be included in the main text due to space constraints. These curves illustrate trends when combined with each model: high-dimensional inputs (\cref{fig:combinations_hash}) and activations (\cref{fig:combinations_act}).

\noindent \textbf{Lossless Compression} We observed that the hash table generated by our method (Ours$^{++}$) has low entropy, making it highly suitable for compression. \cref{tab:sup_compression} present a quantitative result on compressing 16-bit images. We applied quantization to the hash table, followed by entropy coding. Despite the challenges of compressing 16-bit images, this approach outperforms traditional codecs, demonstrating superior performance.

\noindent \textbf{Computational Complexity} In \cref{tab:comp_time_base}, we present a comparison of the computational resources including a SIREN \cite{sitzmann2019siren} and our experiment group in \cref{tab:hypothesis_test} of the main paper.
FLOPs and time are reported for all models based on the computation of a single pixel. Our method requires FLOPs proportional to the bit depth linearly, but memory usage and computation time remain nearly unchanged. Due to the parallel processing nature of GPUs, the computation time shows marginal differences. In \cref{tab:comp_time_compress},  We show a decoding time for a single image used in the compression, demonstrating marginal differences.

\end{document}